\documentclass[10pt,twocolumn,letterpaper]{article}

\usepackage{iccv}
\usepackage{times}
\usepackage{epsfig}
\usepackage{graphicx}
\usepackage{amsmath}
\usepackage{amssymb}
\usepackage{datetime2}
\usepackage{bm}
\usepackage{color,soul}          
\usepackage{microtype}      
\usepackage{dcolumn}        
\usepackage{url}            
\usepackage{booktabs}       
\usepackage{amsfonts}       
\usepackage{nicefrac}       
\usepackage{enumitem}
\usepackage{multirow}
\usepackage{float}
\usepackage{algorithm}
\usepackage[noend]{algpseudocode}
\usepackage{amsthm}
\usepackage{subfiles}
\usepackage{adjustbox}
\usepackage{subcaption}
\usepackage{makecell}
\usepackage[normalem]{ulem}
\usepackage{caption}
\usepackage{nopageno}
\usepackage[flushmargin,para]{footmisc}

\usepackage[pagebackref=true,breaklinks=true,letterpaper=true,colorlinks,bookmarks=false]{hyperref}
\iccvfinalcopy 
\def\iccvPaperID{6754} 

\ificcvfinal\fi

\begin{document}

\title{Multi-Instance Pose Networks: Rethinking Top-Down Pose Estimation}

\author{{Rawal Khirodkar}\textsuperscript{{1\thanks{Work done during an internship at Amazon}}} \hspace{1cm} {Visesh Chari}\textsuperscript{{2\thanks{Now at Waymo}}} \hspace{1cm} {Amit Agrawal}\textsuperscript{2} \hspace{1cm} {Ambrish Tyagi}\textsuperscript{2}\\
\textsuperscript{1}Carnegie Mellon University \hspace{1cm} \textsuperscript{2}Amazon Lab 126\\
{\urlstyle{sf} \href{https://rawalkhirodkar.github.io/mipnet}{https://rawalkhirodkar.github.io/mipnet}}
} 


\maketitle


\begin{abstract}
\vspace*{-0.3cm}
A key assumption of top-down human pose estimation approaches is their expectation of having a single person/instance present in the input bounding box. This often leads to failures in crowded scenes with occlusions. We propose a novel solution to overcome the limitations of this fundamental assumption. Our Multi-Instance Pose Network (MIPNet) allows for predicting multiple 2D pose instances within a given bounding box. We introduce a Multi-Instance Modulation Block (MIMB) that can adaptively modulate channel-wise feature responses for each instance and is parameter efficient. We demonstrate the efficacy of our approach by evaluating on COCO, CrowdPose, and OCHuman datasets. Specifically, we achieve $70.0$ AP on CrowdPose and $42.5$ AP on OCHuman test sets, a significant improvement of $2.4$ AP and $6.5$ AP over the prior art, respectively. When using ground truth bounding boxes for inference, MIPNet achieves an improvement of $0.7$ AP on COCO, $0.9$ AP on CrowdPose, and $9.1$ AP on OCHuman validation sets compared to HRNet. Interestingly, when fewer, high confidence bounding boxes are used, HRNet's performance degrades (by $5$ AP) on OCHuman, whereas MIPNet maintains a relatively stable performance (drop of $1$ AP) for the same inputs.
\end{abstract}

\vspace*{-1.0cm}
\section{Introduction}

 \begin{figure}[t]
 \captionsetup{font=small, skip=0pt}
 \begin{center}
  \includegraphics[width=0.9\linewidth,height=0.8\linewidth]{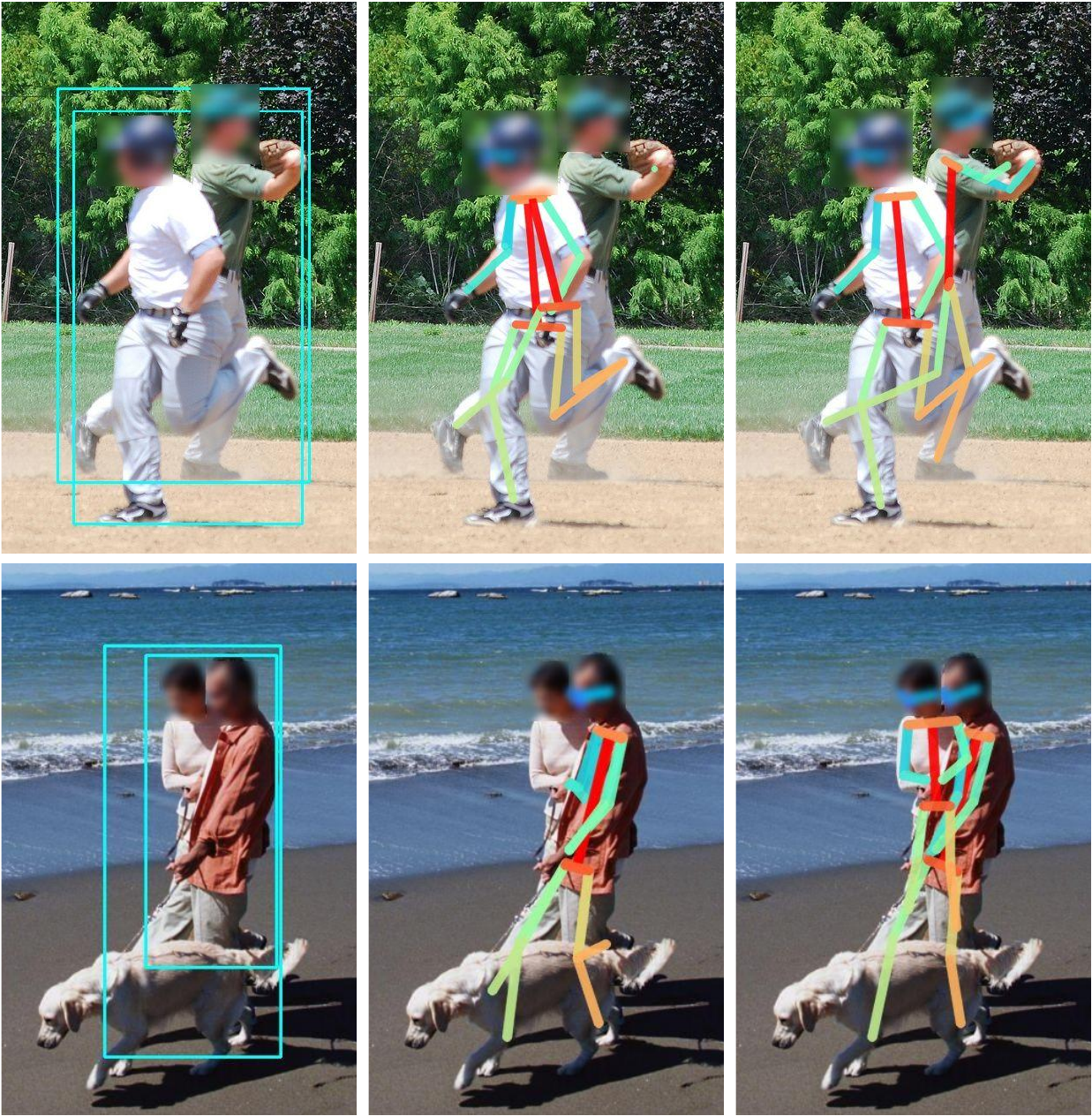}
 \end{center}
 \vspace*{-0.1in}
 \caption{2D pose estimation networks often fail in presence of heavy occlusion. (Left) Bounding boxes corresponding to two persons. (Middle) For both bounding boxes, HRNet predicts the pose for the front person and misses the occluded person. (Right) MIPNet allows multiple instances for each bounding box and recovers the pose of the occluded person.}
\vspace*{-0.3in}
\label{fig:introduction}
 \end{figure}

Human pose estimation aims at localizing 2D human anatomical keypoints (\eg, elbow, wrist, \etc) in a given image. Current human pose estimation methods can be categorized as \textit{top-down} or \textit{bottom-up} methods. Top-down methods~\cite{chen2018cascaded, he2017mask,papandreou2017towards, sun2019deep,sun2017compositional,wang2020deep, xiao2018simple} take as input an image region within a bounding box, generally the output of a human detector, and reduce the problem to the simpler task of \emph{single human pose estimation}. Bottom-up methods \cite{cao2016realtime, kreiss2019pifpaf, newell2017associative,
papandreou2018personlab}, in contrast, start by independently localizing keypoints
in the entire image, followed by grouping them into 2D human pose instances.

The single human assumption made by top-down approaches limits the inference to a \emph{single} configuration of human joints (a single instance) that can best explain the input. Top-down pose estimation approaches~\cite{chen2018cascaded,huang2017coarse, newell2016stacked, sun2019deep, xiao2018simple} are currently the best performers on datasets such as COCO~\cite{lin2014microsoft}, MPII~\cite{andriluka14cvpr}. However, when presented with inputs containing multiple humans like crowded or occluded instances, top-down methods are forced to select a single plausible configuration per human detection. In such cases, top-down methods may erroneously identify pose landmarks corresponding to the occluder (person in the front). See, for example, Fig.~\ref{fig:introduction} (Middle). Therefore, on datasets such as CrowdPose~\cite{li2019crowdpose} and OCHuman~\cite{zhang2019pose2seg}, which have a relatively higher proportion of occluded instances (Table \ref{tab:intro_performance}), the performance of top-down methods suffer due to the single person assumption~\cite{cheng2019higherhrnet,li2019crowdpose, zhang2019pose2seg}. 

\begin{figure}
\captionsetup{font=small}
\centering
\includegraphics[width=0.9\linewidth]{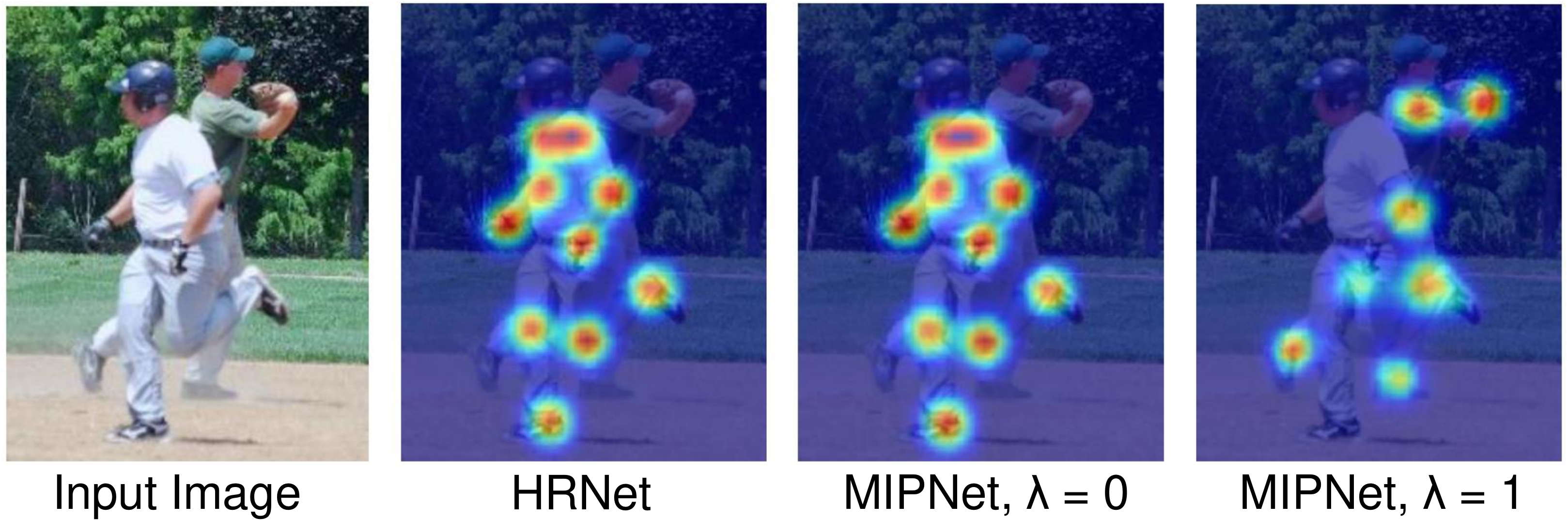}
 \vspace*{-0.1in}
\caption{Heatmap predictions for a few keypoints from HRNet vs MIPNet. HRNet only focuses on the foreground person. MIPNet enables prediction of the multiple instances from the \textit{same} input bounding box by varying $\lambda$ during inference.}
\vspace*{-0.2in}
\label{fig:heatmaps}
\end{figure}

In this paper, we rethink the architecture for top-down 2D pose estimators by predicting \emph{multiple} pose
instances for the input bounding box. The key idea of our proposed architecture is to allow the model to predict more than one pose instance for each bounding box. We demonstrate that this conceptual change improves the performance of top-down methods, especially for images with crowding and heavy occlusion. A na\"ive approach to predict multiple instances per bounding box would be to add multiple prediction heads to an
existing top-down network with a shared feature-extraction backbone. However, such an approach fails to learn different features corresponding to the various instances. A brute-force approach would then be to replicate the feature-extraction backbone, though at a cost of an $N$-fold increase in parameters, for $N$ instances. In contrast, our approach enables predicting multiple instances for any existing top-down architecture with a small increase in the number of parameters (\textit{$<3\%$}) and inference time ($<9\text{ms}, 16\%$). Technically, our approach can handle $N>2$ instances. However, as shown in Figure~\ref{fig:dataset_statistics}, number of examples with 3+ annotated pose instances per ground truth bounding box in existing datasets is extremely small. Thus, similar to ~\cite{qiu2020peeking, zhang2019pose2seg}, we primarily focus on the dominant occlusion scenario involving two persons.

\begin{figure*}[t!]
\captionsetup{font=small}
 \begin{center}
  \includegraphics[width=\linewidth]{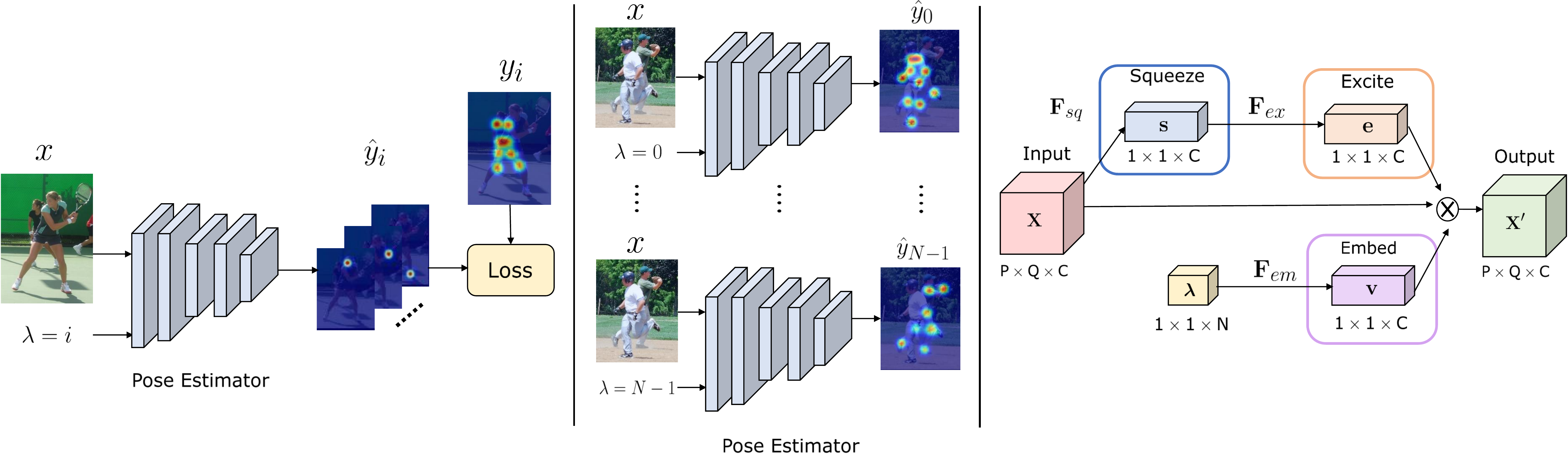}
 \end{center}
 \vspace*{-0.1in}
  \caption{(Left) MIPNet is trained to predict the $i^{th}$ instance from an input $x$ by conditioning the network using $\lambda=i$, $\forall~i = 0,\dots,N-1$. (Middle) During inference, we obtain the $N$ pose predictions by varying $\lambda$. (Right) MIMB uses squeeze, excitation and embed modules that enables $\lambda$ to modulate the feature responses for each instance.}
  \label{fig:method}
  \vspace*{-0.2in}
 \end{figure*}

To enable efficient training and inference of multiple instances in a given bounding box, we propose a novel Multi-Instance Modulation Block (MIMB). MIMB modulates the feature tensors based on a scalar \textit{instance-selector}, $\lambda$, and allows the network to index on one of the $N$ instances (Fig.~\ref{fig:heatmaps}). MIMB can be incorporated in any existing feature-extraction backbone, with a relatively simple ($<15$ lines) code change (refer supplemental). At inference, for a given bounding box, we vary the instance-selector $\lambda$ to generate multiple pose predictions (Fig.~\ref{fig:method}).

Since top-down approaches rely on the output from an object detector, they typically process a large number of bounding box hypotheses. For example, HRNet~\cite{sun2019deep} uses more than $100K$ bounding boxes from Faster R-CNN~\cite{ren2015faster} to predict 2D pose for $\sim6000$ persons in the COCO \texttt{val} dataset. Many of these bounding boxes overlap and majority have low detection scores ($<0.4$). This also adversely impacts the inference time, which increases linearly with the number of input bounding boxes. As shown in Fig.~\ref{fig:bb_decay}, using fewer, high confidence bounding boxes degrades the performance of HRNet from $37.8$ to $32.8$ AP on OCHuman, a degradation of $5$ AP in performance. In contrast, MIPNet is robust and maintains a relatively stable performance for the same inputs (drop of $1$ AP). Intuitively, our method can predict the 2D pose instance corresponding to a mis-detected bounding box based on predictions from its neighbors.

Overall, MIPNet outperforms top-down methods and occlusion specific methods on various datasets as shown in Table~\ref{tab:intro_performance}. For challenging datasets such as CrowdPose and OCHuman, containing a larger proportion of cluttered scenes (with multiple overlapping people), MIPNet sets a new state-of-the-art achieving $70.0$ AP and $42.5$ AP respectively on the \texttt{test} set outperforming bottom-up methods. Our main contributions are
\begin{itemize}
    \item We advance top-down 2D pose estimation methods by addressing limitations caused by the single person assumption during training and inference. Our approach achieves the state-of-the-art results on CrowdPose and OCHuman datasets. 
    \item MIPNet allows predicting multiple pose instances for a given bounding box efficiently by modulating feature responses for each instance independently.
    \item The ability to predict multiple instances makes MIPNet resilient to bounding box confidence and allows it to deal with missing bounding boxes with minimal impact on performance. 
\end{itemize}


\begin{table}[t]
\captionsetup{font=small}
\small
\begin{center}
\begin{tabular}{@{}l|l| c c c@{}}
\Xhline{3\arrayrulewidth}
     Dataset &IoU$>0.5$   & $\Delta AP_{0}$  & $\Delta AP_{0.9}$ & $\Delta AP_{gt}$\\
\hline
COCO & 1.2K (1\%)  &  0.0 & +1.9 & +0.7\\
CrowdPose & 2.9K (15\%)  & +0.8 & +2.3 & +0.9\\
OCHuman & 3.2K (68\%)  & +4.2 & +8.2 & +9.1\\
\Xhline{3\arrayrulewidth}
\end{tabular}
\vspace*{-0.1in}
\caption{MIPNet's relative improvement in AP compared to HRNet-W48 on the \texttt{val} set, using Faster R-CNN ($AP_{0}$: all, $AP_{0.9}$: high confidence) and ground truth ($AP_{gt}$) bounding boxes. For each dataset, the number ($\%$) of instances with occlusion IoU~$>0.5$ is reported~\cite{qiu2020peeking}. Datasets with more occlusions and crowding demonstrate higher gains.}
\vspace*{-0.2in}
\label{tab:intro_performance}
\end{center}
\end{table}

\vspace*{-0.2cm}
\section{Related Work}
\label{sec:related_work}

\textbf{Biased benchmarks:} Most human pose estimation benchmarks \cite{andriluka2018posetrack,andriluka14cvpr, guler2018densepose,   joo2015panoptic,lin2014microsoft} do not uniformly represent possible poses and occlusions in the real world. Popular datasets such as COCO \cite{lin2014microsoft} and MPII \cite{andriluka14cvpr} have less than $3\%$ annotations with crowding at IoU of 0.3 \cite{qiu2020peeking}. More than $86\%$ of annotations in COCO \cite{lin2014microsoft} have 5 or more keypoints visible \cite{ronchisupplementary}. These biases have seeped into our state-of-the-art data driven deep learning models \cite{yamada2012no}, not only in the form of poor generalization to ``in-the-tail'' data but surprisingly in critical design decisions for network architectures. Recently, challenging datasets such as OCHuman \cite{zhang2019pose2seg} and CrowdPose \cite{li2019crowdpose} containing heavy occlusion have been proposed to capture these biases. These datasets demonstrate the failures of the state-of-art models under severe occlusions (Section \ref{sec:experiments_ochuman}). MIPNet shows a significant improvement in performance under such challenging conditions.

\textbf{Top-down methods:} Top-down methods \cite{chen2018cascaded,fang2017rmpe,he2017mask, huang2017coarse,   newell2016stacked, papandreou2017towards, sun2019deep, xiao2018simple} detect the keypoints of a single person within a bounding box. These bounding boxes are usually generated by an object detector \cite{cheng2018revisiting, lin2017feature, liu2016ssd,  redmon2018yolov3,ren2015faster}. As top-down methods can normalize all the persons to approximately the same scale by cropping and resizing the bounding boxes, they are generally less sensitive to scale variations in images. Thus, state-of-the-art performances on various human pose estimation benchmarks are mostly achieved by top-down methods \cite{sun2019deep} in contrast to bottom-up methods~\cite{cao2016realtime,cheng2019higherhrnet, insafutdinov2016deepercut, iqbal2016multi,  newell2017associative,pishchulin2015deepcut,zhang2019pose2seg}. However, these methods inherently assume a single person (instance) in the detection window and often fail under occlusions in multi-person cases. It is the ambiguity of the implicit bounding-box level representation that leads to this failure. MIPNet resolves this issue by predicting multiple instances within a single bounding box.

\textbf{Occluded pose estimation:} Many methods~\cite{tang2014detection, ouyang2013single, stewart2016end, chu2020detection} have made good progress in occluded person detection. Recent methods~\cite{li2019crowdpose, zhang2019pose2seg, qiu2020peeking, jin2020differentiable} have also focused on occluded pose estimation.~\cite{li2019crowdpose} uses a top-down model to make a multi-peak prediction and joint peaks are then grouped into persons using a graph model.~\cite{zhang2019pose2seg} uses instance segmentation for occlusion reasoning.~\cite{qiu2020peeking} use a graph neural network to refine pose proposals from a top-down model. ~\cite{jin2020differentiable} is a bottom-up method which uses a differentiable hierarchical graph grouping for joint association. In contrast, our approach is much simpler and does not require initial pose estimates, grouping or solving for joint association.

Lastly, in machine learning, many models have been trained to behave differently depending on a conditional input~\cite{caruana1997multitask, dosovitskiy2019you, kokkinos2017ubernet, maninis2019attentive, mirza2014conditional, zamir2018taskonomy}. Instead of training multiple models, our approach enables training a single network for predicting multiple outputs on the same input. Rather than duplicating the feature backbone, our novel MIMB block leads to a parameter efficient design. Our multi-instance pose network is fully supervised and not related to multiple instance learning~\cite{carbonneau2018multiple, yang2005review}, which is a form of weakly-supervised learning paradigm where training instances are arranged in sets. 

\vspace*{-0.2cm}
\section{Method}
\label{sec:method}
Human pose estimation aims to detect the locations of $K$ keypoints from an input image $x \in \mathbb{R}^{H \times W \times 3}$. Most top-down methods transform this problem to estimating $K$ heatmaps, where each heatmap indicates the probability of the corresponding keypoint at any spatial location. Similar to~\cite{chen2018cascaded,newell2016stacked, xiao2018simple} we define a convolutional pose estimator, $P$,  for human keypoint detection. The bounding box at training and inference is scaled to $H \times W$ and is provided as an input to $P$. Let $y \in \mathbb{R}^{H^\prime \times W^\prime \times K}$ denote the $K$ heatmaps corresponding to the ground truth keypoints for a given input $x$. The pose estimator transforms input $x$ to a single set of predicted heatmaps, $\hat{y} \in \mathbb{R}^{H^\prime \times W^\prime \times K}$, such that $\hat{y} = P(x)$. $P$ is trained to minimize the mean squared loss $\mathcal{L}= \mathtt{MSE}(y, \hat{y})$.

\subsection{Training Multi-Instance Pose Network}
\label{sec:method_multihypo}
We propose to modify the top-down pose estimator $P$ to predict multiple instances as follows. Our pose estimator $P$ predicts $N$ instances, $\hat{y}_0, \dots, \hat{y}_{N-1}$ for an input $x$. This is achieved by conditioning the network $P$ on a scalar \textit{instance-selector} $\lambda$, $0\leq \lambda \leq N-1$. $P$ accepts both $x$ and $\lambda$ as input and predicts $\hat{y}_i = P(x, \lambda=i)$, where $i \in \{0,1,\ldots,N-1\}$.

Let $B_0$ denote the ground truth bounding box used to crop the input $x$. Let $B_i$, $i\in\{1,..n-1\}$, denote additional $n-1$ ground truth bounding boxes which overlap $B_0$, such that at least $k=3$ keypoints from $B_i$ fall within $B_0$. Thus, $B_0,\ldots,B_{n-1}$ represents the bounding boxes for $n$ ground truth pose instances present in $x$. We denote the ground truth heatmaps corresponding to these $n$ instances by $y_0, \dots y_{n-1}$.

To define a loss, we need to assign the predicted pose instances to the ground truth heatmaps. The primary instance $\hat{y}_0 = P(x, \lambda=0)$ is assigned to $y_0$, the pose instance corresponding to $B_0$. The next $N-1$ instances are assigned to the remaining ground truth heatmaps ordered according to the distance of their corresponding bounding box from $B_0$. We train the network $P$ to minimize the loss $\mathcal{L} = \frac{1}{N}\sum_{i=0}^{N-1}  \mathcal{L}_i$, where,
\begin{eqnarray}
\mathcal{L}_i =
\begin{cases}
\mathtt{MSE}(y_i, P(x, {\lambda=i})),& \forall ~0 \le i < \text{min}(n, N),\\
\mathtt{MSE}(y_{0}, P(x, {\lambda=i})),& \forall ~\text{min}(n, N) \le i < N.
\end{cases}
\label{eqn:main_loss}
\end{eqnarray}

When $n<=N$, the available $n$ ground truth pose instances are used to compute the loss for $n$ predictions, and the loss for \textit{residual} $N-n$ instances is computed using $y_0$. For example, when $n=1$ and $N=2$, both the predictions are encouraged to predict the heatmaps corresponding to the single ground truth instance present in $x$. In contrast, when $n>N$, only $N$ ground truth pose instances (closest to $B_0$) are used to compute the loss.

In our experience, employing other heuristics such as not propagating the loss, \textit{i.e., don't care} for residual instances resulted in less stable training. Additionally, a \textit{don't care} based training scheme for residual instances resulted in significantly higher false positives, especially as we do not know the number of valid person instances per input at run-time. During inference, we vary $\lambda$ to extract different pose predictions from the same input $x$ as shown in Fig. \ref{fig:method}.

\subsection{Multi-Instance Modulation Block} 
\label{sec:method_sea}
In this section, we describe the Multi-Instance Modulation Block (MIMB) that can be easily introduced in any existing feature extraction backbone. The MIMBs allow a top-down pose estimator $P$ to predict multiple instances from an input image $x$. Using MIMBs, $P$ can now accept both $x$ and the instance-selector $\lambda$ as inputs. The design of MIMB is inspired by the squeeze excite block of~\cite{hu2018squeeze}. Let $\mathbf{X} \in \mathbb{R}^{P \times Q \times C}$ be an intermediate feature map with $C$ channels, such that $\mathbf{X}=[\mathbf{x}_1,\mathbf{x}_2,\dots,\mathbf{x}_{C}]$. We use an instance-selector $\lambda$ to modulate the channel-wise activations of the output of the excite module as shown in Fig.~\ref{fig:method} (Right). The key insight of our design is that we can use the same set of convolutional filters to dynamically cater to different instances in the input. Compared to a brute force approach of replicating the feature backbone or assigning a fixed number of channels per instance, our design is parameter efficient.

\begin{table*}[t]
\captionsetup{font=small}
    \centering
    \small
    \renewcommand{\arraystretch}{1.0} 
    \begin{tabular}{@{}l|l|l|c|l|l c c c c c c@{}}
    \hline
Method  & Arch   & \#Params  & $\text{AP}$ & $\text{AP}^{50}$ & $\text{AP}^{75}$ & $\text{AP}^\text{M}$ & $\text{AP}^\text{L}$ & $\text{AR}$  \\
    \hline
      SBL$\dagger$  & R-50    &  34.0M & 
      72.4 & 91.5 & 80.4 & 69.7 & 76.5 & 75.6\\
  MIPNet$\dagger$ & R-50    &    35.0M (+2.8\%) & 
      \textbf{73.3 (+0.9)} & \textbf{93.3} & \textbf{81.2} & \textbf{70.6} & \textbf{77.6} & \textbf{76.7}\\
      \hline
      SBL$\dagger$  & R-101    &   53.0M & 
      73.4 & 92.6 & 81.4 & 70.7 & 77.7 & 76.5\\
  MIPNet$\dagger$ & R-101     & 54.0M (+1.7\%) & 
      \textbf{74.1 (+0.7)} & \textbf{93.3} & \textbf{82.3} & \textbf{71.3} & \textbf{78.6} & \textbf{77.4}\\
      \hline
      SBL$\dagger$  & R-152    &   68.6M & 
      74.3 & 92.6 & 82.5 & 71.6 & 78.7 & 77.4\\
  MIPNet$\dagger$ & R-152  & 69.6M (+1.4\%) & 
      \textbf{74.8 (+0.5)} & \textbf{93.3} & \textbf{82.4} & \textbf{71.7} & \textbf{79.4} & \textbf{78.2}\\
      
    \hline
      
      SBL$\star$  & R-50   &  34.0M & 
      74.1 & 92.6 & 80.5 & 70.5 & 79.6 & 76.9\\
  MIPNet$\star$ & R-50    &   35.0M (+0.4\%) & 
      \textbf{75.3 (+1.2)} & \textbf{93.4} & \textbf{82.4} & \textbf{72.0} & \textbf{80.4} & \textbf{78.4}\\
      \hline
      SBL$\star$ & R-101    &   35.0M & 
      75.5 & 92.5 & 82.6 & 72.4 & 80.8 & 78.4\\
  MIPNet$\star$& R-101     &   54.0M (+0.3\%) & 
      \textbf{76.0 (+0.5)} & \textbf{93.4} & \textbf{83.5} & \textbf{72.6} & \textbf{81.1} & \textbf{79.1}\\
      \hline
      SBL$\star$  & R-152    &    68.6M & 
      76.6 & 92.6 & 83.6 & 73.7 & 81.3 & 79.3\\
  MIPNet$\star$ & R-152 &  69.6M (+2.8\%) & 
      \textbf{77.0 (+0.4)} & \textbf{93.5} & \textbf{84.3} & \textbf{73.7} & \textbf{81.9} & \textbf{80.0}\\
      
    \hline
    \hline
    
      HRNet$\dagger$  & H-32   & 28.5M  & 
      76.5 & 93.5 & 83.7 & 73.9 & 80.8 & 79.3\\
  MIPNet$\dagger$ & H-32 &  28.6M (+1.7\%) & 
      \textbf{77.6 (+1.1)} & \textbf{94.4} & \textbf{85.3} & \textbf{74.7} & \textbf{81.9} & \textbf{80.6}\\
      \hline
      HRNet$\dagger$  & H-48    &  63.6M  & 
      77.1 & 93.6 & 84.7 & 74.1 & 81.9 & 79.9\\
  MIPNet$\dagger$ & H-48     &    63.7M (+1.4\%) & 
      \textbf{77.6 (+0.5)} & \textbf{94.4} & \textbf{85.4} & \textbf{74.6} & \textbf{82.1} & \textbf{80.6}\\
    
     \hline
     
      HRNet$\star$  & H-32     &  28.5M  & 
      77.7 & 93.6 & 84.7 & 74.8 & 82.5 & 80.4\\
  MIPNet$\star$ & H-32     &  28.6M (+0.4\%) &
      \textbf{78.5 (+0.8)} & \textbf{94.4} & \textbf{85.7} & \textbf{75.6} & \textbf{83.0} & \textbf{81.4}\\
      \hline
      HRNet$\star$  & H-48    &  63.6M  & 
      78.1 & 93.6 & 84.9 & 75.3 & 83.1 & 80.9\\
  MIPNet$\star$ & H-48    &  63.7M (+0.3\%) & 
      \textbf{78.8 (+0.7)} & \textbf{94.4} & \textbf{85.7} & \textbf{75.5} & \textbf{83.7} & \textbf{81.6}\\
\hline
  \end{tabular}
    \vspace*{-0.1in}
    \caption{MIPNet improves performance on COCO~\texttt{val} set across various architectures and input sizes (using ground-truth bounding boxes). R-@ and H-@ stands for ResNet-@ and HRNet-W@ respectively. $\dagger$ and $\star$ denotes input resolution of $256\times192$ and $384 \times 288$ respectively. SBL refers to SimpleBaseline~\cite{xiao2018simple}. \#Params are only of the pose estimation network, excluding bounding box computation.}
    \label{tab:coco_quantitative}
    \vspace*{-0.2in}
\end{table*}


Let $\mathbf{F}_{sq}$, $\mathbf{F}_{ex}$, $\mathbf{F}_{em}$ denote the \textit{squeeze}, \textit{excite}, and \textit{embed} operations, respectively, within MIMB. We represent $\boldsymbol{\lambda}$ as the one hot representation of scalar $\lambda$. The feature map $\mathbf{X}$ is transformed to $\mathbf{X^\prime} = [\mathbf{x}^\prime_1, \mathbf{x}^\prime_2, \dots, \mathbf{x}^\prime_C]$ as follows,
\begin{eqnarray}
s_c & = &\mathbf{F}_{sq}(\mathbf{x}_c), \\
\mathbf{e} & = &\mathbf{F}_{ex}(\mathbf{s}), \\
\mathbf{v} & = &\mathbf{F}_{em}(\boldsymbol{\lambda}), \\
\mathbf{x}^\prime_c & = & (v_c \times e_c) \mathbf{x}_c,
\end{eqnarray}
s.t. $\mathbf{s} = [s_1, \dots, s_C]$, $\mathbf{v} = [v_1, \dots, v_C]$ and $\mathbf{e} = [e_1, \dots, e_C]$. $\mathbf{F}_{sq}$ \textit{squeezes} the global spatial information into a channel descriptor using global average pooling. $\mathbf{F}_{ex}$ allows modeling for channel-wise interactions on the output of $\mathbf{F}_{sq}$. $\mathbf{F}_{ex}$ is implemented as a two layer, fully-connected, neural network. Following the output of the excite module, we modulate the channel-wise activations using the embedding of $\boldsymbol{\lambda}$ from another simple neural network $\mathbf{F}_{em}$. $\mathbf{F}_{em}$ has a similar design to $\mathbf{F}_{ex}$. 

During inference, we vary the instance-selector $\lambda$ from $0$ to $N-1$ to get $N$ predictions and then apply OKS-NMS~\cite{sun2019deep} after merging all predictions. Please refer supplemental for details. Figure~\ref{fig:heatmaps} visualizes the predicted heatmaps from HRNet and MIPNet (using $N=2$). Note that HRNet only outputs the heatmap corresponding to the foreground person while MIPNet predicts heatmaps for both persons using different values of $\lambda$ at inference.

\vspace*{-0.2cm}
\section{Experiments}
\label{sec:experiments}
 \begin{figure} 
 \captionsetup{font=small}
     \centering
     \includegraphics[width=0.25\textwidth]{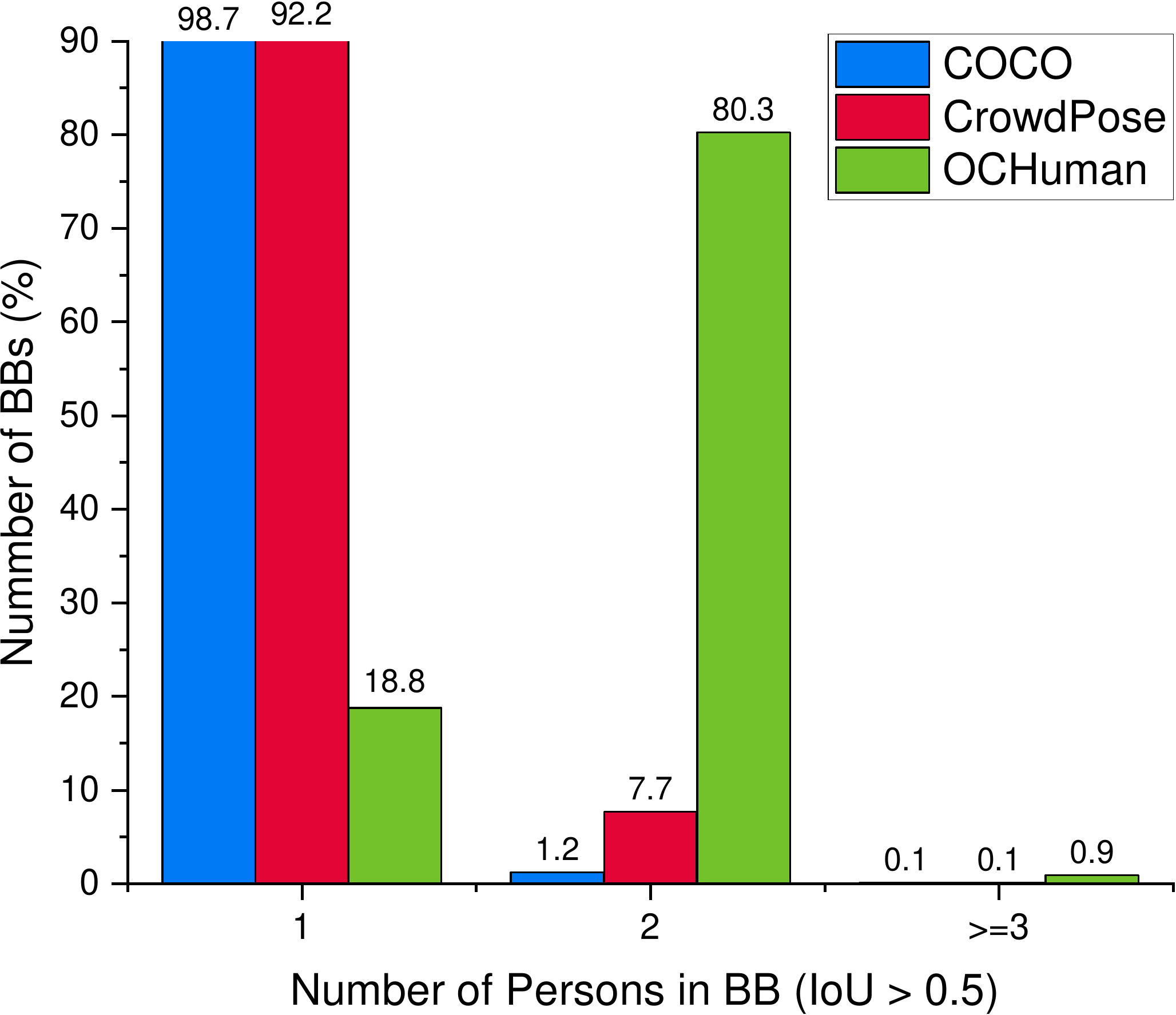}
      \vspace*{-0.1in}
     \caption{Percentage of examples with $1$, $2$ and $3+$ pose instances per ground truth bounding box in various datasets.}
     \vspace*{-0.2in}
     \label{fig:dataset_statistics}
 \end{figure}

We evaluate MIPNet on three datasets: \textit{Common-Objects in Context}-COCO~\cite{lin2014microsoft}, \textit{CrowdPose}~\cite{li2019crowdpose} and \textit{Occluded Humans}-OCHuman~\cite{zhang2019pose2seg}. These datasets represent varying degrees of occlusion/crowding (see Table~\ref{tab:intro_performance}) and help illustrate the benefits of predicting multiple instances in top-down methods. We report standard metrics such as $\text{AP}, \text{AP}^{50}, \text{AP}^{75}, \text{AP}^\text{M}, \text{AP}^\text{L}, \text{AR}, \text{AP}^{easy}, \text{AP}^{med}$ and $\text{AP}^{hard}$  at various Object Keypoint Similarity as defined in~\cite{lin2014microsoft,li2019crowdpose}. We report results using ground truth bounding boxes as well as bounding boxes obtained via YOLO~\cite{redmon2018yolov3} and Faster R-CNN~\cite{ren2015faster} detectors.

We base MIPNet on recent state-of-the-art top-down architectures, namely, SimpleBaseline~\cite{xiao2018simple} and HRNet~\cite{sun2019deep}. When comparing with HRNet, MIPNet employs a similar feature extraction backbone and adds MIMBs' at the output of the convolutional blocks at the end of stages $3$ and $4$~\cite{sun2019deep}. For comparisons with SimpleBaseline~\cite{xiao2018simple}, two MIMB's are added to the last two ResNet blocks in the encoder.


\textbf{Number of instances $N$:} Trivially, $N=1$ is equivalent to baseline top-down methods. By design, MIPNet supports predicting multiple instances. Empirically, on average we observed a small improvement of $0.3$ AP, $0.5$ AP using $N=3$ and $N=4$ on top of $N=2$ respectively on the datasets. This is consistent with the fact that most datasets have very few examples with three or more ground-truth pose instances per bounding box (Fig.~\ref{fig:dataset_statistics}). However, $N=2$ provides a substantial improvement over $N=1$ baseline as shown in our experiments. Note that since the MIMBs are added to the last few stages in our experiments, the increase in inference time due to predicting $N=2$ instances is small (Table \ref{tab:crowdpose_quantitative}). For bigger HRNet-48 network with input resolution of $384\times288$, inference time increases by $8.2  \text{ms}$ ($16.7\%$). For smaller HRNet-32 network, increase in run-time is $4.7 \text{ms}$ ($11.9\%$). This is significantly better than replicating the backbone for each instance, which would lead to a 2x increase in inference time for $N=2$. Please refer supplemental for more details. 

\subsection{COCO Dataset}
\textbf{Dataset:} COCO contains $64K$ images and $270K$ persons labeled with 17 keypoints. For training we use the \texttt{train} set ($57K$ images, $150K$ persons) and for evaluation we use the \texttt{val} ($5K$ images, $6.3K$ persons) and the \texttt{test-dev} set ($20K$ images). The input bounding box is extended in either height or width to obtain a fixed aspect ratio of $4:3$. The detection box is then cropped from the image and is resized to a fixed size of either $256\times192$ or $384\times288$, depending on the experiment. Following~\cite{newell2017associative}, we use data augmentation with random rotation ($[{-45}^{\circ}, {45}^{\circ}]$), random scale ($[0.65, 1.35]$), flipping, and half-body crops. Following~\cite{newell2016stacked,sun2019deep, xiao2018simple}, we use flipping and heatmap offset during inference.

\textbf{Results:} Table~\ref{tab:coco_quantitative} compares the performance of MIPNet with SimpleBaseline (denoted as SBL) and HRNet using ground truth bounding boxes. MIPNet outperforms the baseline across various backbones and input sizes. Using ResNet-50 backbone, MIPNet improves the SimpleBaseline results by $0.9$ AP for smaller input size and $1.2$ AP for larger input size. Comparing with HRNet, MIPNet shows an improvement ranging from $0.7$ to $1.1$ AP on various architectures and input sizes. Note that MIPNet results in $<3\%$ increase in parameters compared to the baselines. 

When using bounding boxes obtained from a person detector, as expected, MIPNet performs comparably to SBL and HRNet when using the same backbone (Table~\ref{tab:detector_performance}). Unsurprisingly, since most of the COCO bounding boxes contain a single person. The benefits of MIPNet are apparent on more challenging CrowdPose and OCHuman datasets (Sect.~\ref{sec:experiments_crowdpose},~\ref{sec:experiments_ochuman}).

    
    
    

\begin{table}
\captionsetup{font=small}
    \centering
    \small
    \setlength\tabcolsep{3pt}
    \begin{tabular}{@{}l|c|c| c c c c c c}
    \hline
    Arch & Latency & $\text{AP}$ & $\text{AP}^{50}$ & $\text{AP}^{75}$ & $\text{AP}^\text{easy}$ & $\text{AP}^\text{med}$ & $\text{AP}^\text{hard}$ \\
    
    \hline
    HRNet-32$\dagger$ & 27.5 \scriptsize{ms}  & 70.0 & 91.0 & 76.3 & \textbf{78.8} & 70.3 & 61.7 \\
    MIPNet$\dagger$ & 30.9 \scriptsize{ms} & \textbf{71.2} & \textbf{91.9} & \textbf{77.4} & \textbf{78.8} & \textbf{71.5} & \textbf{63.8} \\
    \hline
    HRNet-48$\dagger$ & 33.8 \scriptsize{ms}  &  71.3 & 91.1 & 77.5 & 80.5 & 71.4 & 62.5 \\
    MIPNet$\dagger$ & 39.6 \scriptsize{ms} & \textbf{72.8} & \textbf{92.0} & \textbf{79.2} & \textbf{80.6} & \textbf{73.1} & \textbf{65.2} \\
    
    \hline
    HRNet-32$\star$ & 39.4 \scriptsize{ms} &  71.6 & 91.1 & 77.7 & 80.4 & 72.1 & 62.6 \\
    MIPNet$\star$ & 44.1 \scriptsize{ms} & \textbf{73.0} & \textbf{91.8} & \textbf{79.3} & \textbf{80.7} & \textbf{73.3} & \textbf{65.5}\\
    \hline
    HRNet-48$\star$ & 49.1 \scriptsize{ms}  & 72.8 & \textbf{92.1} & 78.7 & \textbf{81.3} & 73.3 & 64.0 \\
    MIPNet$\star$ & 57.3 \scriptsize{ms} & \textbf{73.7} & 91.9 & \textbf{80.0} & 80.7 & \textbf{74.1} & \textbf{66.5}\\
    
    \hline
    \end{tabular}
    \vspace*{-0.1in}
    \caption{MIPNet outperforms HRNet on CrowdPose \texttt{val} set. $\dagger$ and $\star$ denote  input resolution of $256 \times 192$ and $384 \times 288$, respectively. Average GPU latency is reported with batch size 24.}
    \label{tab:crowdpose_quantitative}
    \vspace*{-0.3in}
\end{table}

\label{sec:experiments_coco}

\subsection{CrowdPose Dataset}
\textbf{Dataset:} CrowdPose contains $20K$ images and $80K$ persons labeled with 14 keypoints. CrowdPose has more crowded scenes as compared to COCO, but the index of crowding is less compared to the OCHuman~\cite{zhang2019pose2seg}. For training, we use the \texttt{train} set ($10K$ images, $35.4K$ persons) and for evaluation we use the \texttt{val} set ($2K$ images, $8K$ persons) and \texttt{test} set ($8K$ images, $29K$ persons).


\textbf{Results:} Table \ref{tab:crowdpose_quantitative} compares the performance of MIPNet with HRNet when evaluated using ground-truth bounding boxes. MIPNet outperforms HRNet with improvements in AP ranging from $0.9$ to $1.5$ across different input sizes. As shown in Table~\ref{tab:detector_performance}, when evaluated using person detector bounding boxes, MIPNet 
improves SBL by $7.3$ AP on the \texttt{test} set with an increase of less than 25 ms in inference time. For completeness, we also trained and evaluated HRNet on CrowdPose. MIPNet outperforms HRNet by $0.7$ AP on the \texttt{test} set and $0.8$ AP on the \texttt{val} set. MIPNet achieves state-of-the-art performance of $70.0$ AP comparable to the two-stage method OPECNet~\cite{qiu2020peeking} which refines initial pose estimates from AlphaPose+~\cite{qiu2020peeking}. We report additional results in the supplemental.


\begin{table}[!h]
\captionsetup{font=small}
    \small
    \renewcommand{\arraystretch}{1.0} 
    \begin{tabular}{@{}p{1.1cm}|p{0.8cm}|p{1.45cm}|p{0.3cm} p{0.3cm} p{0.3cm} p{0.3cm} p{0.3cm}@{}}
    \hline
Method  & Arch     &  $\text{AP}$ & $\text{AP}^{50}$ & $\text{AP}^{75}$ & $\text{AP}^\text{M}$ & $\text{AP}^\text{L}$ & $\text{AR}$  \\
    \hline
      SBL$\dagger$  & R-50    & 
      56.3 & 76.1 & 61.2 & 66.4 & 56.3 & 61.0\\
  MIPNet$\dagger$ & R-50     &  
      \textbf{64.4 (+8.1)} & \textbf{86.0} & \textbf{70.4} & \textbf{66.8} & \textbf{64.4} & \textbf{72.3}\\
    
    \hline
      SBL$\dagger$  & R-101    & 
      60.5 & 77.2 & 66.6 & \textbf{68.3} & 60.5 & 64.7\\
  MIPNet$\dagger$ & R-101    &   
      \textbf{68.2 (+7.7)} & \textbf{87.4} & \textbf{75.1} & 67.0 & \textbf{68.2} & \textbf{75.5}\\
      
    \hline
      SBL$\dagger$  & R-152    & 
      62.4 & 78.3 & 68.1 & \textbf{68.3} & 62.4 & 66.5\\
  MIPNet$\dagger$ & R-152    &   
      \textbf{70.3 (+7.9)} & \textbf{88.6} & \textbf{77.9} & 66.9 & \textbf{70.2} & \textbf{77.0}\\
    
     \hline
     
      SBL$\star$  & R-50    &  
      55.8 & 74.8 & 60.4 & 64.7 & 55.9 & 60.7\\
  MIPNet$\star$ & R-50    &  
      \textbf{65.3 (+9.5)} & \textbf{87.5} & \textbf{72.2} & \textbf{66.0} & \textbf{66.3} & \textbf{74.1}\\
    \hline  
      SBL$\star$  & R-101    &  
      61.6 & 77.2 & 66.6 & 62.1 & 61.6 & 65.8\\
  MIPNet$\star$ & R-101     &   
      \textbf{70.3 (+8.7)} & \textbf{88.4} & \textbf{77.1} & \textbf{64.1} & \textbf{70.4} & \textbf{77.7}\\
  \hline    
      SBL$\star$  & R-152    &  
      64.2 & 78.3 & 69.1 & 66.5 & 64.2 & 68.1\\
  MIPNet$\star$ & R-152 &  
      \textbf{72.4 (+8.2)} & \textbf{89.5} & \textbf{79.5} & \textbf{67.7} & \textbf{72.5} & \textbf{79.6}\\
      
    \hline
    
      HRNet$\dagger$  & H-32  &
      63.1 & 79.4 & 69.0 & 64.2 & 63.1 & 67.3\\
  MIPNet$\dagger$ & H-32 &  
     \textbf{72.5 (+9.4)} & \textbf{89.2} & \textbf{79.4} & \textbf{65.1} & \textbf{72.6} & \textbf{79.1}\\
 \hline     
      HRNet$\dagger$  & H-48    &  
      64.5 & 79.4 & 70.1 & 65.1 & 64.5 & 68.5\\
  MIPNet$\dagger$ & H-48     &  
    \textbf{72.2 (+7.7)} & \textbf{89.5} & \textbf{78.7} & \textbf{66.5} & \textbf{72.3} & \textbf{79.2}\\
     
     \hline
     
      HRNet$\star$  & H-32     &
      63.7 & 78.4 & 69.0 & 64.3 & 63.7 & 67.6\\
  MIPNet$\star$ & H-32     &  
      \textbf{72.7 (+9.0)} & \textbf{89.6} & \textbf{79.6} & \textbf{66.5} & \textbf{72.7} & \textbf{79.7}\\
\hline      
      HRNet$\star$ & H-48    &  
      65.0 & 78.4 & 70.3 & \textbf{68.4} & 65.0 & 68.8\\
  MIPNet$\star$ & H-48    &  
      \textbf{74.1 (+9.1)} & \textbf{89.7} & \textbf{80.1} & \textbf{68.4} & \textbf{74.1} & \textbf{81.0}\\
\hline
  \end{tabular}
    \vspace*{-0.1in}
    \caption{Comparisons on OCHuman \texttt{val} set with ground-truth bounding box evaluation after training on COCO \texttt{train} set. $\dagger$ and $\star$ denotes input resolution of $256 \times 192$ and $384 \times 288$ respectively. R-@ and H-@ stands for ResNet-@ and HRNet-W@ respectively. SBL refers to SimpleBaseline~\cite{xiao2018simple}.}
    \label{tab:ochuman_quantitative}
    \vspace*{-0.2in}
\end{table}

\label{sec:experiments_crowdpose}

\subsection{OCHuman Dataset}
\textbf{Dataset:} OCHuman is focused on heavily occluded humans. It contains $4731$ images and $8110$ persons labeled with $17$ keypoints. In OCHuman, on an average $67\%$ of the bounding box area has overlap with other bounding boxes~\cite{zhang2019pose2seg}, compared to only $0.8\%$ for COCO. Additionally, the number of examples with occlusion IoU $>0.5$ is $68\%$ for OCHuman, compared to $1\%$ for COCO (Table~\ref{tab:intro_performance}). This makes the OCHuman dataset complex and challenging for human pose estimation under occlusion. The single person assumption made by existing top-down methods is not entirely applicable to examples in this dataset.

Similar to~\cite{zhang2019pose2seg}, we use the \texttt{train} set of COCO for training. Note that we do not train on the OCHuman \texttt{train} set. For evaluation, we use the \texttt{val} set ($2,500$ images, $4,313$ persons) and the \texttt{test} set ($2,231$ images, $3,819$ persons). 

\textbf{Results:} Table~\ref{tab:ochuman_quantitative} compares the performance of MIPNet with SimpleBaseline and HRNet on OCHuman when evaluated with ground truth bounding boxes on the \texttt{val} set. MIPNet significantly outperforms SimpleBaseline with improvements in AP ranging from $7.7$ to $10.5$, across various architectures and input sizes. Similarly, for HRNet the performance gains between $7.7$ to $9.4$ AP are observed.

Current state-of-the-art results on OCHuman are reported by HGG~\cite{jin2020differentiable} (bottom-up method, multi-scale testing) as shown in Table~\ref{tab:detector_performance}. In addition, we also evaluated MIPNet using person detector boxes on OCHuman with same backbones as baselines for a fair comparison. MIPNet with ResNet101 backbone and YOLO bounding boxes outperforms OPEC-Net by $5.9$ AP on the \texttt{test} set.  When using Faster R-CNN bounding boxes, MIPNet outperforms HRNet and HGG by $5.3$ AP and $6.5$ AP, respectively, on the \texttt{test} set. The improvements are significant and to the best of our knowledge, this is the first time a top-down method has outperformed the state-of-the-art bottom-up method using multi-scale testing on OCHuman. 

\begin{figure*}
 \captionsetup{font=small}
 \begin{center}
\includegraphics[height=0.14\textheight,width=0.3\linewidth]{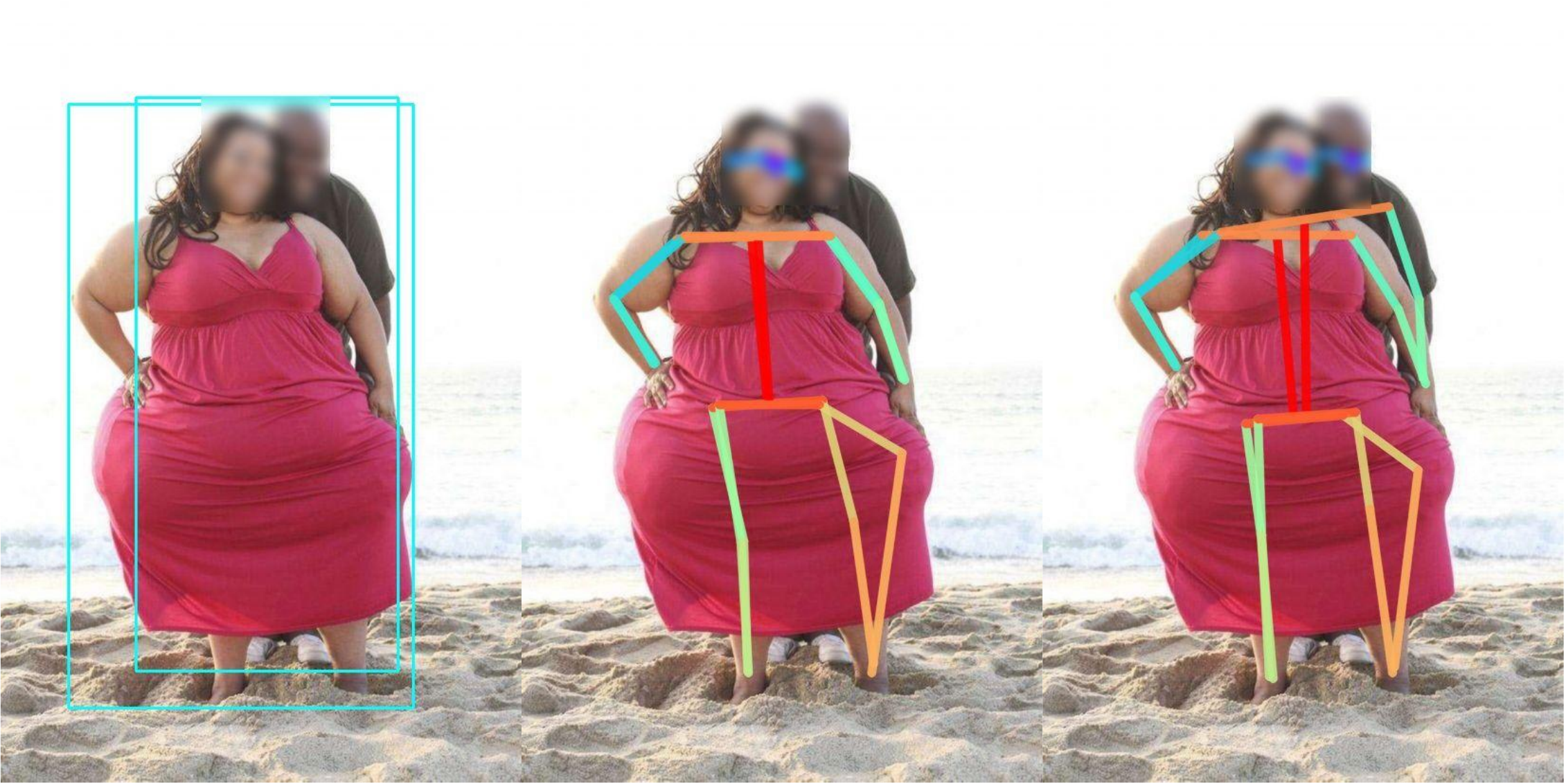}
\includegraphics[height=0.14\textheight,width=0.3\linewidth]{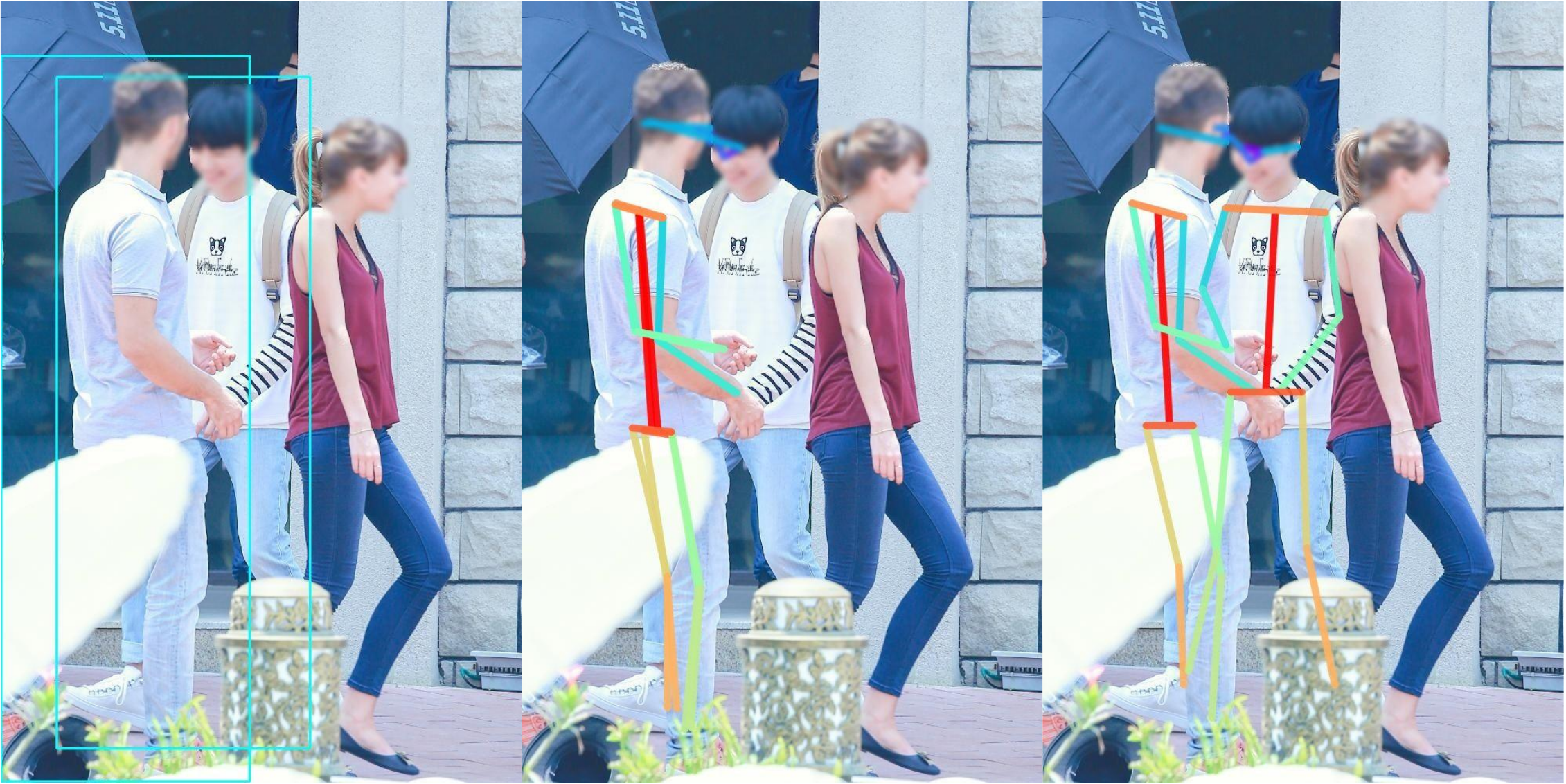}
\includegraphics[height=0.14\textheight,width=0.3\linewidth]{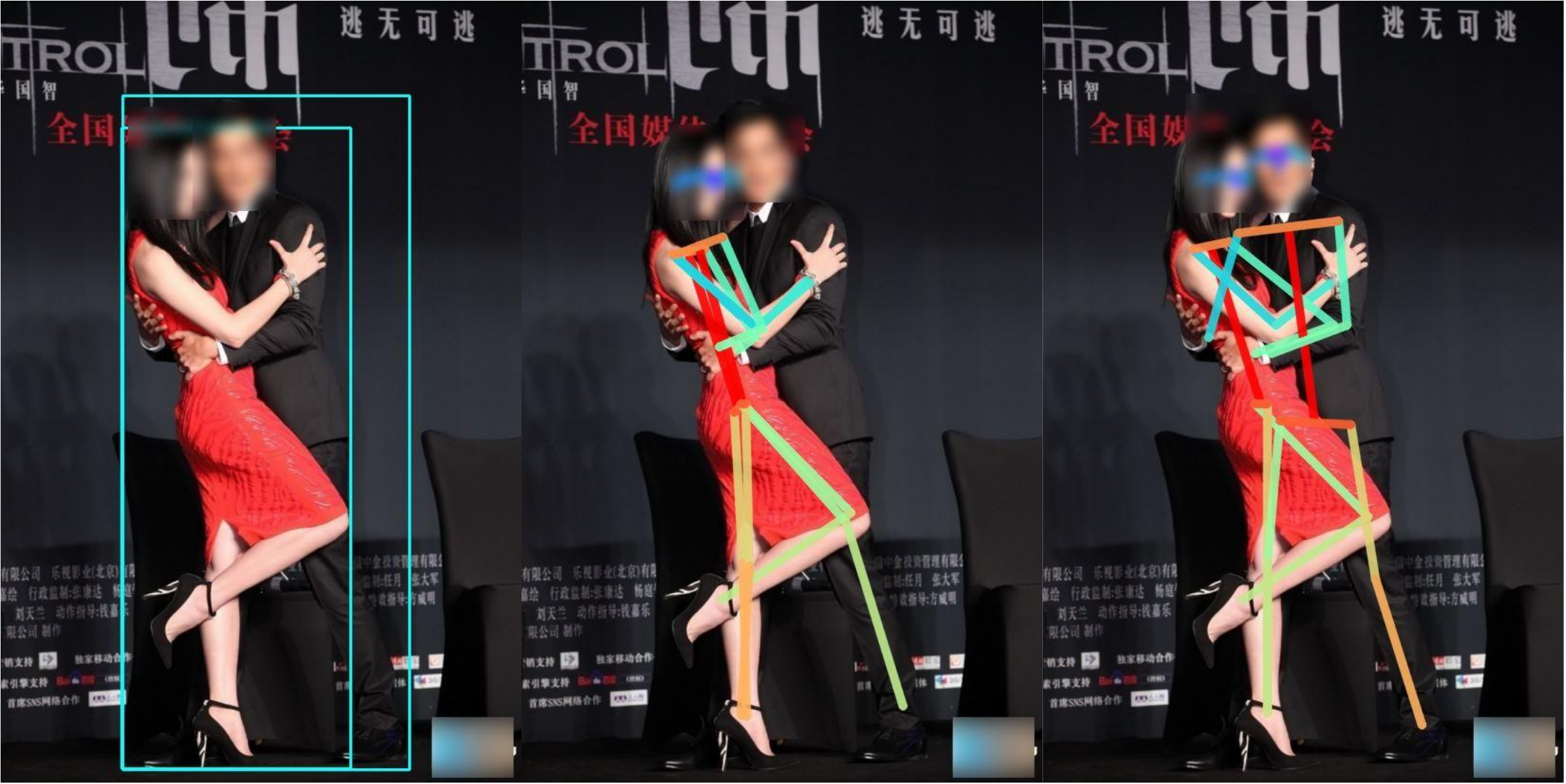}

\includegraphics[height=0.14\textheight,width=0.3\linewidth]{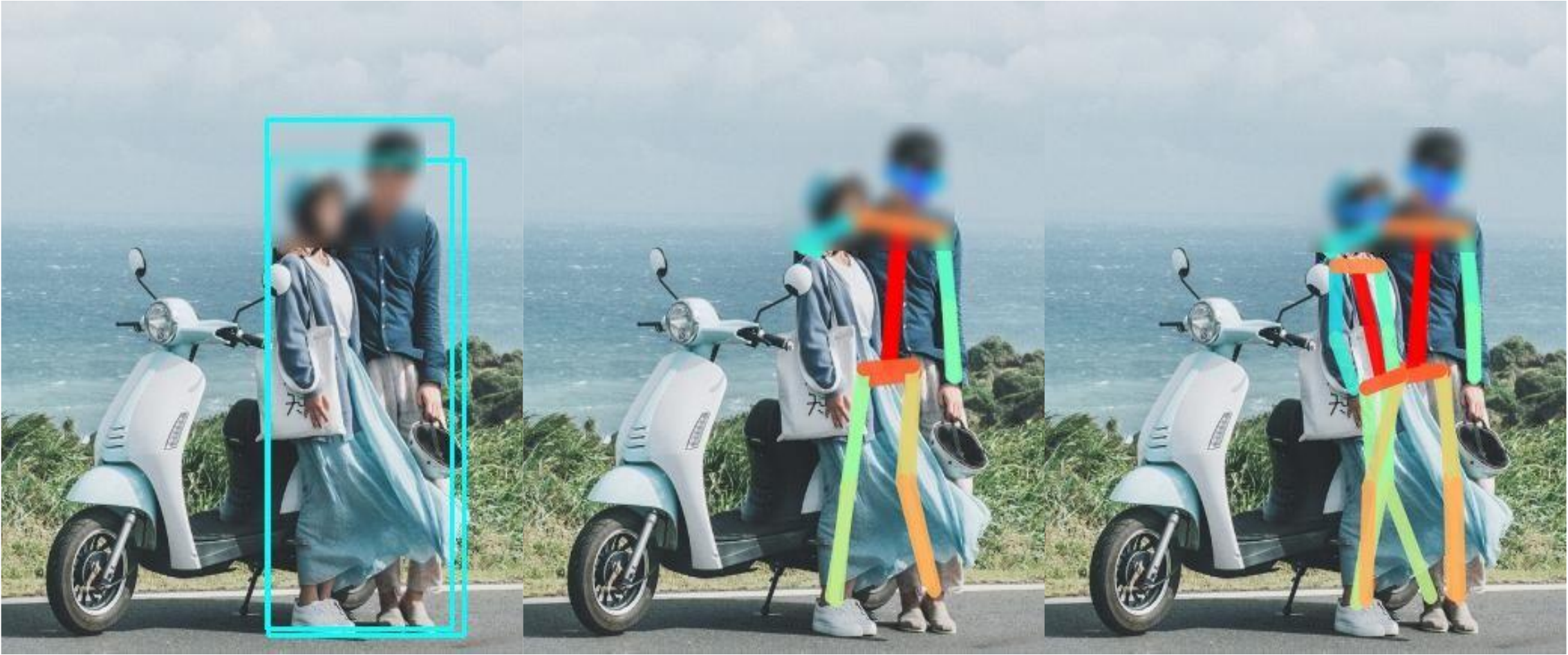}
\includegraphics[height=0.14\textheight,width=0.3\linewidth]{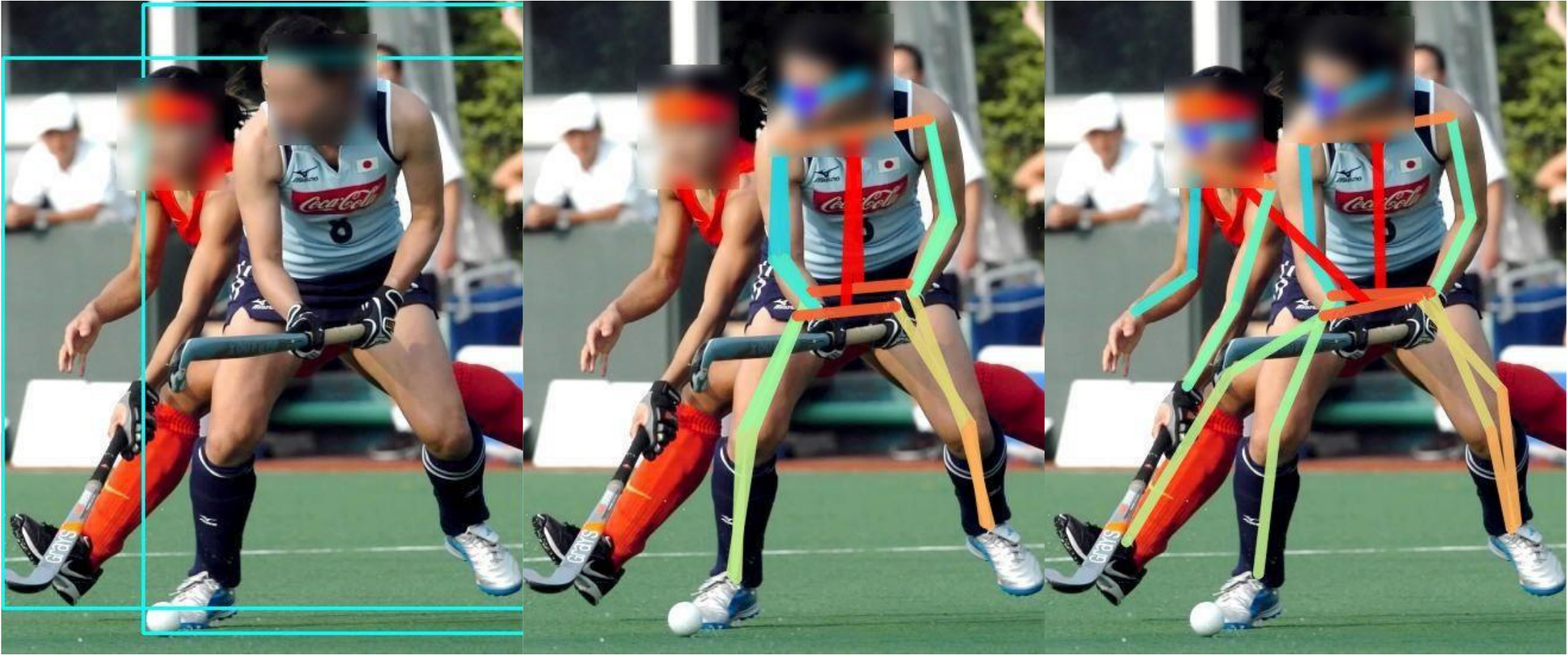}
\includegraphics[height=0.14\textheight,width=0.3\linewidth]{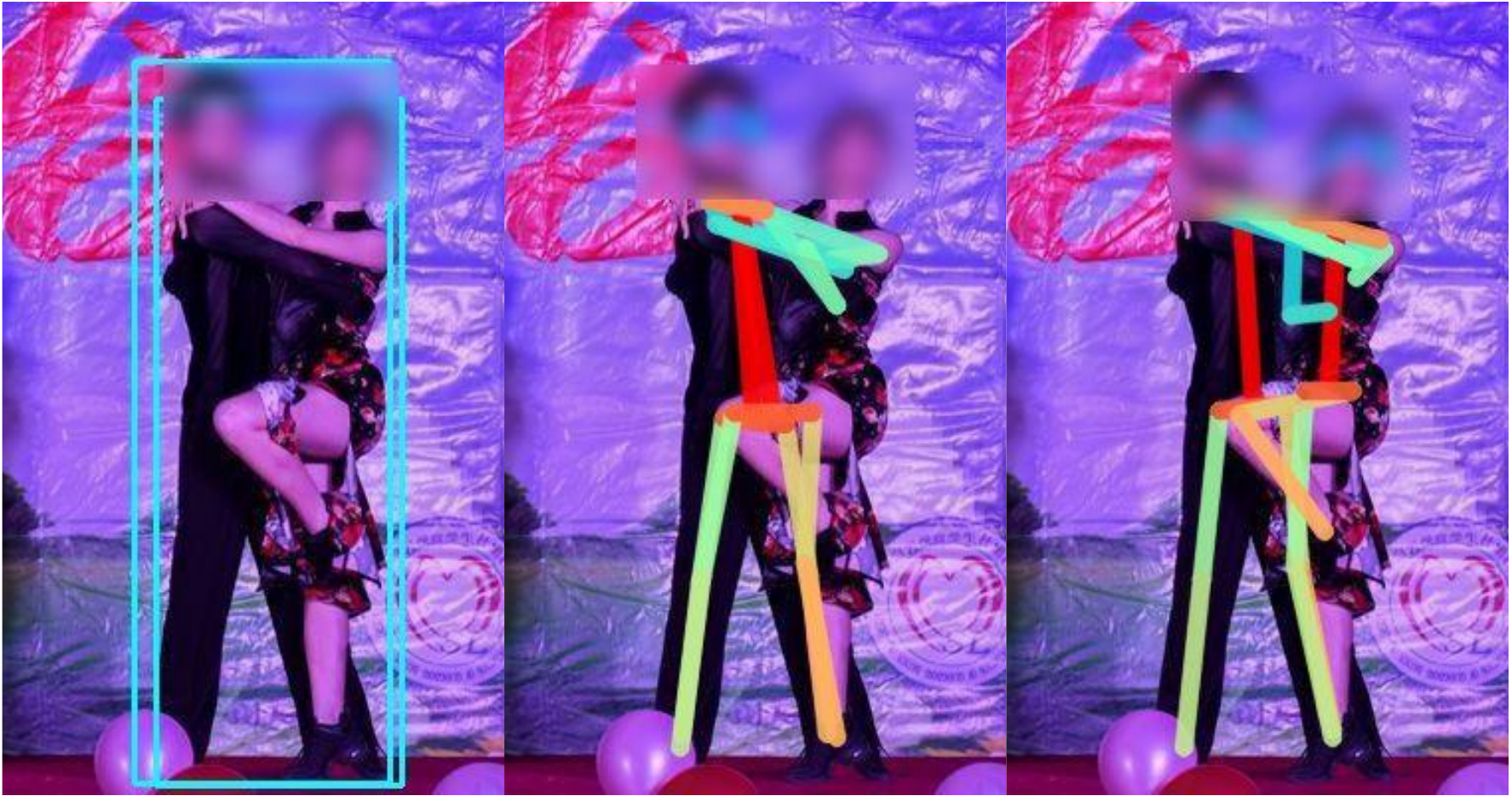}

\includegraphics[height=0.14\textheight,width=0.3\linewidth]{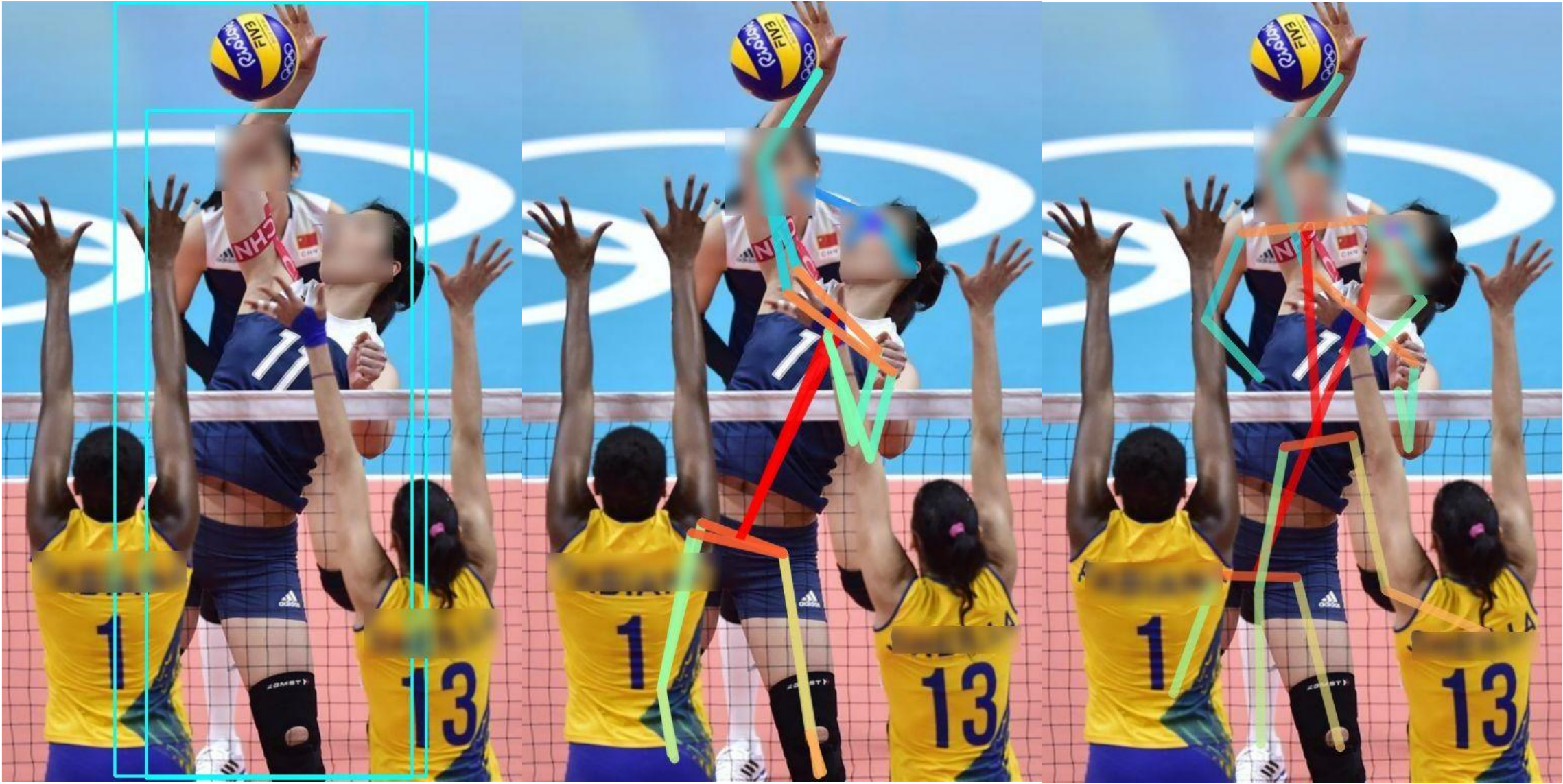}
\includegraphics[height=0.14\textheight,width=0.3\linewidth]{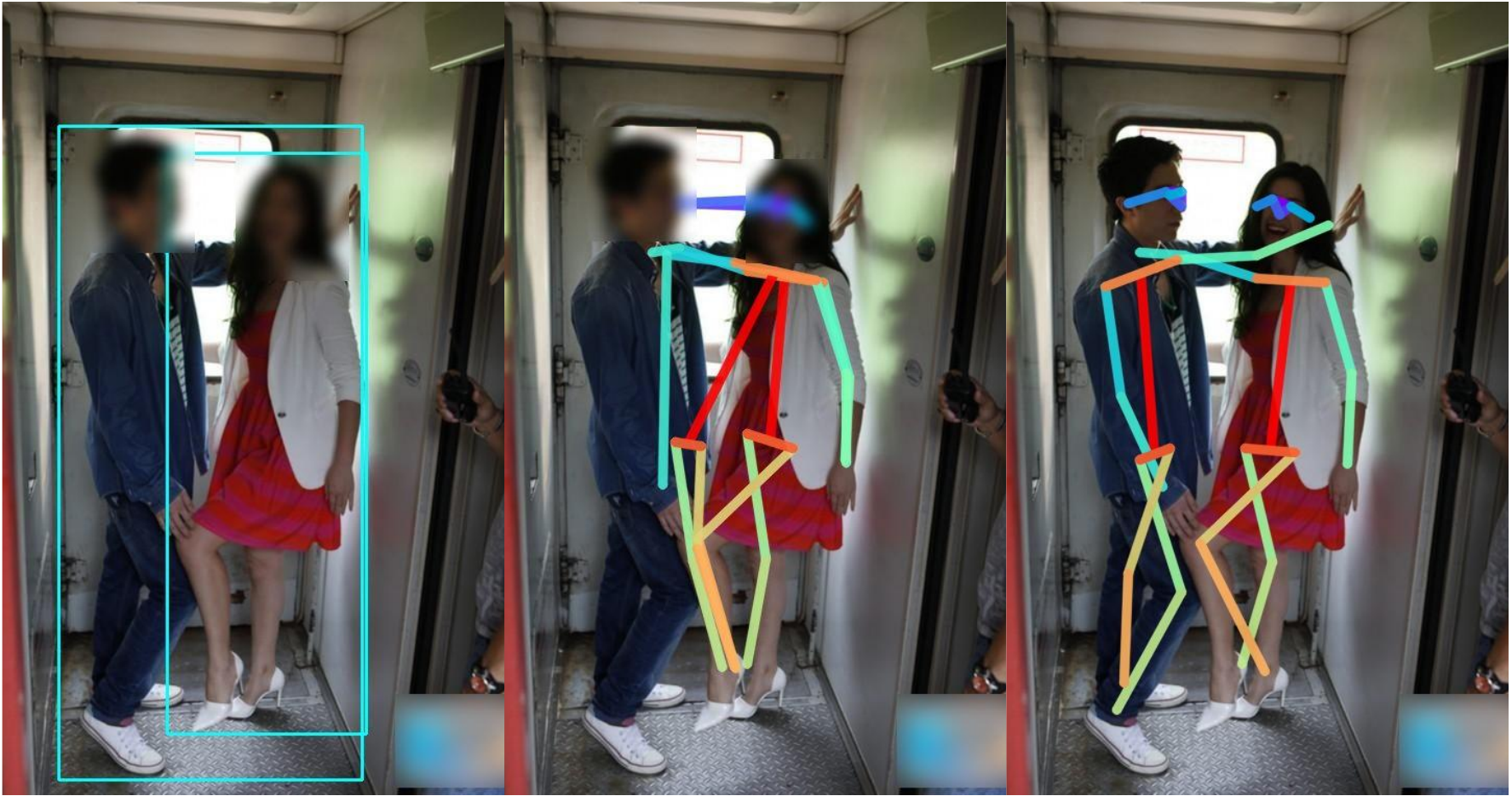}
\includegraphics[height=0.14\textheight,width=0.3\linewidth]{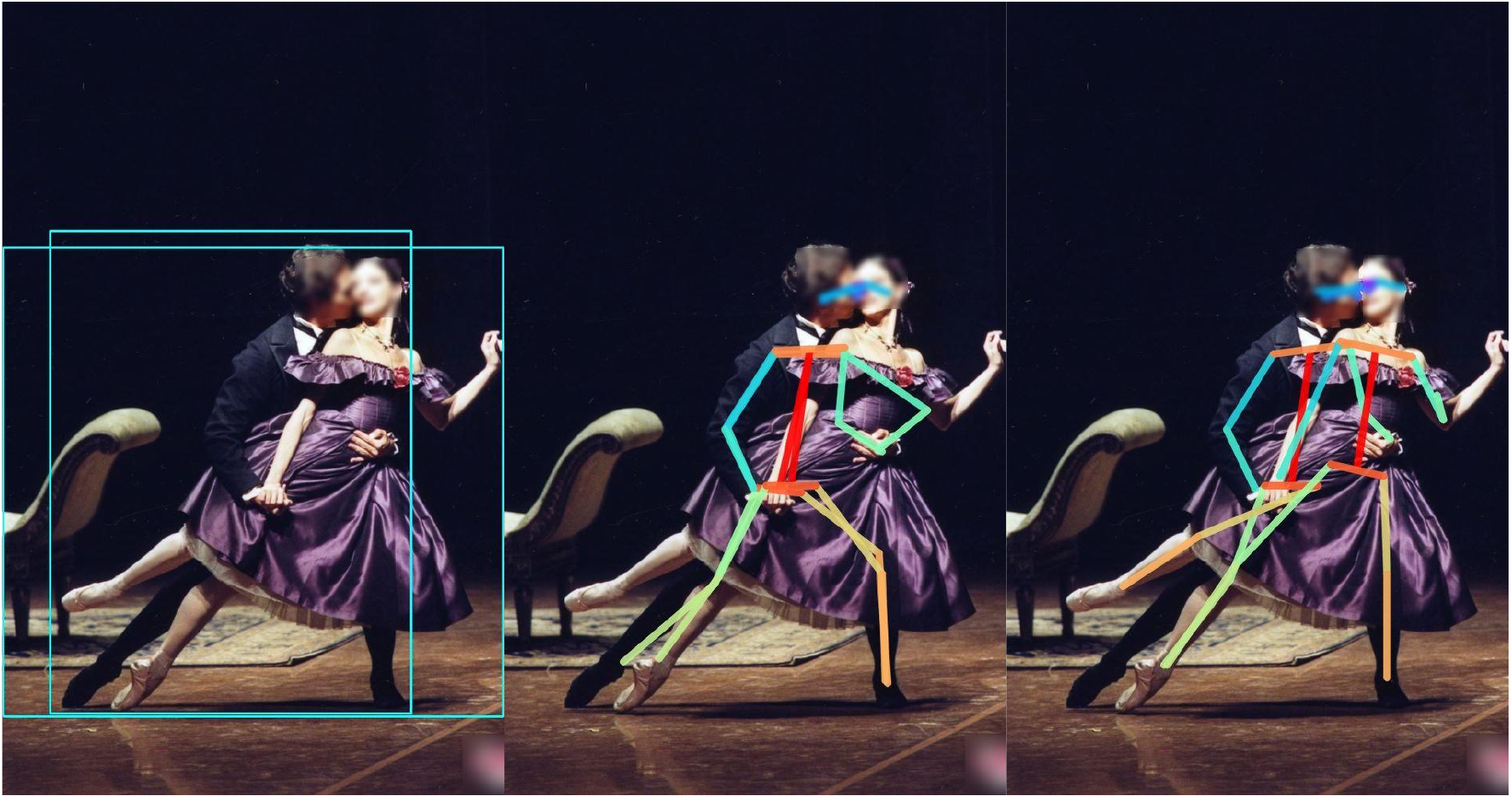}

\includegraphics[height=0.14\textheight,width=0.3\linewidth]{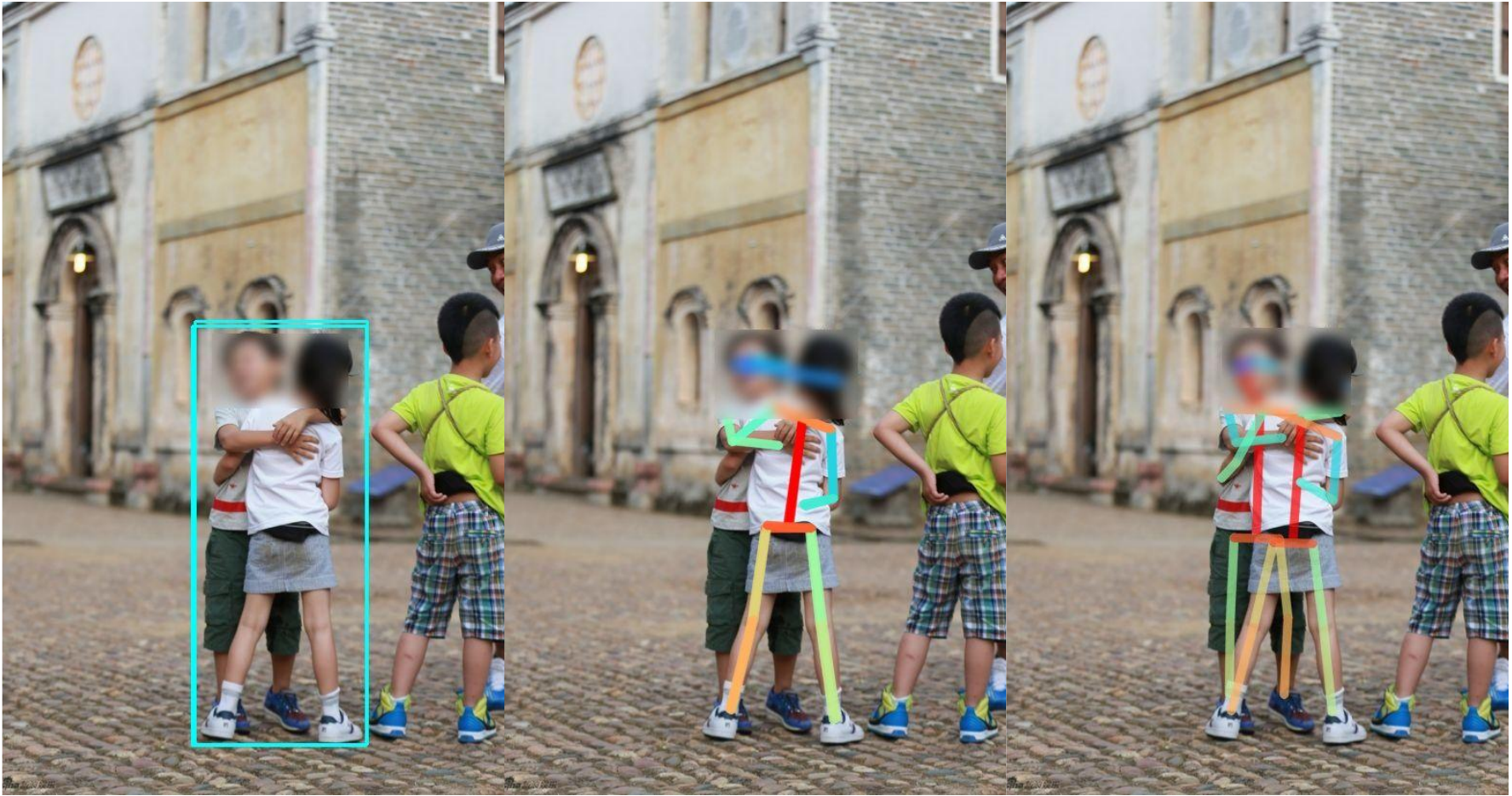}
\includegraphics[height=0.14\textheight,width=0.3\linewidth]{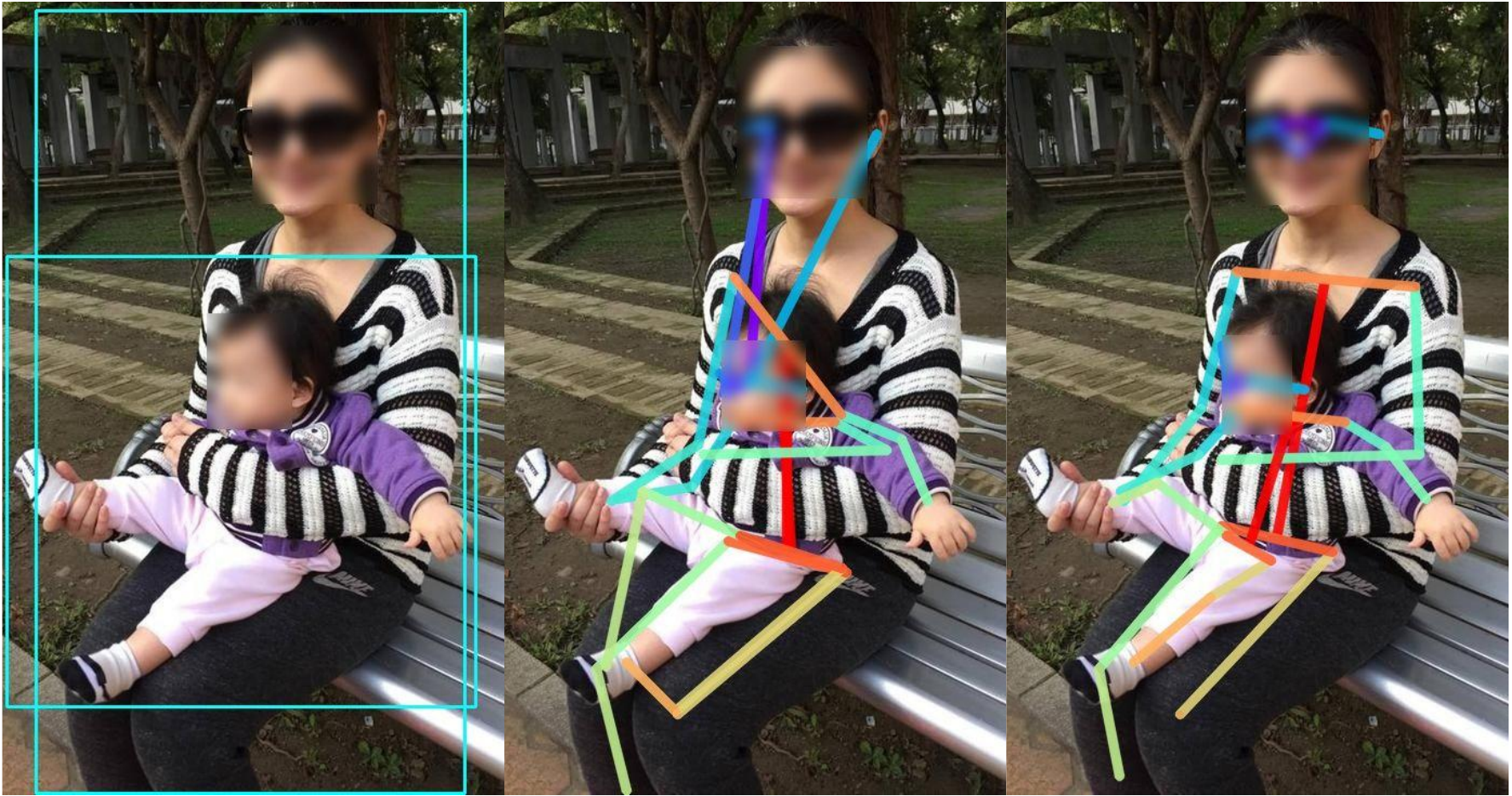}
\includegraphics[height=0.14\textheight,width=0.3\linewidth]{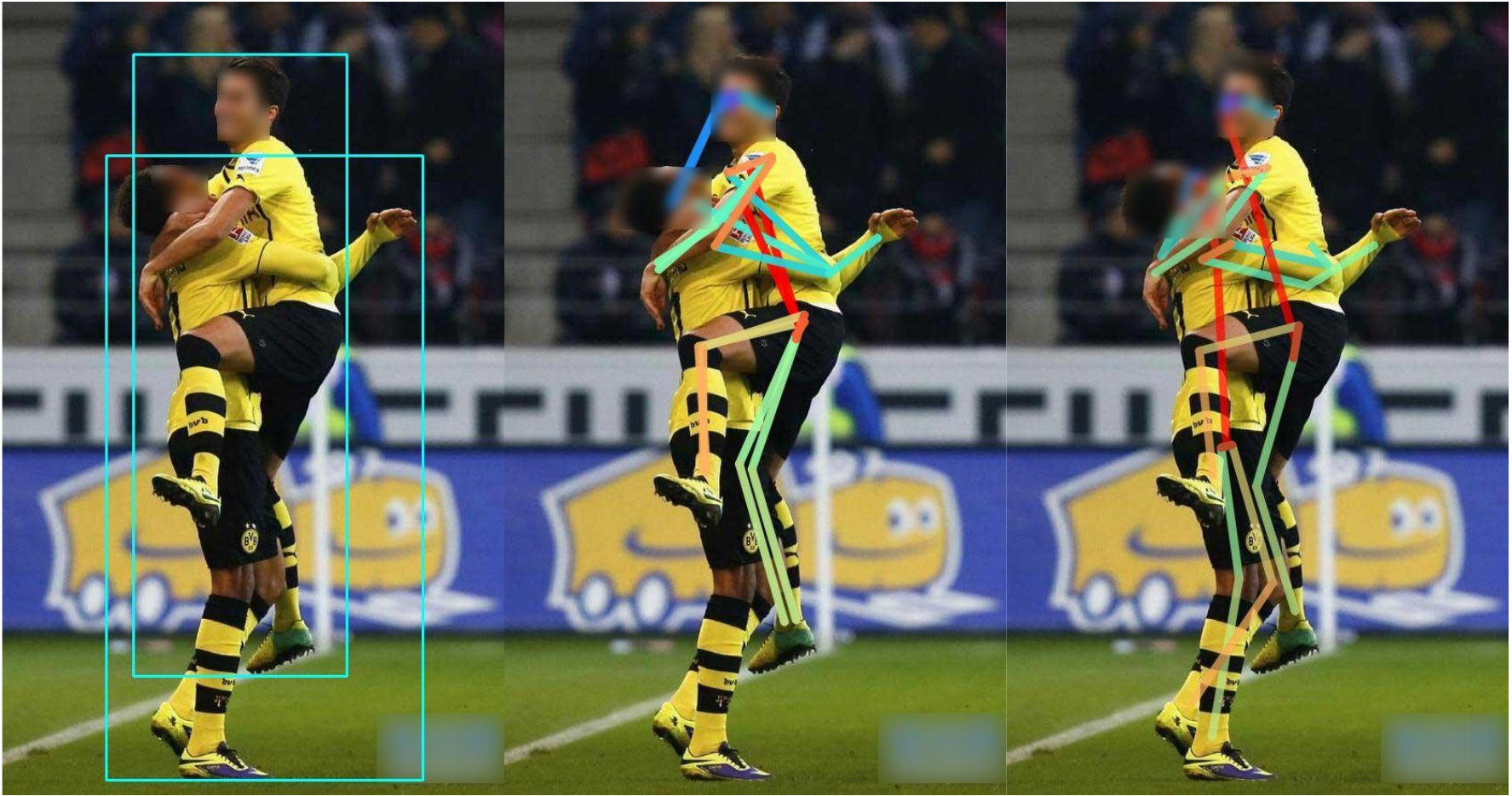}

\includegraphics[height=0.14\textheight,width=0.3\linewidth]{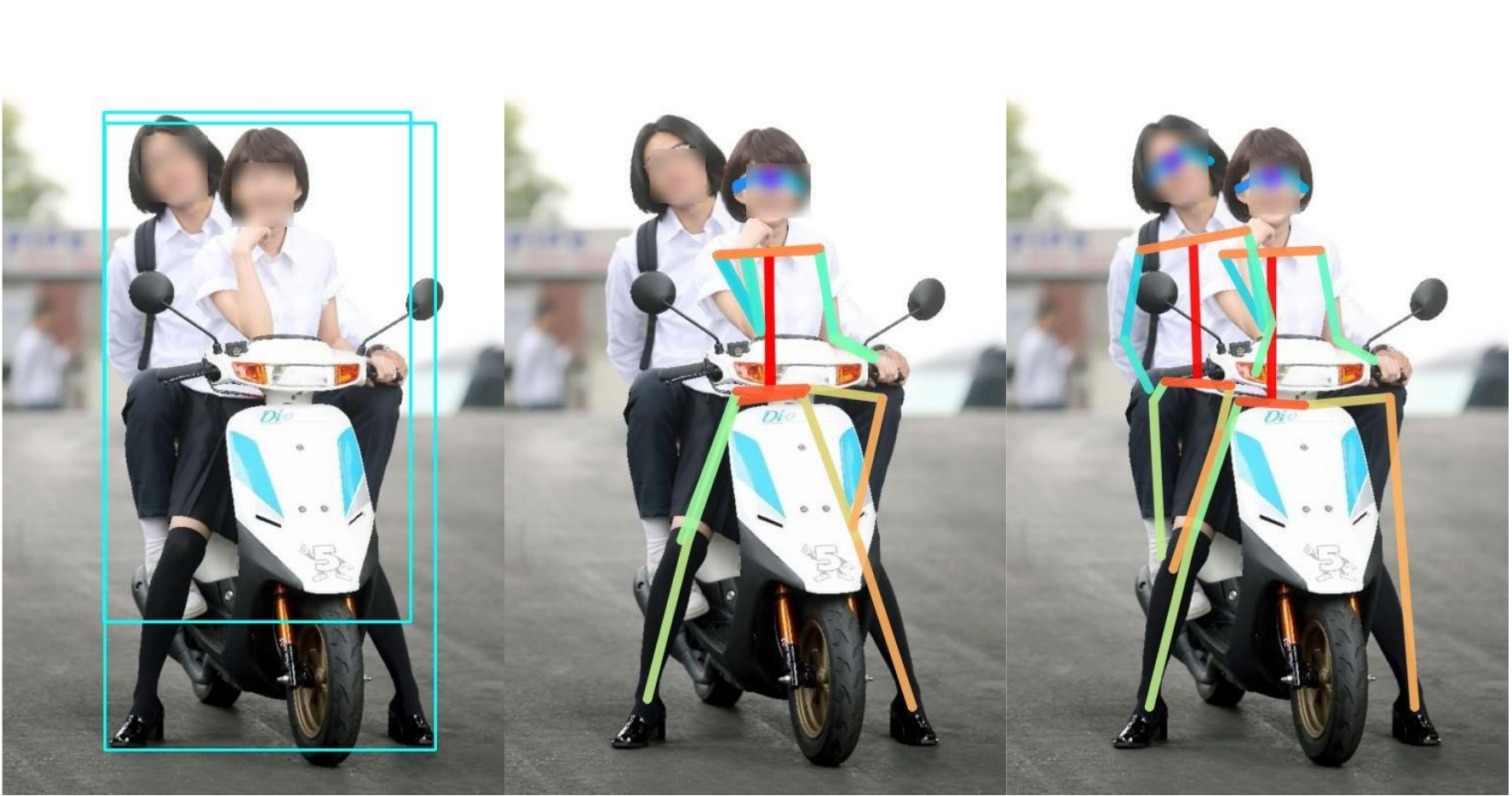}
\includegraphics[height=0.14\textheight,width=0.3\linewidth]{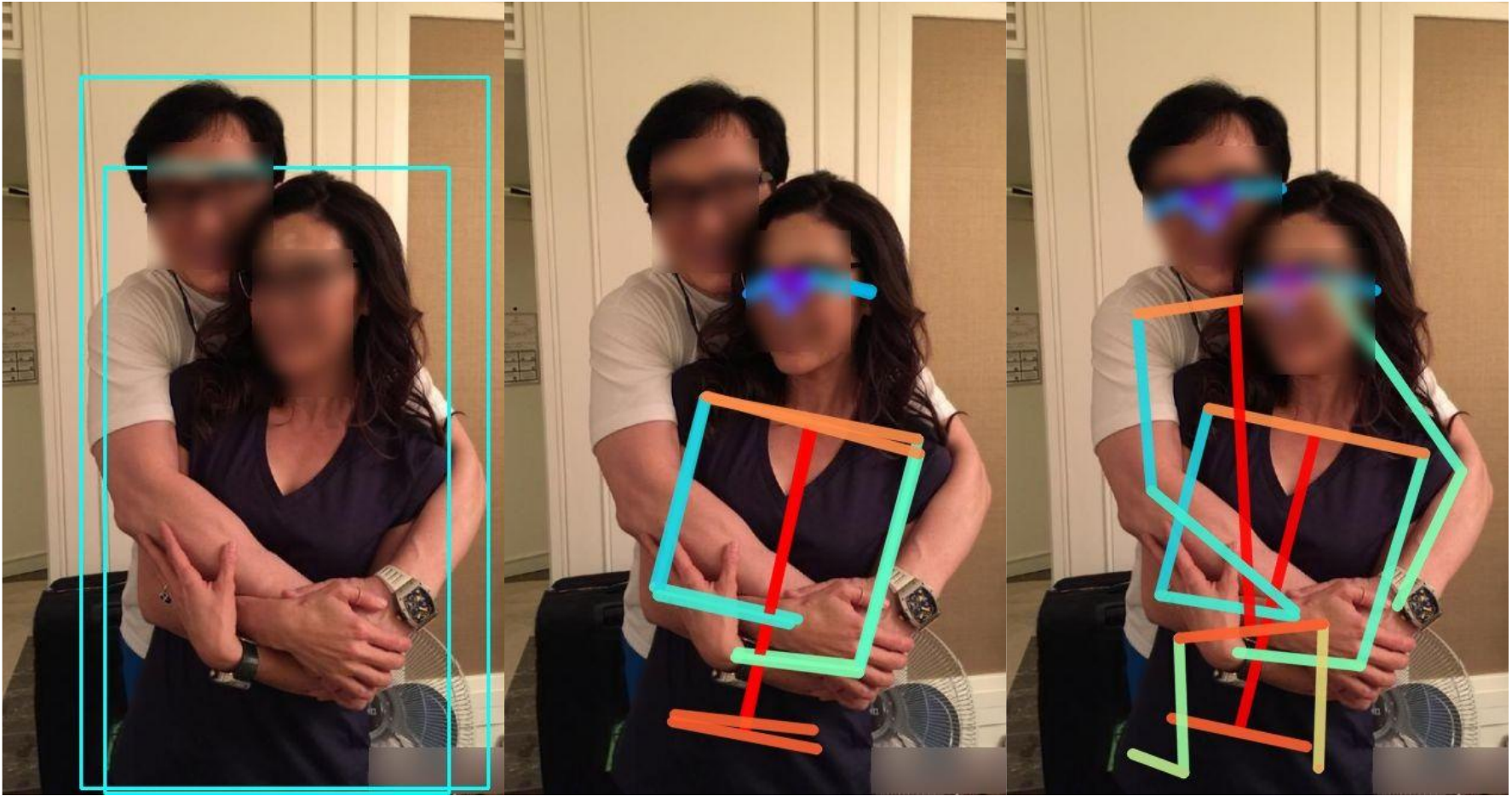}
\includegraphics[height=0.14\textheight,width=0.3\linewidth]{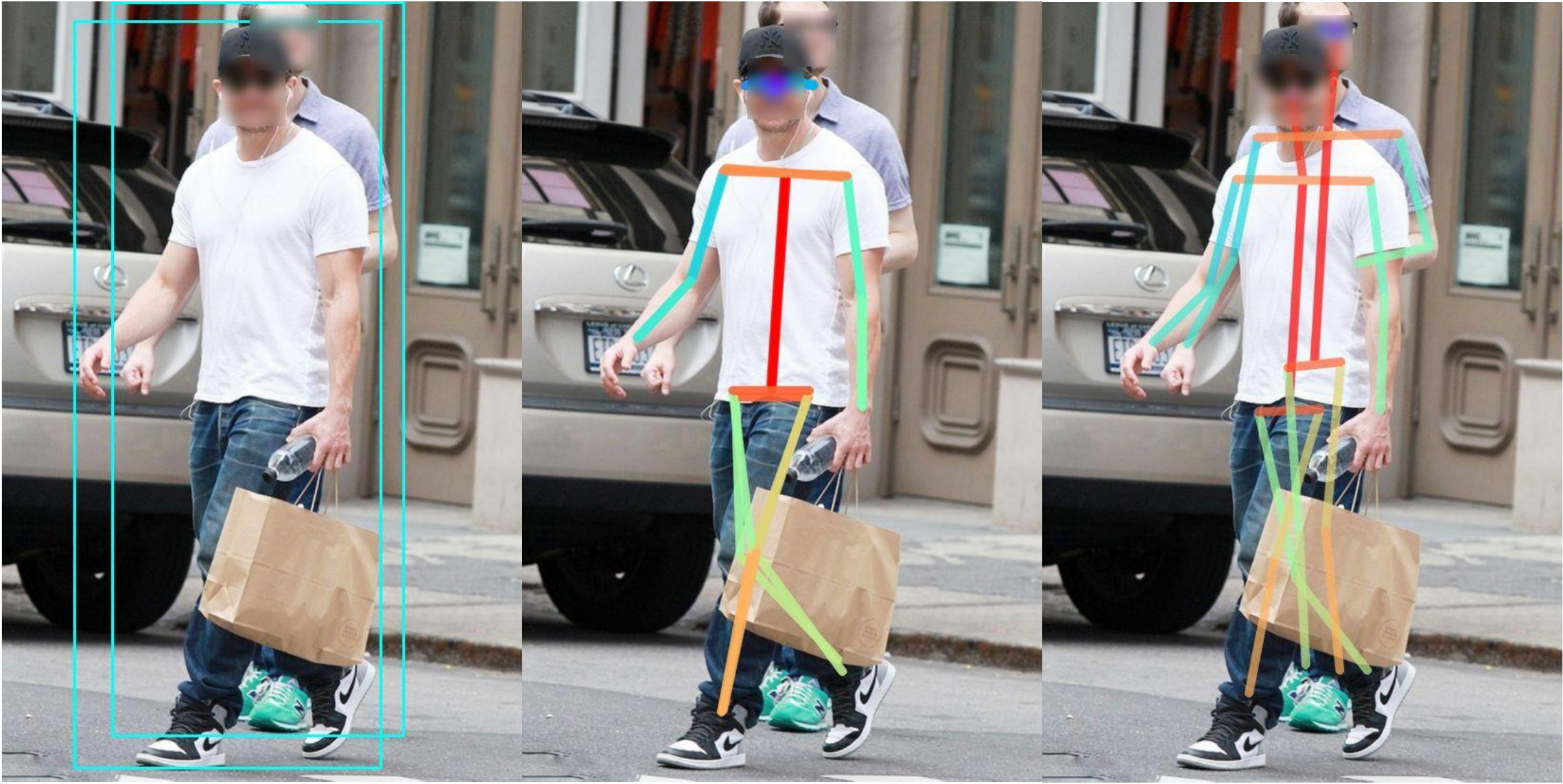}

\includegraphics[height=0.14\textheight,width=0.3\linewidth]{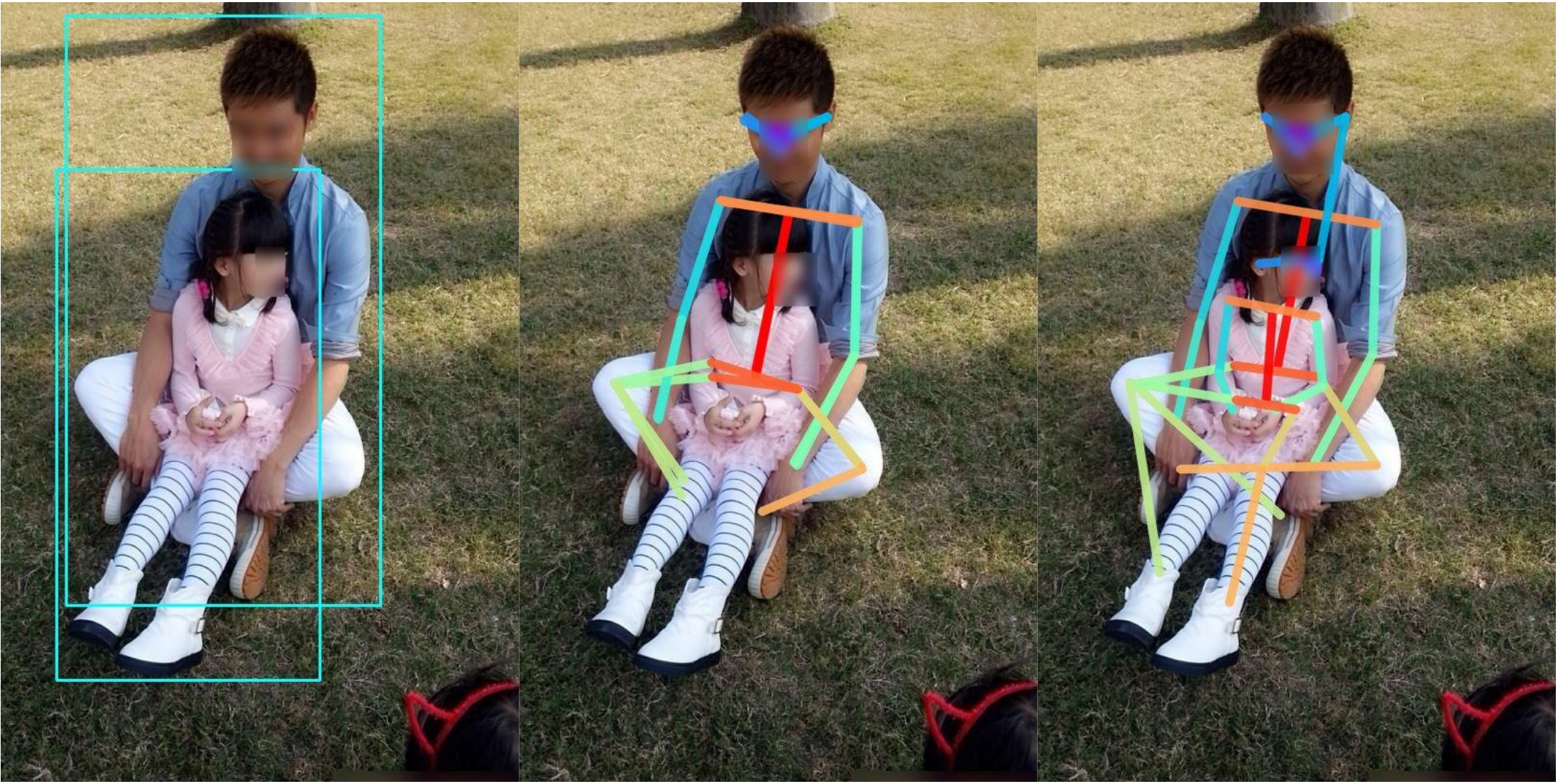}
\includegraphics[height=0.14\textheight,width=0.3\linewidth]{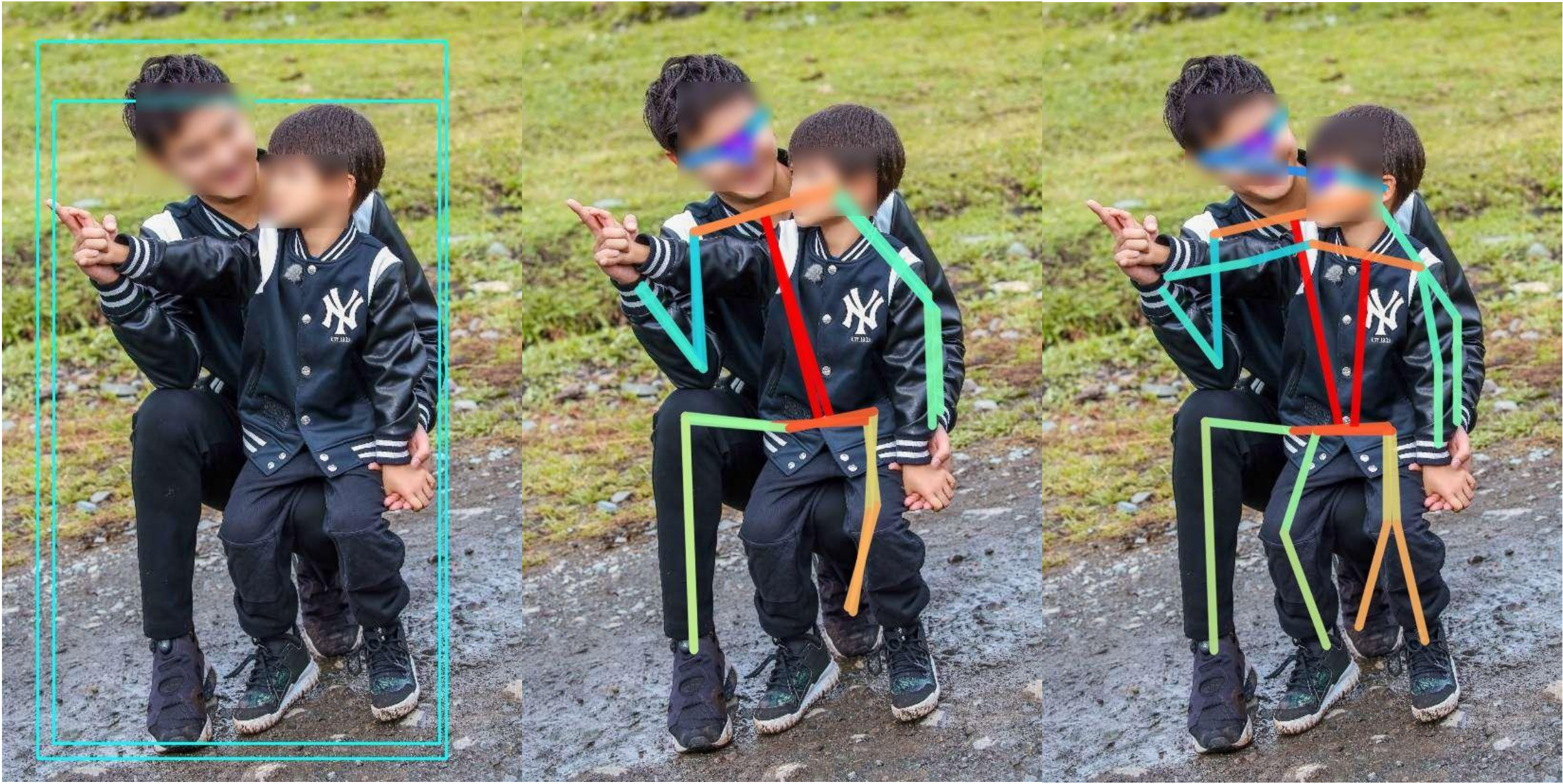}
\includegraphics[height=0.14\textheight,width=0.3\linewidth]{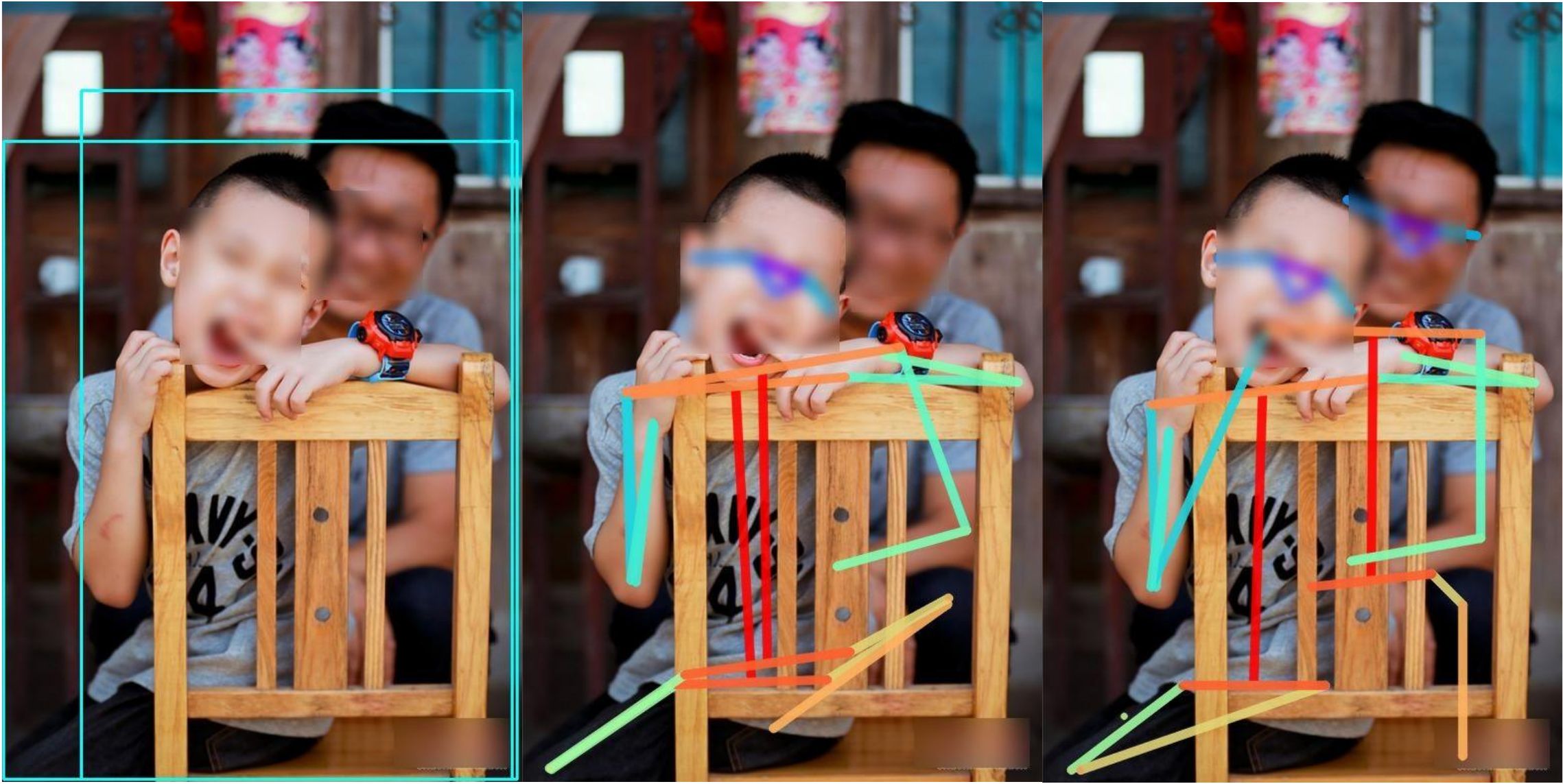}

\includegraphics[height=0.14\textheight,width=0.3\linewidth]{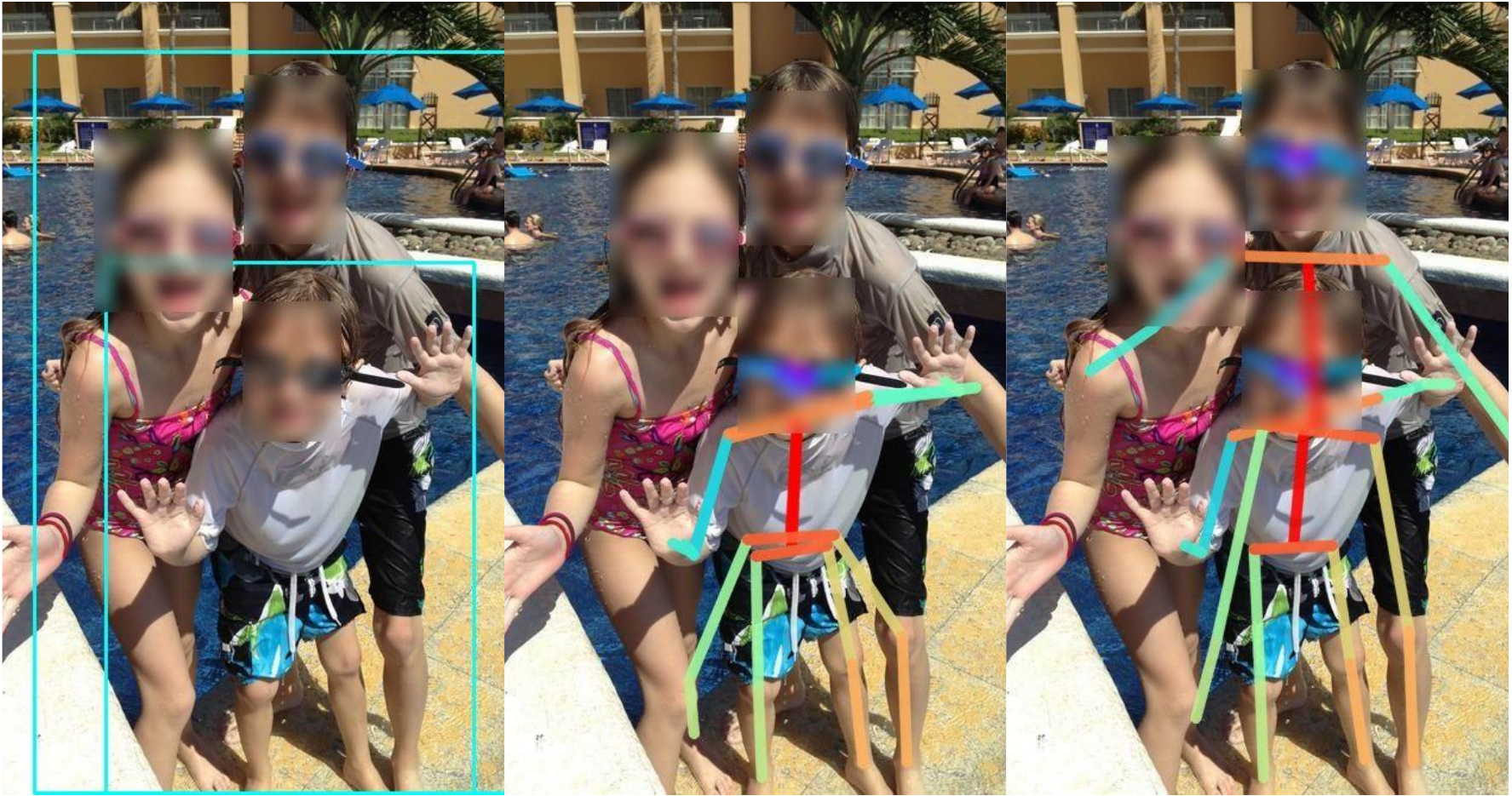}
\includegraphics[height=0.14\textheight,width=0.3\linewidth]{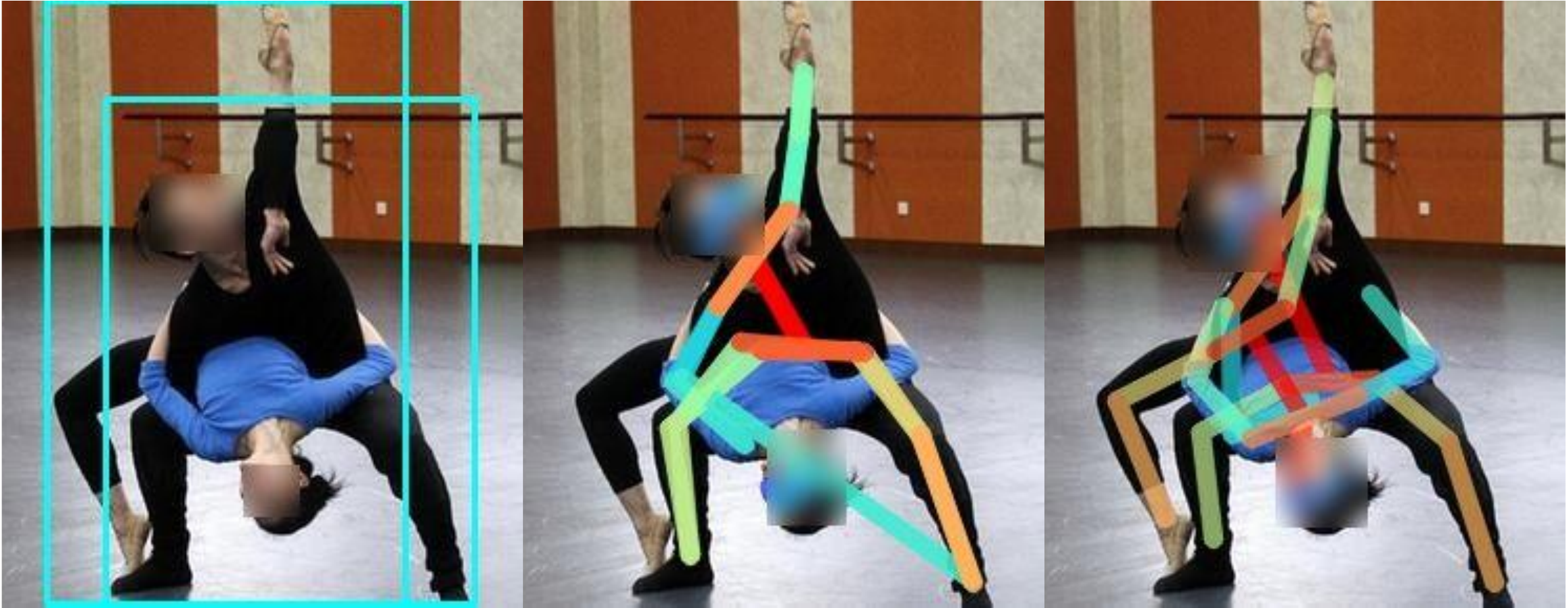}
\includegraphics[height=0.14\textheight,width=0.3\linewidth]{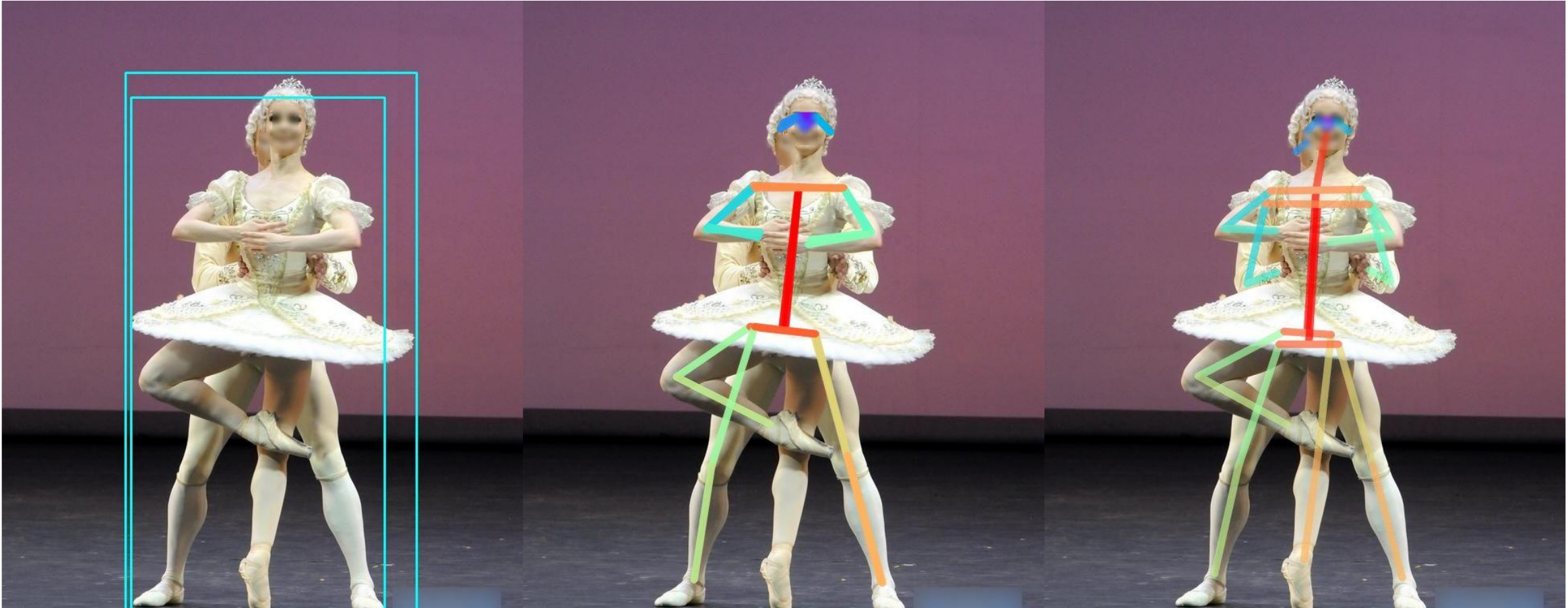}

\end{center}
 \vspace*{-0.2in}
\caption{Qualitative results on OCHuman \texttt{val} set. Each image (left to right) shows input bounding boxes, HRNet predictions and MIPNet predictions. Due to occlusions, HRNet often misses the person in the background which is recovered by MIPNet. Please see additional results in supplemental.}
\label{fig:qualitative}
 \end{figure*}

Figure~\ref{fig:qualitative} shows qualitative results on several examples from OCHuman, highlighting the effectivness of MIPNet in recovering multiple poses under challenging conditions.


\label{sec:experiments_ochuman}

 \begin{figure}
 \captionsetup{font=small}
 \centering
  \includegraphics[width=0.7\linewidth]{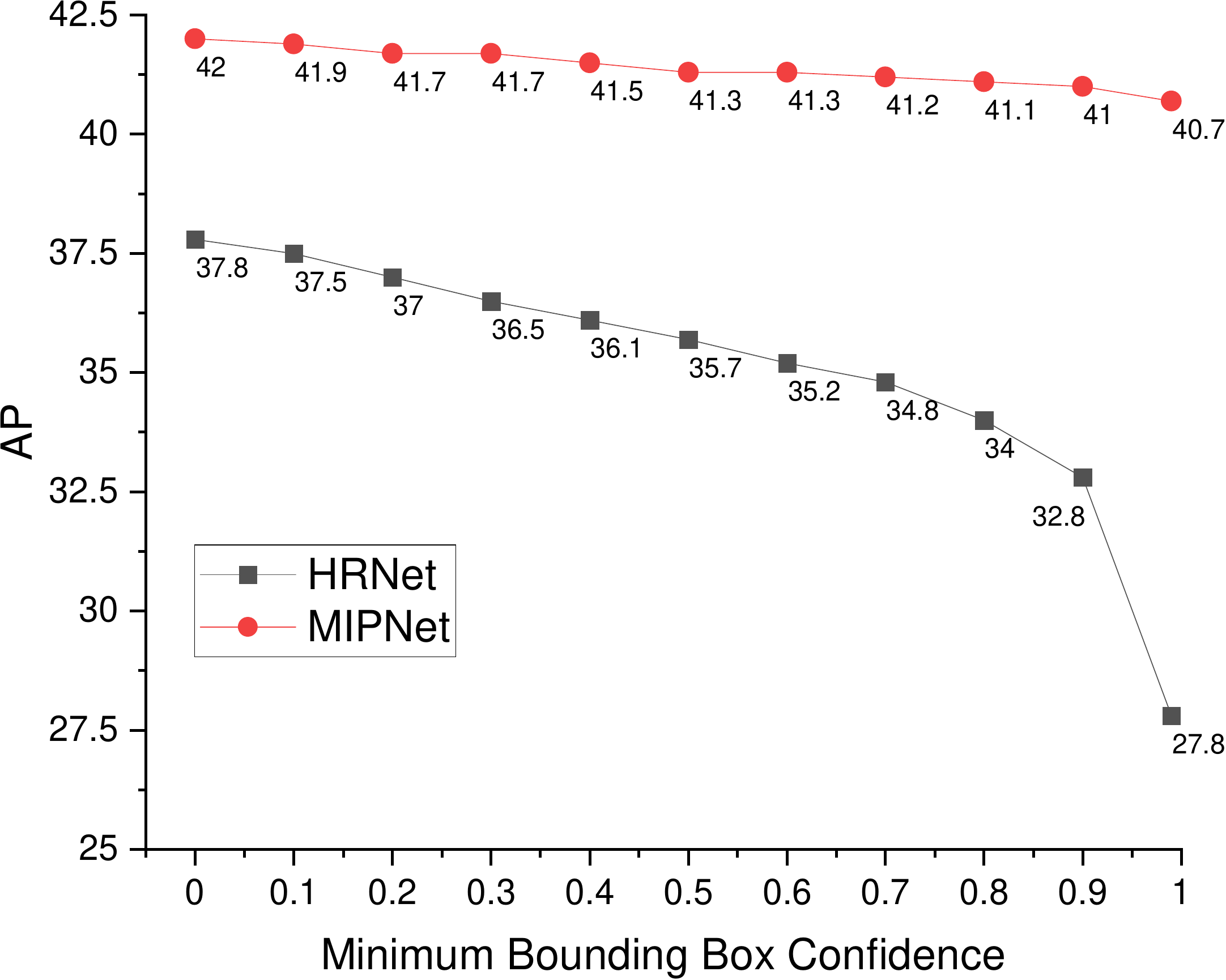}
  \vspace*{-0.1in}
    \caption{Unlike HRNet, MIPNet maintains a stable performance as a function of detector confidence for selecting input bounding boxes. Results are shown using HRNet-W48-$384\times288$ evaluated on OCHuman \texttt{val} set.}
 \label{fig:bb_decay}
 \vspace*{-0.3in}
 \end{figure}
 
\textbf{Robustness to Human Detector Outputs:} The performance of top-down methods is often gated by the quality of human detection outputs. We analyze the robustness of HRNet and MIPNet with varying detector confidence on OCHuman in Fig.~\ref{fig:bb_decay}. As expected, HRNet performance degrades as low confidence bounding boxes are filtered out, leading to missed detections on occluded persons. Specifically, HRNet performance degrades from $37.8$ AP ($30637$ bounding boxes) to $32.8$ AP ($6644$ bounding boxes), when the detector confidence is varied from $0$ to $0.9$. Since HRNet is only able to provide a single output per bounding box, the average precision drops corresponding to misdetections on the occluded persons. In contrast, MIPNet maintains a relatively stable performance (drop of $1$ AP) as shown in Fig.~\ref{fig:bb_decay} for the same inputs. Since MIPNet can predict multiple instances, it can recover pose configurations for occluded persons despite misdetection of their corresponding bounding boxes. This is a desirable property afforded by the proposed MIPNet.

\begin{table}
    \small
    \centering
    \setlength{\tabcolsep}{2pt}
    \renewcommand{\arraystretch}{1.1} 
    \begin{tabular}{@{}l | c c | c c | c c@{}}
    \hline
    \multirow{2}{*}{Method} & \multicolumn{2}{c|}{COCO} & \multicolumn{2}{c|}{CrowdPose} & \multicolumn{2}{c}{OCHuman}  \\
    & \texttt{val} & \texttt{test} & \texttt{val} & \texttt{test} & \texttt{val} & \texttt{test} \\
    \hline
    \multicolumn{7}{c}{\scriptsize{Comparison with Top Down Methods, ResNet101 + YOLO-v3}} \\
    \hline
    MaskRCNN~\cite{he2017mask} & - &  $64.8$ & - & $57.2$ & - & $20.2$ \\
    AlphaPose~\cite{li2019crowdpose} & - &  $70.1$ & - & $61.0$ & - & - \\
    JC-SPPE~\cite{li2019crowdpose} & - &  $70.9$ & - & $66.0$ & - & - \\
    AlphaPose+~\cite{qiu2020peeking} & - &  $72.2$ & - & $68.5$ & - & $27.5$ \\
    OPEC-Net~\cite{qiu2020peeking} & - &  $73.9$ & - & $\mathbf{70.6}$ & - & $29.1$ \\
    SBL~\cite{xiao2018simple} & - &  $73.7$ & - & $60.8$ & - & $24.1$ \\
    MIPNet (Ours) & $\mathbf{72.7}$ &  $\mathbf{74.2}$ & $\mathbf{63.4}$ & $68.1$ & $\mathbf{32.8}$ & $\mathbf{35.0}$ \\
    \hline
    \multicolumn{7}{c}{\scriptsize{Comparison with Top Down Methods, HRNet-W48-384 + Faster R-CNN}} \\
    \hline
    HRNet~\cite{sun2019deep}   & $\mathbf{76.3}$ &  $75.5$ &  $68.0$ & $69.3$ & $37.8$ & $37.2$ \\
    MIPNet~(Ours)   & $\mathbf{76.3}$ &  $\mathbf{75.7}$ & $\mathbf{68.8}$ & $\mathbf{70.0}$ & $\mathbf{42.0}$ & $\mathbf{42.5}$ \\
    \hline
        \multicolumn{7}{c}{\scriptsize{Comparison with Bottom Up Methods, Multi-scale $[\times 2, \times 1, \times 0.5]$}} \\
    \hline
    $\text{AE}$~\cite{newell2017associative}   & $-$ &  $65.5$ & $-$ & $-$ & $40.0$ & $32.8$ \\
    $\text{HghrHRNet}$~\cite{cheng2019higherhrnet}   & $67.1$ &  $70.5$ & - & $67.6$ & - & - \\
    $\text{HghrHRNet+UDP}$~\cite{huang2020devil}   & $-$ &  $70.5$ & - & $68.2$ & - & - \\
    $\text{HGG}$~\cite{jin2020differentiable}   & $68.3$ & $67.6$  & - & - & $41.8$  & $36.0$ \\
    MIPNet~(Ours, Top Down)   & $\mathbf{76.3}$ &  $\mathbf{75.7}$ & $\mathbf{68.8}$ & $\mathbf{70.0}$ & $\mathbf{42.0}$ & $\mathbf{42.5}$ \\
    \hline
    
    \end{tabular}
    \vspace*{-0.1in}
    \caption{Comparison with state-of-the-art methods using bounding boxes from a human detector on various datasets. Other numbers are reported from the respective publications.}
    \label{tab:detector_performance}
    \vspace*{-0.2in}
\end{table}

\label{sec:experiments_person_detector}

 \vspace*{-0.1in}
\section{Discussions}
\label{sec:discussions}

\noindent
\textbf{Comparison to Two-Heads baseline:} We compare MIPNet against the Two-Heads baseline which has a primary head ($\lambda=0)$ and a secondary head ($\lambda=1)$ in Table \ref{tab:two_head_baseline}. To analyze the effect of head capacity in multi-instance prediction, we create two baselines: Two-Heads (\textit{light}), and Two-Heads (\textit{heavy}). MIPNet consistently outperforms the Two-Heads baseline on the OCHuman dataset. Please refer supplemental for more details.

\begin{table}
\centering
\captionsetup{font=small}
    \setlength{\tabcolsep}{1.9pt}
    \small
    \renewcommand{\arraystretch}{1.0} 
    \begin{tabular}{@{}l|c|c c c|c c c@{}}
    \hline
   \multirow{2}{*}{Method} & \multirow{2}{*}{\#Params} & \multicolumn{3}{c|}{COCO} & \multicolumn{3}{c}{OCHuman}\\
    & & $\text{AP}$ & $\text{AP}^{50}$ & $\text{AP}^{75}$ & $\text{AP}$ & $\text{AP}^{50}$ & $\text{AP}^{75}$   \\
    \hline
    HRNet & 28.5M & 76.5 & 93.5 & 83.7 & 63.1 & 79.4 & 69.0 \\
    Two-Heads (\textit{light}) & 28.6M & 76.7 & 93.4 & 84.0 & 64.0 & 78.7 & 71.2 \\
    Two-Heads (\textit{heavy}) & 48.9M & 77.1 & 94.1 & \textbf{85.5} & 69.8 & 84.5 & 74.9 \\
    MIPNet & 28.6M & \textbf{77.6} & \textbf{94.4} & 85.3 & \textbf{72.5} & \textbf{89.2} & \textbf{79.4} \\
    \hline
    \end{tabular}
    \vspace*{-0.1in}
    \caption{Comparison with the Two-Heads baseline (\textit{light}, \textit{heavy}) and HRNet on the \texttt{val} sets using HRNet-W32 backbone with $256\times192$ input resolution and ground-truth bounding boxes.}
    \label{tab:two_head_baseline}
\end{table}

\begin{figure}
 \begin{center}
  \includegraphics[height=0.18\textheight,width=1\linewidth]{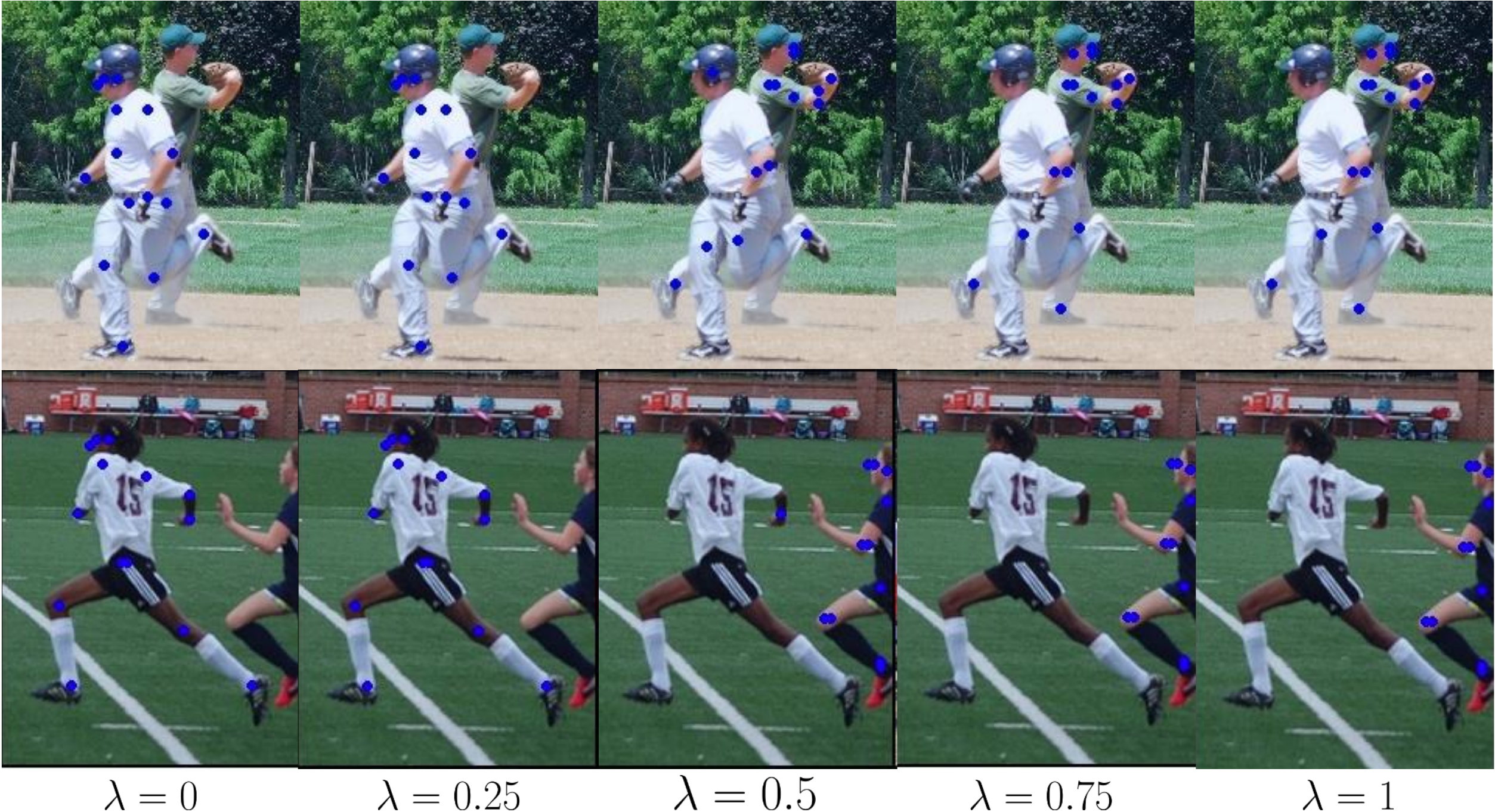}
 \end{center}
 \vspace*{-0.2in}
   \caption{As $\lambda$ is varied from $0$ to $1$ during inference, the keypoints (in blue) gradually shift from the foreground person to the other pose instance within the bounding box.} 
 \label{fig:continuous_lambda}
  \end{figure}

\noindent
\textbf{Visualization with continuous $\lambda$:} MIPNet's ability to predict multiple instances provides a useful tool to visualize how predictions can dynamically switch between various pose configurations. After training MIPNet using an one-hot representation of $\lambda$, during inference, we use a soft representation of $[\lambda, 1-\lambda]$ as instance-selector for the MIPNet. Fig.~\ref{fig:continuous_lambda} shows how the predicted keypoints gradually shift from the foreground person to the other pose instance within the bounding box, as $\lambda$ is varied from $0$ to $1$.

\begin{figure}
    \centering
    \includegraphics[height=0.15\textheight,width=0.35\linewidth]{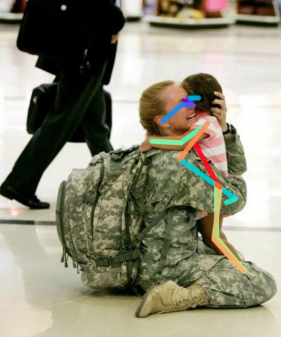}
    \includegraphics[height=0.15\textheight,width=0.35\linewidth]{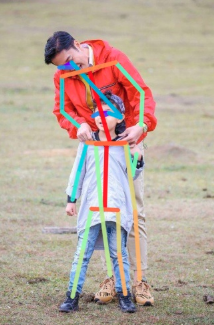}
     \vspace*{-0.1in}
    \caption{MIPNet fails in some cases with significant scale difference between multiple persons in the bounding box.}
    \label{fig:failure}
    \vspace*{-0.2in}
\end{figure}

\noindent
\textbf{Limitations:} In some cases, MIPNet can fail due to large difference in the scale of the various pose instances in a given bounding box, as shown in Figure~\ref{fig:failure}. 

\label{sec:experiments_ablation}

\vspace*{-0.2cm}
\section{Conclusion}
\label{sec:conclusion}
Top-down 2D pose estimation methods make the key assumption of a single person within the input bounding box. While these methods have shown impressive results, the single person assumption limits their ability to perform well in crowded scenes with occlusions. Our proposed Multi-Instance Pose Network, MIPNet, enables top-down methods to predict multiple instances for a given input. MIPNet is efficient in terms of the number of additional network parameters and is stable with respect to the quality of the input bounding boxes. MIPNet achieves state-of-art results on challenging datasets with significant crowding and occlusions. We believe that the concept of predicting multiple instances is an important conceptual change and will inspire a new research direction for top-down methods.

\clearpage
\newpage
{\small
\bibliographystyle{ieee_fullname}
\bibliography{references}
}

\end{document}


\title{Multi-Instance Pose Networks: Rethinking Top-Down Pose Estimation}
\author{
}

\maketitle



\section{Multi-Instance Modulation Block (MIMB) Code}
In this section, we describe the code of MIMB in \texttt{PyTorch}. The code in Listing.~\ref{code:mimb} outlines the details of functions $\mathbf{F}_{sq}$, $\mathbf{F}_{ex}$ and $\mathbf{F}_{em}$. $\mathbf{F}_{sq}$ is a simple global average pool and $\mathbf{F}_{ex}$ and $\mathbf{F}_{em}$ are two-layered neural networks. MIMB can be incorporated in any existing feature extraction
backbone, with a relatively simple ($<15$ lines) code change.

\begin{lstlisting}[language=Python, caption=Code for MIMB., label=code:mimb]
class MIMB(nn.Module):
  def __init__(self, num_channels=c, reduce=r):
    super(MIMB, self).__init__()
    self.F_sqn = nn.AdaptiveAvgPool2d(1)
        
    self.F_ex = nn.Sequential(
        nn.Linear(c, c // r, bias=False),
        nn.ReLU(inplace=True),
        nn.Linear(c // r, c, bias=False),
        nn.Sigmoid()
    )

    self.F_em = nn.Sequential(
        nn.Linear(2, c // r),
        nn.BatchNorm1d(c // r),
        nn.ReLU(inplace=True),
        nn.Linear(c // r, c),
        nn.Sigmoid()
    )
        return

    def forward(self, x, lambda):
        b, c, _, _ = x.size()
        y = self.F_sqn(x).view(b, c)
        y = self.F_ex(y).view(b, c, 1, 1)

        z = self.F_em(lambda).view(b, c, 1, 1)

        out = x * y.expand_as(x) * z.expand_as(x)
        return out
\end{lstlisting}

\section{Implementation Details}
We merge all the instances from $\lambda = 0$ to $N-1$ and then apply oks-nms. During the merger, we discount the confidence of the instance $\lambda=i$ by $\gamma^i$. As the primary instance ($\lambda = 0$) is always centralized in the input, this confidence discounting avoids suppression of a high resolution primary predictions by a low resolution $\lambda > 0$ prediction. We use $\gamma = 0.9$ in all our experiments.

\clearpage

\begin{figure*}
\centering
\includegraphics[width=0.9\linewidth]{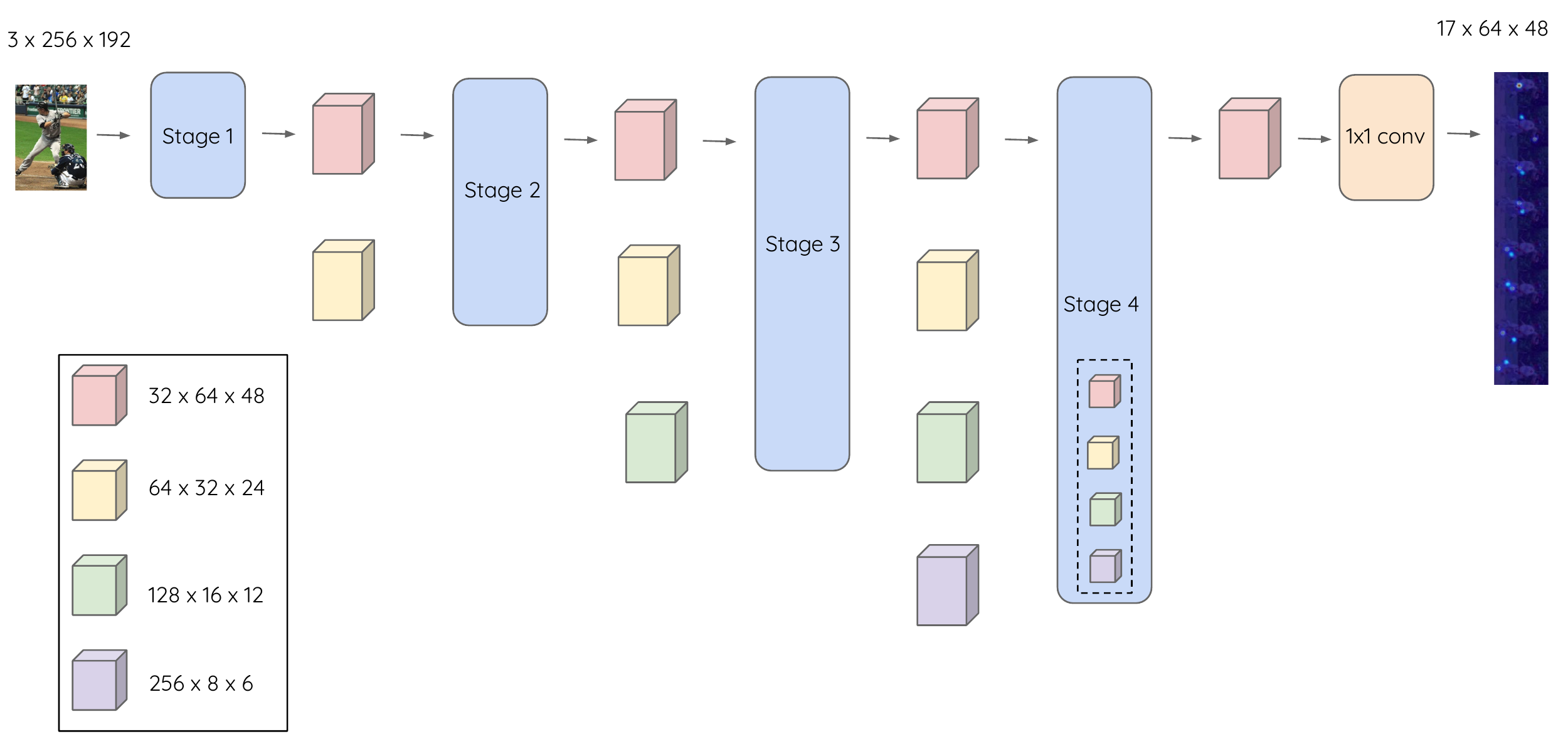}
\caption{Illustration of HRNet-W32 backbone at input resolution $256 \times 192$. The blue blocks depict the four stages in the architecture.}
\label{fig:supplementary:hrnet}
\end{figure*}

\begin{figure*}
\centering
\includegraphics[width=0.9\linewidth]{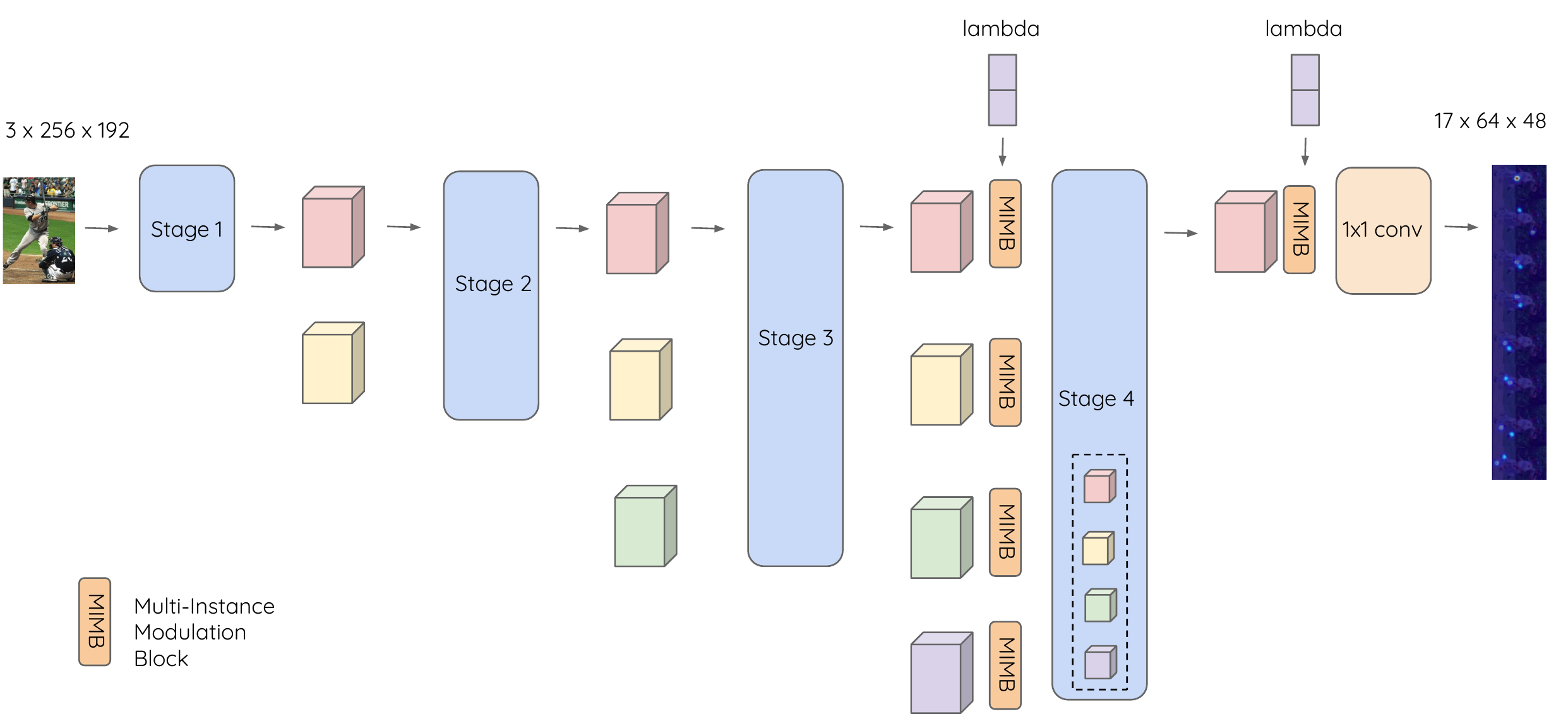}
\caption{Illustration of MIPNet with HRNet-W32 backbone at input resolution $256 \times 192$. We insert 5 MIMBs into the HRNet, 4 MIMBs after Stage 3 and 1 MIMB after Stage 4.}
\label{fig:supplementary:hrnet_mimb}
\end{figure*}

\textbf{MIPNet-HRNet:}
Figure.~\ref{fig:supplementary:hrnet} shows the architecture details of HRNet~\cite{sun2019deep}. For simplicity, we only show backbone HRNet-W32 at input size $256 \times 192$, other HRNet backbones follow similar pipeline. Figure.~\ref{fig:supplementary:hrnet_mimb} shows the architecture of MIPNet, where multiple MIMBs are inserted at various stages.

\clearpage

\begin{figure*}
\centering
\includegraphics[width=0.9\linewidth]{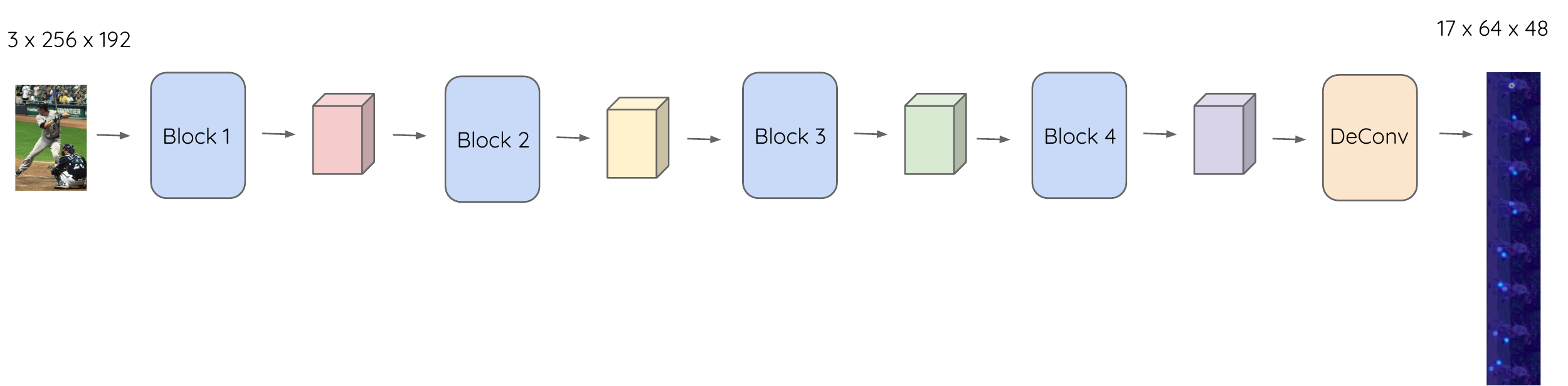}
\caption{Illustration of SimpleBaseline architecture. The blue blocks represent the four blocks in the encoder of SimpleBaseline.}
\label{fig:supplementary:resnet}
\end{figure*}

\begin{figure*}
\centering
\includegraphics[width=0.9\linewidth]{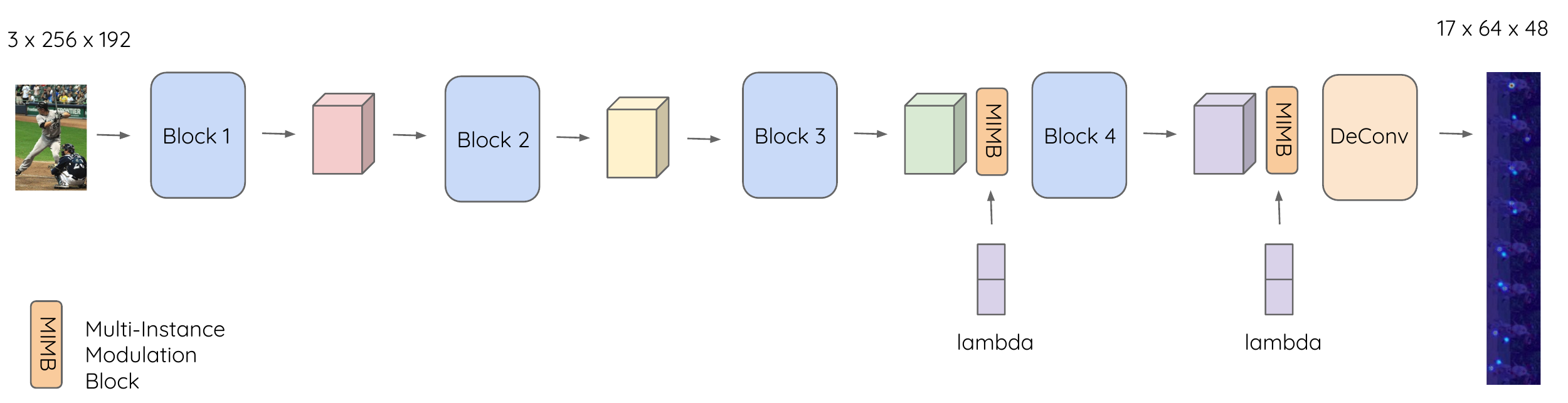}
\caption{Illustration of MIPNet with SimpleBaseline architecture. We insert 2 MIMBs into the encoder after Block 3 and Block 4.}
\label{fig:supplementary:resnet_mimb}
\end{figure*}

\textbf{MIPNet-SimpleBaseline:}
Figure.~\ref{fig:supplementary:resnet} shows the architecture details of SimpleBaseline~\cite{xiao2018simple}. Figure.~\ref{fig:supplementary:resnet_mimb} shows the architecture of MIPNet, where multiple MIMBs are inserted in the encoder of the pose estimator.

\section{Diminishing Returns with $\mathbf{N=3, 4}$}
We observed a small improvement in AP using $N=3$ and $N=4$ on top of $N=2$ respectively on the datasets when evaluated using ground-truth bounding boxes. This is consistent with the fact that most datasets have very few examples with three or more ground-truth pose instances per bounding box (Refer data statistics in the paper). Note, on the more occluded OCHuman dataset, increasing $N$ gives better performance.

\begin{table}[H]
    \centering
    \small
    \renewcommand{\arraystretch}{1.0} 
    \begin{tabular}{@{}l|c|c|c@{}}
    \hline

Inference  & COCO  & CrowdPose & OCHuman    \\
    \hline
    HRNet  & 78.1 & 72.8 & 65.0 \\
    \hline
MIPNet, $N=2$ & \textbf{78.8}  & 73.7  & 74.1 \\
MIPNet, $N=3$ & 78.4  & \textbf{73.9}  & 74.3 \\
MIPNet, $N=4$ & 78.6  & 73.7  & \textbf{74.7} \\
\hline
  \end{tabular}

    \caption{Performance of MIPNet on \texttt{val} sets using ground truth bounding boxes with increasing $N$. We use the backbone W48 with image resolution $384 \times 288$, and compare with the same HRNet configuration. By default, HRNet only predicts a single instance.}
    \label{tab:supplementary:diminishing_N}
\end{table}

\clearpage
\section{Additional Results on COCO, CrowdPose and OCHuman}

\subsection{Additional results on COCO}

Table~\ref{tab:supplementary:coco_gt} shows additional metrics for comparison between MIPNet ($N=2$) and various baseline architectures on COCO {\tt val} dataset using ground truth bounding boxes for evaluation. We also report GFLOPs for each model. Note that for all baseline evaluations for HRNet and SimpleBaseline, we follow the same protocol as outlined in the respective papers~\cite{sun2019deep, xiao2018simple}.

\begin{table*}[b]
    \centering
    \small
    \renewcommand{\arraystretch}{1.0} 
    \begin{tabular}{@{}l|l|l|c|l|l c c c c c c c c c@{}}
    \hline

Method  & Arch  & Input Size  & GFLOPS &$\text{AP}$ & $\text{AP}^{50}$ & $\text{AP}^{75}$ & $\text{AP}^\text{M}$ & $\text{AP}^\text{L}$ & $\text{AR}$ & $\text{AR}^{50}$ & $\text{AR}^{75}$ & $\text{AR}^\text{M}$ & $\text{AR}^\text{L}$  \\
    \hline
      SBL  & R-50    & $256 \times 192$ &  8.90 & 
      72.4 & 91.5 & 80.4 & 69.7 & 76.5 & 75.6 & 93.0 & 82.3 & 72.3 & 80.4\\
  MIPNet & R-50    &  $256 \times 192$ & 16.3 &
      \textbf{73.3 (+0.9)} & \textbf{93.3} & \textbf{81.2} & \textbf{70.6} & \textbf{77.6} & \textbf{76.7} & \textbf{94.2} & \textbf{83.4} & \textbf{73.4} & \textbf{81.6}\\
      \hline
      SBL  & R-101    &   $256 \times 192$ & 12.4 &
      73.4 & 92.6 & 81.4 & 70.7 & 77.7 & 76.5 & 93.4 & 83.1 & 73.3 & 81.2\\
  MIPNet & R-101     & $256 \times 192$ & 23.1  &
      \textbf{74.1 (+0.7)} & \textbf{93.3} & \textbf{82.3} & \textbf{71.3} & \textbf{78.6} & \textbf{77.4} & \textbf{94.4} & \textbf{84.4} & \textbf{74.1} & \textbf{82.3}\\
      \hline
      SBL  & R-152    &   $256 \times 192$ & 15.7  &
      74.3 & 92.6 & \textbf{82.5} & 71.6 & 78.7 & 77.4 & 93.8 & 84.2 & 74.4 & 82.0\\
  MIPNet & R-152  & $256 \times 192$ & 29.1  &
      \textbf{74.8 (+0.5)} & \textbf{93.3} & 82.4 & \textbf{71.7} & \textbf{79.4} & \textbf{78.2} & \textbf{94.6} & \textbf{84.9} & \textbf{74.7} & \textbf{83.2}\\
      
    \hline

      SBL  & R-50   &  $384 \times 288$ & 20.2 &
      74.1 & 92.6 & 80.5 & 70.5 & 79.6 & 76.9 & 93.2 & 82.7 & 73.0 & 82.6\\
  MIPNet & R-50    &   $384 \times 288$ & 36.7 &
      \textbf{75.3 (+1.2)} & \textbf{93.4} & \textbf{82.4} & \textbf{72.0} & \textbf{80.4} & \textbf{78.4} & \textbf{94.6} & \textbf{84.7} & \textbf{74.6} & \textbf{83.8}\\
      \hline
     SBL & R-101    &  $384 \times 288$ & 27.8  &
      75.5 & 92.5 & 82.6 & 72.4 & 80.8 & 78.4 & 93.6 & 84.5 & 74.9 & 83.8\\
  MIPNet & R-101     &   $384 \times 288$ & 51.9 &
      \textbf{76.0 (+0.5)} & \textbf{93.4} & \textbf{83.5} & \textbf{72.6} & \textbf{81.1} & \textbf{79.1} & \textbf{94.8} & \textbf{85.6} & \textbf{75.5} & \textbf{84.5}\\
      \hline
      SBL  & R-152    &   $384 \times 288$ & 35.5 &
      76.6 & 92.6 & 83.6 & \textbf{73.7} & 81.3 & 79.3 & 94.0 & 85.3 & 75.9 & 84.5\\
  MIPNet & R-152 & $384 \times 288$ & 65.4 & 
      \textbf{77.0 (+0.4)} & \textbf{93.5} & \textbf{84.3} & \textbf{73.7} & \textbf{81.9} & \textbf{80.0} & \textbf{94.9} & \textbf{86.1} & \textbf{76.4} & \textbf{85.3}\\
      
    \hline
    \hline
      HRNet  & H-32   & $256 \times 192$ & 7.10 &
      76.5 & 93.5 & 83.7 & 73.9 & 80.8 & 79.3 & 94.5 & 85.8 & 76.2 & 84.1\\
  MIPNet & H-32 & $256 \times 192$ & 9.80 &
      \textbf{77.6 (+1.1)} & \textbf{94.4} & \textbf{85.3} & \textbf{74.7} & \textbf{81.9} & \textbf{80.6} & \textbf{95.6} & \textbf{87.1} & \textbf{77.3} & \textbf{85.4}\\
      \hline
      HRNet  & H-48    &  $256 \times 192$ & 14.6 &
      77.1 & 93.6 & 84.7 & 74.1 & 81.9 & 79.9 & 94.5 & 86.3 & 76.5 & 85.1\\
  MIPNet & H-48     &   $256 \times 192$ & 20.7 &
      \textbf{77.6 (+0.5)} & \textbf{94.4} & \textbf{85.4} & \textbf{74.6} & \textbf{82.1} & \textbf{80.6} & \textbf{95.6} & \textbf{87.0} & \textbf{77.3} & \textbf{85.5}\\
    
     \hline
      HRNet  & H-32     & $384 \times 288$ & 16.0 &
      77.7 & 93.6 & 84.7 & 74.8 & 82.5 & 80.4 & 94.4 & 86.4 & 77.0 & 85.6\\
  MIPNet & H-32     &  $384 \times 288$ & 22.1 &
      \textbf{78.5 (+0.8)} & \textbf{94.4} & \textbf{85.7} & \textbf{75.6} & \textbf{83.0} & \textbf{81.4} & \textbf{95.6} & \textbf{87.4} & \textbf{78.0} & \textbf{86.3}\\
      \hline
      HRNet  & H-48    &  $384 \times 288$ & 32.9 &
      78.1 & 93.6 & 84.9 & 75.3 & 83.1 & 80.9 & 94.7 & 86.7 & 77.5 & 86.0\\
  MIPNet & H-48    &  $384 \times 288$ & 46.5  &
      \textbf{78.8 (+0.7)} & \textbf{94.4} & \textbf{85.7} & \textbf{75.5} & \textbf{83.7} & \textbf{81.6} & \textbf{95.5} & \textbf{87.5} & \textbf{78.0} & \textbf{86.8}\\
\hline
  \end{tabular}

    \caption{Additional metrics for comparison between MIPNet and various architectures on COCO~\texttt{val} set using ground-truth bounding boxes for evaluation.}
    \label{tab:supplementary:coco_gt}
\end{table*}

Table~\ref{tab:supplementary:coco_bb} shows additional metrics for comparison between MIPNet and HRNet on COCO {\tt val} and {\tt test} datasets using Faster-RCNN bounding boxes, provided by authors of ~\cite{sun2019deep} for evaluation. For HRNet, numbers are reported from their paper~\cite{sun2019deep} (some metrics are not provided).

\begin{table*}[t]
    \centering
    \small
    \renewcommand{\arraystretch}{1.0} 
    \begin{tabular}{@{}l|c|c|l|l c c c c c c c c c@{}}
    \hline

Method  & Arch  & Input Size  & $\text{AP}$ & $\text{AP}^{50}$ & $\text{AP}^{75}$ & $\text{AP}^\text{M}$ & $\text{AP}^\text{L}$ & $\text{AR}$ & $\text{AR}^{50}$ & $\text{AR}^{75}$ & $\text{AR}^\text{M}$ & $\text{AR}^\text{L}$  \\
    \hline
    \multicolumn{13}{c}{\texttt{val}} \\
     \hline 
      HRNet  & H-48    &  $384 \times 288$ &
      \textbf{76.3} & \textbf{90.8} & 82.9 & \textbf{72.3} & \textbf{83.4} & 81.2 & - & - & - & -\\
  MIPNet & H-48    &  $384 \times 288$ &
      \textbf{76.3 (+0.0)} & 90.6 & \textbf{83.0} & 72.1 & 83.3 & \textbf{81.4} & 94.2 & 87.6 & 76. & 88.2\\
       \hline
    \multicolumn{13}{c}{\texttt{test}} \\
    \hline
    \multicolumn{13}{c}{\scriptsize{Bottom-Up}} \\
    \hline
    OpenPose~\cite{cao2018openpose} & - &  - &
      61.8 & 84.9 & 67.5 & 57.1 & 68.2 & 66.5 & - & - & - & -\\
     AE~\cite{newell2017associative} & - &  - &
      65.5 & 86.8 & 72.3 & 60.6 & 72.6 & 70.2 & - & - & - & -\\
     PersonLab~\cite{papandreou2018personlab} & - &  - &
      68.7 & 89.0 & 75.4 & 64.1 & 75.5 & 75.4 & - & - & - & -\\
      MultiPoseNet~\cite{kocabas2018multiposenet} & - &  - &
      69.6 & 86.3 & 76.6 & 65.0 & 76.3 & 73.5 & - & - & - & -\\
    
    \hline
    \multicolumn{13}{c}{\scriptsize{Top-Down}} \\
    \hline
    MaskRCNN~\cite{he2017mask} & R-50 &  - &
      63.1 & 87.3 & 68.7 & 57.8 & 71.4 & - & - & - & - & -\\
    G-RMI~\cite{papandreou2017towards} & R-101 &  $353 \times 257$ &
      64.9 & 85.5 & 71.3 & 62.3 & 70.0 & 69.7 & - & - & - & -\\
    CPN~\cite{chen2018cascaded} & R-Incep &  $384 \times 288$ &
      72.1 & 91.4 & 80.0 & 68.7 & 77.2 & 78.5 & - & - & - & -\\
    RMPE~\cite{fang2017rmpe} & PyraNet &  $320 \times 256$ &
      72.3 & 89.2 & 79.1 & 68.0 & 78.6 & - & - & - & - & -\\
    HRNet  & H-48    &  $384 \times 288$ &
     75.5 & \textbf{92.5} & \textbf{83.3} &\textbf{ 71.9} & \textbf{81.5} & \textbf{80.5} & - & - & - & -\\
  MIPNet & H-48    &  $384 \times 288$ &
      \textbf{75.7 (+0.2)} & 92.4 & 83.3 & 71.4 & 81.2 & \textbf{80.5} & 95.5 & 87.4 & 76.1 & 86.5\\

\hline
  \end{tabular}

    \caption{Additional metrics for comparison between MIPNet and various architectures on COCO \texttt{val} and \texttt{test} set using Faster-RCNN bounding boxes for evaluation.}
    \label{tab:supplementary:coco_bb}
\end{table*}

\begin{table}
\centering

    \renewcommand{\arraystretch}{1.0} 
  \begin{tabular}{@{}l|l|l|l|l@{}}
    \hline
Method  & Arch & $\text{AP}$ & $\text{AP}^{50}$ & $\text{AP}^{75}$   \\
    \hline
    
    HRNet  & H-32 & $76.5$ & $93.5$ & $83.7$ \\
    MIPNet & H-32 & $\mathbf{77.44 \pm 0.185}$ & $\mathbf{94.42 \pm 0.039}$ & $\mathbf{85.32\pm 0.025}$  \\
    
    \hline
    
    HRNet & H-48 & 77.1 & 93.6 & 84.7\\
    MIPNet & H-48 & $\mathbf{77.84 \pm 0.162}$ & $\mathbf{94.44 \pm 0.079}$ & $\mathbf{85.4\pm 0.012}$ \\
    
    \hline

    \end{tabular}

    \caption{We report mean $\pm$ std-dev of MIPNet over five runs on the COCO \texttt{val} set with ground-truth bounding boxes using $256 \times 192$ input resolution. H-@ stands for HRNet-W@ backbone.}
    \label{tab:averaged_runs}

\end{table}

\subsection{Additional results on CrowdPose}

Table~\ref{tab:supplementary:crowdpose_gt} compares MIPNet to HRNet for various widths and image resolutions on CrowdPose {\tt val} dataset using ground truth bounding boxes. Similarly, Table.~\ref{tab:supplementary:crowdpose_bb} compares MIPNet to HRNet on CrowdPose {\tt val} and {\tt test} datasets using Faster-RCNN bounding boxes~\cite{ren2015faster}. Note that, commensurate with increasing percentage of occlusions in the dataset, MIPNet consistently does better than HRNet in most metrics on both datasets.

\begin{table*}[t]
    \centering
    \small
    \renewcommand{\arraystretch}{1.0} 
    \begin{tabular}{@{}l|l|l|l|l c c c c c c c c@{}}
    \hline

Method  & Arch  & Input Size & $\text{AP}$ & $\text{AP}^{50}$ & $\text{AP}^{75}$ & $\text{AR}$ & $\text{AR}^{50}$ & $\text{AR}^{75}$ & $\text{AP}^\text{easy}$ & $\text{AP}^\text{med}$ & $\text{AP}^\text{hard}$   \\
    \hline
      HRNet  & H-32   & $256 \times 192$ &
      70.0 & 91.0 & 76.3 & 73.9 & 92.6 & 79.4 & \textbf{78.8} & 70.3 & 61.7 \\
  MIPNet & H-32 & $256 \times 192$ & 
      \textbf{71.2 (+1.2)} & \textbf{91.9} & \textbf{77.4} & \textbf{76.1} & \textbf{94.4} & \textbf{81.7} & \textbf{78.8} & \textbf{71.5} & \textbf{63.8} \\
      \hline
      HRNet  & H-48   & $256 \times 192$  &  
      71.3 & 91.1 & 77.5 & 74.8 & 92.4 & 80.5 & 80.5 & 71.4 & 62.5\\
  MIPNet & H-48    & $256 \times 192$  &   
      \textbf{72.8 (+1.5)} & \textbf{92.0} & \textbf{79.2} & \textbf{77.4} & \textbf{94.8} & \textbf{83.0} & \textbf{80.6} & \textbf{73.1} & \textbf{65.2}\\
      
     \hline
      HRNet  & H-32     &  $384 \times 288$ & 
      71.6 & 91.1 & 77.7 & 75.0 & 92.6 & 80.4 & 80.4 & 72.1 & 62.6 \\
  MIPNet & H-32    &  $384 \times 288$  &  
      \textbf{73.0 (+1.4)} & \textbf{91.8} & \textbf{79.3} & \textbf{77.9} & \textbf{94.8} & \textbf{83.4} & \textbf{80.7} & \textbf{73.3} & \textbf{65.5} \\

      \hline
      HRNet  & H-48    &  $384 \times 288$ &  
      72.8 & \textbf{92.1} & 78.7 & 76.3 & 93.3 & 81.4 & \textbf{81.3} & 73.3 & 64.0 \\
  MIPNet & H-48    &  $384 \times 288$ &  
      \textbf{73.7 (+0.9)} & 91.9 & \textbf{80.0} & \textbf{78.4} & \textbf{94.8} & \textbf{84.0} & 80.7 & \textbf{74.1} & \textbf{66.5}\\
\hline
  \end{tabular}

    \caption{Additional metrics for comparison between MIPNet and various architectures on CrowdPose~\texttt{val} set using ground-truth bounding boxes for evaluation. }
    \label{tab:supplementary:crowdpose_gt}
\end{table*}

\begin{table*}[t]
    \centering
    \small
    \renewcommand{\arraystretch}{1.0} 
    \begin{tabular}{@{}l|c|c|l|l c c c c c c c c@{}}
    \hline

Method  & Arch  & Input Size & $\text{AP}$ & $\text{AP}^{50}$ & $\text{AP}^{75}$ & $\text{AR}$ & $\text{AR}^{50}$ & $\text{AR}^{75}$ & $\text{AP}^\text{easy}$ & $\text{AP}^\text{med}$ & $\text{AP}^\text{hard}$   \\
    \hline
    \multicolumn{12}{c}{\texttt{val}} \\
     \hline 
      HRNet  & H-48    &  $384 \times 288$ &
      68.0 & 85.5 & 73.4 & 76.6 & 93.8 & 81.9 & \textbf{77.4} & 68.8 & 57.8 \\
  MIPNet & H-48    &  $384 \times 288$ &
      \textbf{68.8 (+0.8)} & \textbf{85.9} & \textbf{74.5} & \textbf{78.1} & \textbf{94.5} & \textbf{83.6} & 77.1 & \textbf{69.4} & \textbf{59.8} \\
    \hline
    \multicolumn{12}{c}{\texttt{test}} \\
    \hline
    \multicolumn{12}{c}{\scriptsize{Bottom-Up}} \\
    \hline
     OpenPose~\cite{cao2018openpose}  & -    &  - &
     - & - & - & - & - & - & 62.7 & 48.7 & 32.3 \\ 
    HigherHRNet~\cite{cheng2019higherhrnet}  & HH-48  &  $640 \times 640$ &
     67.6 & 87.4 & 72.6 & - & - & - & 75.8 & 68.1 & 58.9 \\ 
    HghHRNet + UDP~\cite{huang2020devil}  & HH-48  &  $640 \times 640$ &
     68.2 & 88.0 & 72.9 & - & - & - & 76.6 & 68.7 & 59.9 \\ 
    
    \hline
    \multicolumn{12}{c}{\scriptsize{Top-Down, YOLO-v3}} \\
    \hline
    MaskRCNN~\cite{he2017mask}  & R-101    &  - &
     57.2 & 83.5 & 60.3 & - & - & - & - & - & - \\ 
    SimpleBaseline~\cite{xiao2018simple}  & R-101    &  - &
     60.8 & 81.4 & 65.7 & - & - & - & - & - & - \\ 
    AlphaPose+~\cite{qiu2020peeking}  & R-101    &  - &
     27.5 & 40.8 & 29.9 & - & - & - & - & - & - \\ 
    OPEC-Net~\cite{qiu2020peeking}  & R-101    &  - &
     \textbf{70.6} & 86.8 & 75.6 & - & - & - & - & - & - \\ 
    MIPNet  & R-101  &  $384 \times 288$ &
     68.1 & 85.2 & 73.8 & 75.1 & 92.3 & 79.2 & 74.6 & 69.2 & 53.4 \\ 
       \hline
    \multicolumn{12}{c}{\scriptsize{Top-Down, Faster-RCNN}} \\
    \hline
    HRNet  & H-48    &  $384 \times 288$ &
     69.3 & \textbf{86.9} & 74.7 & 77.3 & 94.2 & 82.5 & 77.7 & 70.6 & 57.8 \\
  MIPNet & H-48    &  $384 \times 288$ &
      70.0 (+0.7) & 86.8 & \textbf{75.7} & \textbf{78.8} & \textbf{94.9} & \textbf{84.3} & \textbf{78.1} & \textbf{71.1} & \textbf{59.4} \\

\hline
  \end{tabular}

    \caption{Additional metrics for comparison between MIPNet and various architectures on CrowdPose \texttt{val} and \texttt{test} set using Faster-RCNN and YOLO-v3 bounding boxes for evaluation.}
    \label{tab:supplementary:crowdpose_bb}
\end{table*}

\subsection{Additional results on OCHuman}

Table~\ref{tab:supplementary:ochuman_gt} shows our detailed evaluations on the OCHuman {\tt val} dataset using ground truth bounding boxes. As can be seen, MIPNet outpeforms HRNet and SimpleBaseline on {\bf all} metrics, with a maximum improvement of $10.5~\text{AP}$ over SimpleBaseline ($\text{R}-50, 384 \times 288$) and $9.1~\text{AP}$ over HRNet ($\text{H}-48, 384\times 288$).

\begin{table*}[t]
    \centering
    \small
    \renewcommand{\arraystretch}{1.0} 
    \begin{tabular}{@{}l|l|l|l|l c c c c c c c c c@{}}
    \hline
Method  & Arch  & Input Size  & $\text{AP}$ & $\text{AP}^{50}$ & $\text{AP}^{75}$ & $\text{AP}^\text{M}$ & $\text{AP}^\text{L}$ & $\text{AR}$ & $\text{AR}^{50}$ & $\text{AR}^{75}$ & $\text{AR}^\text{M}$ & $\text{AR}^\text{L}$  \\
    \hline
      SimpleBaseline  & R-50    & $256 \times 192$ &   
      56.3 & 76.1 & 61.2 & 66.4 & 56.3 & 61.0 & 78.0 & 65.9 & 70.0 & 61.0\\
  MIPNet & R-50    &  $256 \times 192$ & 
      \textbf{64.4 (+8.1)} & \textbf{86.0} & \textbf{70.4} & \textbf{66.8} & \textbf{64.6} & \textbf{72.3} & \textbf{91.5} & \textbf{78.5} & \textbf{71.4} & \textbf{72.3}\\
      \hline
      SimpleBaseline  & R-101    &   $256 \times 192$ &
      60.5 & 77.2 & 66.6 & 68.3 & 60.5 & 64.7 & 79.6 & 70.1 & 72.9 & 64.7\\
  MIPNet & R-101     & $256 \times 192$ &
      \textbf{68.2 (+7.7)} & \textbf{87.4} & \textbf{75.1} & \textbf{67.0} & \textbf{68.2} & \textbf{75.5} & \textbf{92.9} & \textbf{82.1} & \textbf{74.3} & \textbf{75.5}\\
      \hline
      
      SimpleBaseline  & R-152    &   $256 \times 192$ &
      62.4 & 78.3 & 68.1 & \textbf{68.3} & 62.4 & 66.5 & 80.2 & 71.8 & \textbf{74.3} & 66.5\\
  MIPNet & R-152  & $256 \times 192$ &
      \textbf{70.3 (+7.9)} & \textbf{88.6} & \textbf{77.9} & 66.9 & \textbf{70.2} & \textbf{77.0} & \textbf{93.0} & \textbf{84.1} & 72.9 & \textbf{77.0}\\

    \hline
      SimpleBaseline  & R-50   &  $384 \times 288$ &
      55.8 & 74.8 & 60.4 & 64.7 & 55.9 & 60.7 & 78.0 & 65.2 & \textbf{71.4} & 60.7\\
  MIPNet & R-50    &   $384 \times 288$ &
      \textbf{66.3 (+10.5)} & \textbf{87.5} & \textbf{72.2} & \textbf{66.0} & \textbf{66.3} & \textbf{74.1} & \textbf{92.7} & \textbf{80.3} & \textbf{71.4} & \textbf{74.1}\\
      \hline
     SimpleBaseline & R-101    &  $384 \times 288$ &
      61.6 & 77.2 & 66.6 & 62.1 & 61.6 & 65.8 & 79.4 & 70.5 & \textbf{72.9} & 65.8\\
  MIPNet & R-101     &   $384 \times 288$ &
      \textbf{70.3 (+8.7)} & \textbf{88.4} & \textbf{77.1} & \textbf{64.1} & \textbf{70.4} & \textbf{77.7} & \textbf{93.4} & \textbf{84.0} & \textbf{72.9} & \textbf{77.7}\\
      \hline
      
      SimpleBaseline  & R-152    &   $384 \times 288$ &
      64.2 & 78.3 & 69.1 & 66.5 & 64.2 & 68.1 & 80.4 & 73.0 & \textbf{74.3} & 68.1\\
  MIPNet & R-152 & $384 \times 288$ &
      \textbf{72.4 (+8.2)} & \textbf{89.5} & \textbf{79.5} & \textbf{67.7} & \textbf{72.5} & \textbf{79.6} & \textbf{94.1} & \textbf{86.2} & 71.4 & \textbf{79.6}\\
    \hline
    \hline
      HRNet  & H-32   & $256 \times 192$ &
      63.1 & 79.4 & 69.0 & 64.2 & 63.1 & 67.3 & 81.9 & 72.4 & 68.6 & 67.3\\
  MIPNet & H-32 & $256 \times 192$ &
      \textbf{72.5 (+9.4)} & \textbf{89.2} & \textbf{79.4} & \textbf{65.1} & \textbf{72.6} & \textbf{79.1} & \textbf{93.6} & \textbf{85.2} & \textbf{71.4} & \textbf{79.1}\\
      
      \hline
      HRNet  & H-48    &  $256 \times 192$ &
      64.5 & 79.4 & 70.1 & 65.1 & 64.5 & 68.5 & 81.6 & 73.7 & 68.6 & 68.5\\
  MIPNet & H-48     &   $256 \times 192$ &
      \textbf{72.2 (+7.7)} & \textbf{89.5} & \textbf{78.7} & \textbf{66.5} & \textbf{72.3} & \textbf{79.2} & \textbf{94.2} & \textbf{85.4} & \textbf{70.0} & \textbf{79.2}\\
     \hline
      HRNet  & H-32     & $384 \times 288$ &
      63.7 & 78.4 & 69.0 & 64.3 & 63.7 & 67.6 & 80.8 & 72.6 & \textbf{70.0} & 67.6\\
  MIPNet & H-32     &  $384 \times 288$ &
      \textbf{72.7 (+9.0)} & \textbf{89.6} & \textbf{79.6} & \textbf{66.5} & \textbf{72.7} & \textbf{79.7} & \textbf{94.3} & \textbf{86.1} & \textbf{70.0} & \textbf{79.7}\\
      \hline
      HRNet  & H-48    &  $384 \times 288$ &
      65.0 & 78.4 & 70.3 & 68.4 & 65.0 & 68.8 & 80.6 & 73.4 & 71.4 & 68.8\\
  MIPNet & H-48    &  $384 \times 288$ &
      \textbf{74.1 (+9.1)} & \textbf{89.7} & \textbf{80.1} & \textbf{68.4} & \textbf{74.1} & \textbf{81.0} & \textbf{94.4} & \textbf{87.0} & \textbf{72.9} & \textbf{81.0}\\
\hline

  \end{tabular}

    \caption{Additional metrics for comparison between MIPNet and various architectures on OCHuman~\texttt{val} set using ground-truth bounding boxes for evaluation.}
    \label{tab:supplementary:ochuman_gt}
\end{table*}

Similarly, Table.~\ref{tab:supplementary:ochuman_bb} shows detailed results on the OCHuman {\tt val} and {\tt test} datasets using Faster-RCNN bounding boxes. MIPNet achieves a {\bf state-of-the-art} $42.5 \text{AP}$ across \emph{both} top-down and bottom-up pose estimation networks, to the best of our knowledge. We show a $4.2~\text{AP}$ improvement over HRNet on {\tt val} dataset and a $5.3~\text{AP}$ improvement over HRNet on {\tt test} dataset in this case.

\begin{table*}[t]
    \centering
    \small
    \renewcommand{\arraystretch}{1.0} 
    \begin{tabular}{@{}l|c|c|l|l c c c c c c c c c@{}}
    \hline

Method  & Arch  & Input Size  & $\text{AP}$ & $\text{AP}^{50}$ & $\text{AP}^{75}$ & $\text{AP}^\text{M}$ & $\text{AP}^\text{L}$ & $\text{AR}$ & $\text{AR}^{50}$ & $\text{AR}^{75}$ & $\text{AR}^\text{M}$ & $\text{AR}^\text{L}$  \\
    \hline
    \multicolumn{13}{c}{\texttt{val}} \\
     \hline 
      HRNet  & H-48    &  $384 \times 288$ &
     37.8 & 50.6 & 40.5 & \textbf{3.8} & 40.4 & 69.9 & 89.0 & 73.9 & 67.1 & 69.9  \\
  MIPNet & H-48    &  $384 \times 288$ &
      \textbf{42.0 (+4.2)} & \textbf{51.2} & \textbf{45.6} & 3.2 & \textbf{43.5} & \textbf{82.5} & \textbf{96.7} & \textbf{88.5} & \textbf{71.4} & \textbf{82.5} \\
       \hline
    \multicolumn{13}{c}{\texttt{test}} \\
    \hline
    \multicolumn{13}{c}{\scriptsize{Bottom-Up}} \\
    \hline
    AE~\cite{newell2017associative}  & Hourglass   &  -  &
     29.5 & - & - & - & - & - & - & - & - & -  \\
    AE-multiscale~\cite{newell2017associative}  & Hourglass   &  -  &
     32.8 & - & - & - & - & - & - & - & - & -  \\
    HGG~\cite{jin2020differentiable}  & Hourglass   &  -  &
     34.8 & - & - & - & - & - & - & - & - & -  \\ 
    HGG-multiscale~\cite{jin2020differentiable}  & Hourglass   &  -  &
     36.0 & - & - & - & - & - & - & - & - & -  \\
    \hline
    \multicolumn{13}{c}{\scriptsize{Top-Down, YOLO-v3}} \\
    \hline
    MaskRCNN~\cite{he2017mask}  & R-101    &  -  &
     20.2 & 33.2 & 24.5 & - & - & - & - & - & - & -  \\
    SimpleBaseline  & R-101    &  -  &
     24.1 & 37.4 & 26.8 & - & - & - & - & - & - & -  \\
    
    AlphaPose+~\cite{qiu2020peeking}  & R-101    &  -  &
     27.5 & 40.8 & 29.9 & - & - & - & - & - & - & -  \\
    
    OPEC-Net~\cite{qiu2020peeking}  & R-101    &  -  &
     29.1 & 41.3 & 31.4 & - & - & - & - & - & - & -  \\
    MIPNet  & R-101    &  $384 \times 288$  &
     35.0 & 44.1 & 36.1 & - & 35.1 & 74.5 & 88.6 & 79.1 & - & 72.8  \\
    \hline
    \multicolumn{13}{c}{\scriptsize{Top-Down, FasterRCNN}} \\
    \hline
    HRNet  & H-48    &  $384 \times 288$ &
     37.2 & 46.7 & 40.0 & - & 39.8 & 78.0 & 93.5 & 83.7 & - & 78.0  \\
  MIPNet & H-48    &  $384 \times 288$ &
      \textbf{42.5 (+5.3)} & \textbf{51.8} & \textbf{46.3} & - & \textbf{44.1} & \textbf{83.0} & \textbf{97.1} & \textbf{89.2} & - & \textbf{83.0} \\

\hline
  \end{tabular}

    \caption{Additional metrics for comparison between MIPNet and various architectures on OCHuman \texttt{val} and \texttt{test} set using Faster-RCNN and YOLO-v3 bounding boxes for evaluation.}
    \label{tab:supplementary:ochuman_bb}
\end{table*}

\subsection{Robustness to Bounding Box Confidence}

\begin{table}
    \centering
    \renewcommand{\arraystretch}{1.0} 
    \begin{tabular}{@{}c | r| r@{}}
    \hline
    Min. BB   & \multicolumn{2}{c}{OCHuman} \\
    Confid. & \texttt{val} & \texttt{test} \\
    \hline
    0.0      &  30637   &   26992   \\
    0.1      &  22247   &   19704   \\
    0.2      &  16273   &   14613   \\
    0.3      & 13603   &   12216   \\
    0.4      & 11944   &   10767   \\
    0.5      & 10654   &   9645   \\
    0.6      &  9626   &   8697   \\
    0.7      & 8699   &   7880   \\
    0.8      &  7768   &   7018   \\
    0.9      & 6644   &   5989   \\      
    0.99     & 4416   &   3883   \\
    \hline
    \end{tabular}
    \caption{Number of Faster-RCNN bounding boxes greater than a given confidence score.}
    \label{tab:supplementary:num_bbs}
\end{table}


Table~\ref{tab:supplementary:num_bbs} illustrates the number of Faster-RCNN bounding boxes as a function of minimum bounding box confidence. Notice that a majority of all available bounding boxes  (min. confidence = $0.0$) have confidence $< 0.4$.

\begin{figure}
\centering
\includegraphics[width=0.5\linewidth]{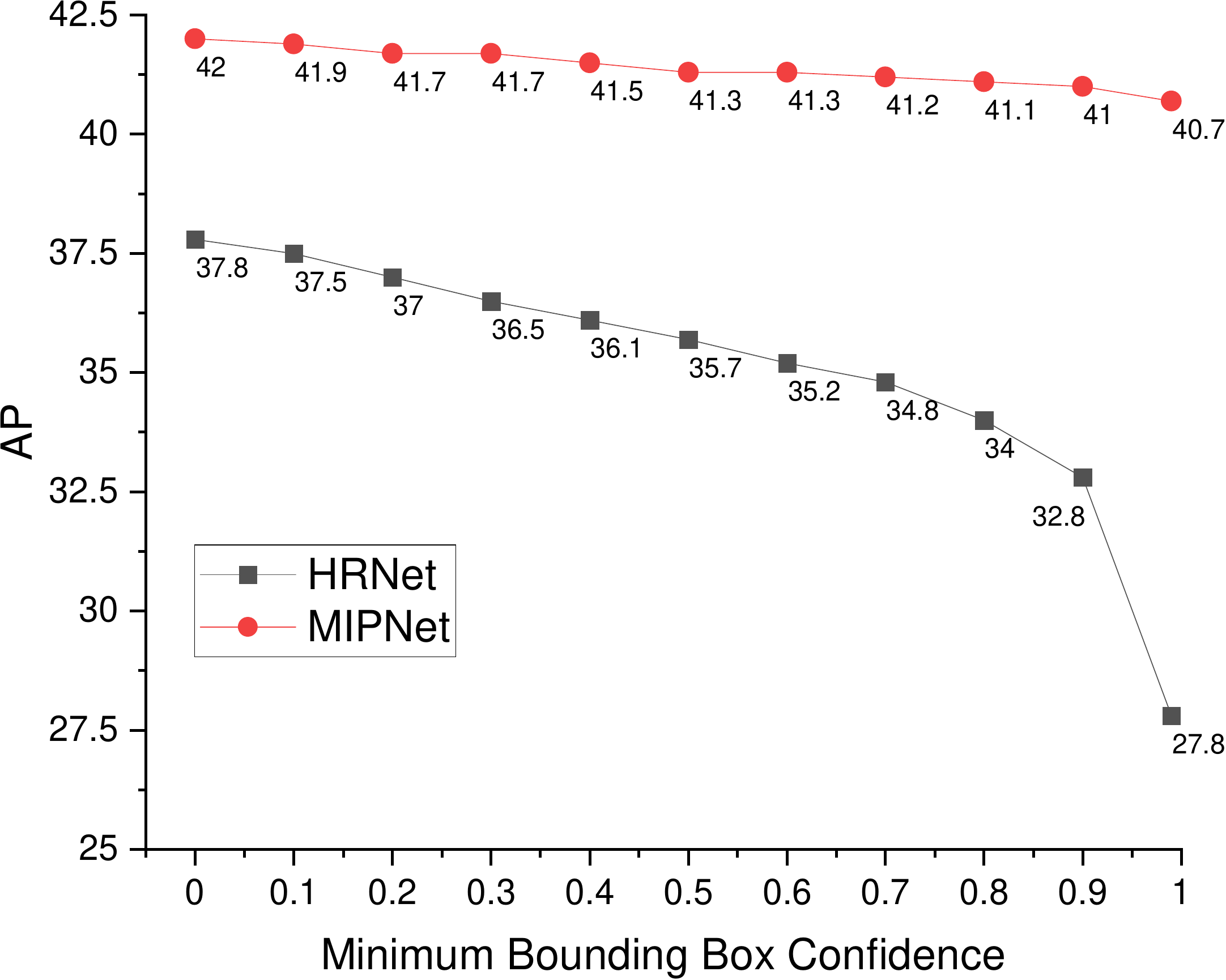}
\caption{Unlike HRNet, MIPNet maintains a stable performance as a function of detector confidence for selecting input bounding boxes. Results are shown using HRNet-W48-$384\times288$ evaluated on the \texttt{val} set of OCHuman.}
\label{fig:supplementary:bb_decay_val}
\end{figure}

\begin{figure}
\centering
\includegraphics[width=0.5\linewidth]{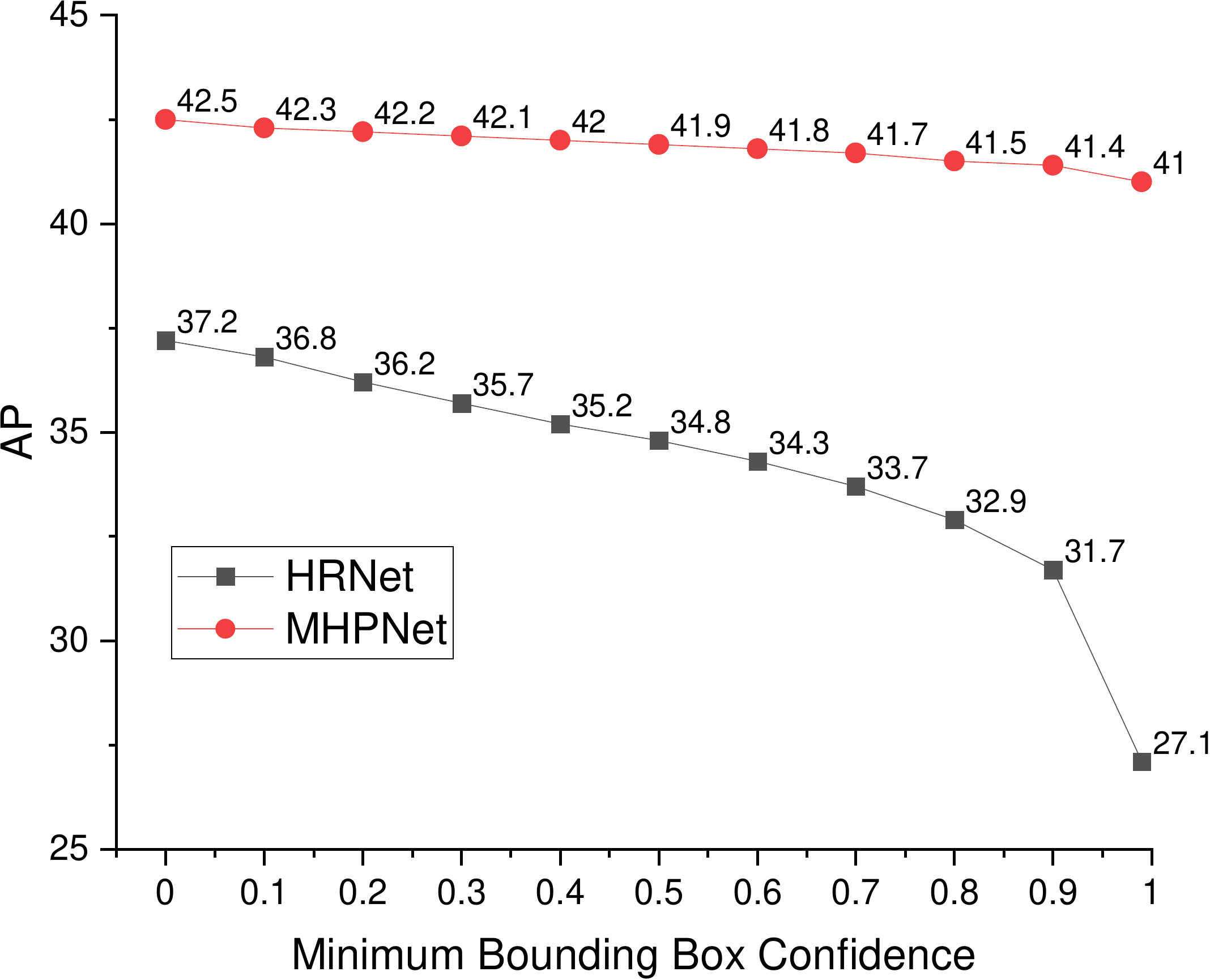}
\caption{Similar to Figure~\ref{fig:supplementary:bb_decay_val} we show results on the \texttt{test} set of OCHuman.}
\label{fig:supplementary:bb_decay_test}
\end{figure}

We compare the performance of MIPNet to HRNet as a function of varying minimum confidence on OCHuman {\tt test} dataset in Fig.~\ref{fig:supplementary:bb_decay_test} and {\tt val} dataset in  Fig.~\ref{fig:supplementary:bb_decay_val} (also shown in the paper). MIPNet is much more stable w.r.t bounding box confidence thresholding, as compared to baseline networks like HRNet. We note that while MIPNet AP drops from $42.5$ to $41.4$ ($1.1~\text{AP}$ drop) on {\tt test} set at minimum confidence of $0.9$, HRNet drops by more than $6~\text{AP}$. This performance is consistent with the performance on the {\tt val} dataset (Fig. 4 in the paper).

\section{Individual Instance Performance}

\begin{table}[t]
    \centering
    \small
    \renewcommand{\arraystretch}{1.0} 
    \begin{tabular}{@{}l|c|c|c@{}}
    \hline

Inference  & COCO  & CrowdPose & OCHuman    \\
    \hline
    HRNet  & 78.1 & 72.8 & 65.0 \\
    \hline
    MIPNet (SIP, $\lambda=1$) & 55.8  & 42.2  & 41.4 \\
    MIPNet (SIP, $\lambda=0$) & 78.3  & 72.7  & 65.7 \\
MIPNet (MIP) & \textbf{78.8}  & \textbf{73.7}  & \textbf{74.1} \\
\hline
  \end{tabular}

    \caption{Performance of each individual instances of MIPNet on \texttt{val} sets using ground truth bounding boxes. We use the backbone W48 with image resolution $384 \times 288$, and compare with the same HRNet configuration. By default, HRNet only predicts a single instance.}
    \label{tab:supplementary:individual_hypothesis}
\end{table}

It is interesting to compare the performance of each individual instances predicted by MIPNet in isolation. Since $\lambda=0$ correspond to the primary instance (centered on the person), only using the primary instance for inference is expected to give better results compared to only using $\lambda=1$ instance during inference. In addition, we also expect $\lambda=0$ instance to provide similar performance as baseline top-down network, if used in isolation. Table~\ref{tab:supplementary:individual_hypothesis} shows the performance of each individual instance mode of MIPNet with HRNet-W48 backbone at input size $384 \times 288$ on various datasets, using ground truth bounding boxes. Note that when using only a single hypothesis from MIPNet for inference, performance of primary instance ($\lambda=0$) is similar to HRNet. When using multiple instances during inference, we get an improvement of $8.4$ AP ($65.7$ to $74.1$ AP) on the OCHuman dataset.

\section{Ablation: MIMB}

\begin{table*}
\centering
\renewcommand{\arraystretch}{1.1} 
    \begin{tabular}{@{}l|l|l|c c c c c c c c c c@{}}
    \hline
    Method & Arch & Ablation & $\text{AP}$ & $\text{AP}^{50}$ & $\text{AP}^{75}$ & $\text{AP}^\text{M}$ & $\text{AP}^\text{L}$ & $\text{AR}$ & $\text{AR}^{50}$ & $\text{AR}^{75}$ &$\text{AR}^\text{M}$ & $\text{AR}^\text{L}$ \\
    \hline
    & & & \multicolumn{9}{c}{COCO} & \\
    \hline
    MIPNet & H-48 & only embed & 78.5 & 94.4 & 85.5 & 75.3 & 83.5 & 81.4 & 95.8 & 87.5 & 77.8 & 86.7 \\
    MIPNet & H-48 & reduce=$1$ & 78.8 & 94.4 & 85.8 & 75.5 & 83.6 & 81.5 & 95.4 & 87.8 & 78.0 & 86.6 \\
    MIPNet & H-48 & reduce=$2$ & 78.8 & 94.4 & 85.6 & 75.8 & 83.6 & 81.7 & 95.7 & 87.7 & 78.3 & 86.8 \\
    MIPNet & H-48    &  reduce=$4$ &
      \textbf{78.8} & \textbf{94.4} & \textbf{85.7} & \textbf{75.5} & \textbf{83.7} & \textbf{81.6} & \textbf{95.5} & \textbf{87.5} & \textbf{78.0} & \textbf{86.8} \\
    \hline
    & & & \multicolumn{9}{c}{OCHuman} & \\
    \hline
    MIPNet & H-48 & only embed & 70.8 & 89.8 & 77.5 & 65.7 & 70.9 & 77.9 & 94.2 & 84.2 & 68.6 & 77.9 \\
    MIPNet & H-48 & reduce=$1$ & 74.4 & 90.7 & 80.9 & 66.9 & 74.4 & 81.2 & 95.1 & 87.2 & 70.0 & 81.2 \\
    MIPNet & H-48 & reduce=$2$ & 74.0 & 90.1 & 80.3 & 63.6 & 74.0 & 80.7 & 94.5 & 86.7 & 68.6 & 80.7 \\
    MIPNet & H-48    &  reduce=$4$ &
      \textbf{74.1 } & \textbf{89.7} & \textbf{80.1} & \textbf{68.4} & \textbf{74.1} & \textbf{81.0} & \textbf{94.4} & \textbf{87.0} & \textbf{72.9} & \textbf{81.0} \\
    \hline
    \end{tabular}
    \caption{We illustrate different ablations of MIMB. For MIPNet with backbone W48 on resolution $384 \times 288$, we train models with varying capacity for \emph{squeeze} $\mathbf{F}_{sq}$ and \emph{excite} $\mathbf{F}_{ex}$ operations. When both operations are disabled, and only \emph{embed} operation $\mathbf{F}_{embed}$ is used within MIMB, we get sub-optimal results on both COCO {\tt val} ($0.3~\text{AP}$ drop) and OCHuman {\tt val} ($3.6~\text{AP}$ drop) datasets (first row of each dataset). When \emph{squeeze} and \emph{excite} operations are employed, we get a good performance boost, especially on the OCHuman {\tt val} dataset. All results in the paper employ {\tt reduce=$4$} (bold).}
    \label{tab:supplementary:attention}
\end{table*}
\label{para:attention_variation}
In this section, we study the effect of ablation for MIMB. As outlined in the paper, MIMB consists of three operations \emph{squeeze} $\mathbf{F}_{sq}$, \emph{excite} $\mathbf{F}_{ex}$ and \emph{embed} $\mathbf{F}_{em}$. Of the three operations, the \emph{embed} operation $\mathbf{F}_{em}$ consumes the $\lambda$ parameter that we pass as additional input to MIMB. In Tab.~\ref{tab:supplementary:attention}, we show the effect of only using the embed block by disabling $\mathbf{F}_{sq}$ and $\mathbf{F}_{ex}$, in the first row for both COCO and OCHuman {\tt val} datasets. Note that these numbers are lower than corresponding experiments that use $\mathbf{F}_{sq}$ and $\mathbf{F}_{ex}$ operations, by $0.3~\text{AP}$ for COCO (Tab. 2, last row in paper) and $3.3~\text{AP}$ (Tab. 4, last row in paper) for OCHuman {\tt val} datasets. This confirms that all three operations contribute to MIMB, and therefore to MIPNet. We further study the effect of varying the intermediate linear layer within $\mathbf{F}_{sq}$ and $\mathbf{F}_{ex}$, which is controlled by the {\tt reduce} 
parameter~\cite{hu2018squeeze} in Listing~\ref{code:mimb}. While all the results reported in the paper use the default value of {\tt reduce=$4$}, we show that {\tt reduce=$2$} and {\tt reduce=$1$} show comparable results.

\section{OCPose Dataset}
For completeness, we also benchmark MIPNet on another occlusion specific OCPose dataset~\cite{qiu2020peeking}. OCPose is a larger dataset than OCHuman with pose annotations of occluded humans. It contains $9K$ images and $18000$ persons labeled with $12$ keypoints. The number of examples with occlusion IoU $>0.5$ is $78\%$ for OCPose~\cite{qiu2020peeking}. Each image in the dataset, is annotated with exactly \emph{two} person keypoints. Further, both the persons have the same bounding box, this is in contrast to tight fitting bounding box annotations in the datasets like COCO, Crowdpose and OCHuman. This results in inflated occlusion levels for the OCPose dataset in comparison to the OCHuman dataset  reported in ~\cite{qiu2020peeking} (refer its Table 1). 

Table. \ref{tab:supplementary:ocpose} reports the MIPNet's results on the OCPose dataset~\cite{qiu2020peeking} with custom \texttt{train}:\texttt{test} splits as the OPEC-Net~\cite{qiu2020peeking} splits are not released. All the models are trained on the COCO dataset and evaluated on the \texttt{test} set of OCPose. We evaluate only on the common keypoints between the both datasets.

\begin{table*}
    \centering
    \small
    \renewcommand{\arraystretch}{1.0} 
    \begin{tabular}{@{}l|l|l c c c c c c c c c@{}}
    \hline


Method  & $\text{AP}$ & $\text{AP}^{50}$ & $\text{AP}^{75}$ & $\text{AP}^\text{M}$ & $\text{AP}^\text{L}$ & $\text{AR}$ & $\text{AR}^{50}$ & $\text{AR}^{75}$ & $\text{AR}^\text{M}$ & $\text{AR}^\text{L}$  \\
    \hline
    \multicolumn{11}{c}{\scriptsize{ground truth bounding box}} \\
     \hline 
    HRNet  &
      34.2 & 48.2 & 36.7 & 36.6 & 34.1 & 36.8 & 48.9 & 39.5 & 38.3 & 36.8 \\
    MIPNet ($\lambda=0$) & 
      34.6 & 49.2 & 36.7 & 37.0 & 34.6 & 37.3 & 49.2 & 39.9 & 36.7 & 37.3 \\
    MIPNet ($\lambda=1$) & 
      23.8 & 34.9 & 25.2 & 30.6 & 24.0 & 28.6 & 39.7 & 30.0 & 45.0 & 28.6 \\
    MIPNet & 
      \textbf{49.7 (+15.5)} & \textbf{72.3} & \textbf{53.0} & \textbf{59.4} & \textbf{49.7} & \textbf{56.4} & \textbf{74.8} & \textbf{60.1} & \textbf{70.0} & \textbf{56.4}\\
      
       \hline
    \multicolumn{11}{c}{\scriptsize{ground truth bounding box - tight fitting}} \\
    \hline
    HRNet  & 
    47.7 & 74.6 & 50.1 & 35.8 & 47.7 & 53.0 & 77.0 & 56.4 & 41.3 & 53.1\\
    MIPNet ($\lambda=0$) & 
      46.6 & 73.2 & 49.1 & 33.7 & 46.9 & 52.5 & 76.2 & 55.8 & 37.0 & 52.6 \\
    MIPNet ($\lambda=1$) & 
      26.5 & 51.2 & 23.9 & 10.7 & 26.9 & 36.4 & 61.4 & 35.7 & 32.3 & 36.4 \\
    MIPNet & 
      \textbf{49.3 (+1.6)} & \textbf{77.3} & \textbf{51.9} & \textbf{33.6} & \textbf{49.5} & \textbf{56.9} & \textbf{82.8} & \textbf{60.7} & \textbf{37.3} & \textbf{57.1}\\
 
\hline
  \end{tabular}

    \caption{Results on the OCPose \texttt{val} set. All the evaluations use the HRNet-W48 backbone at $348 \times 288$ image resolution. We provide both evaluations, using the relaxed gt bounding boxes provided by the OCPose and the tight fitting gt bounding box. The tight fitting bounding box is using the keypoint annotations.}
    \label{tab:supplementary:ocpose}
\end{table*}

\section{Qualitative Results}
\begin{figure*}
 \begin{center}
\includegraphics[height=0.13\textheight,width=0.3\linewidth]{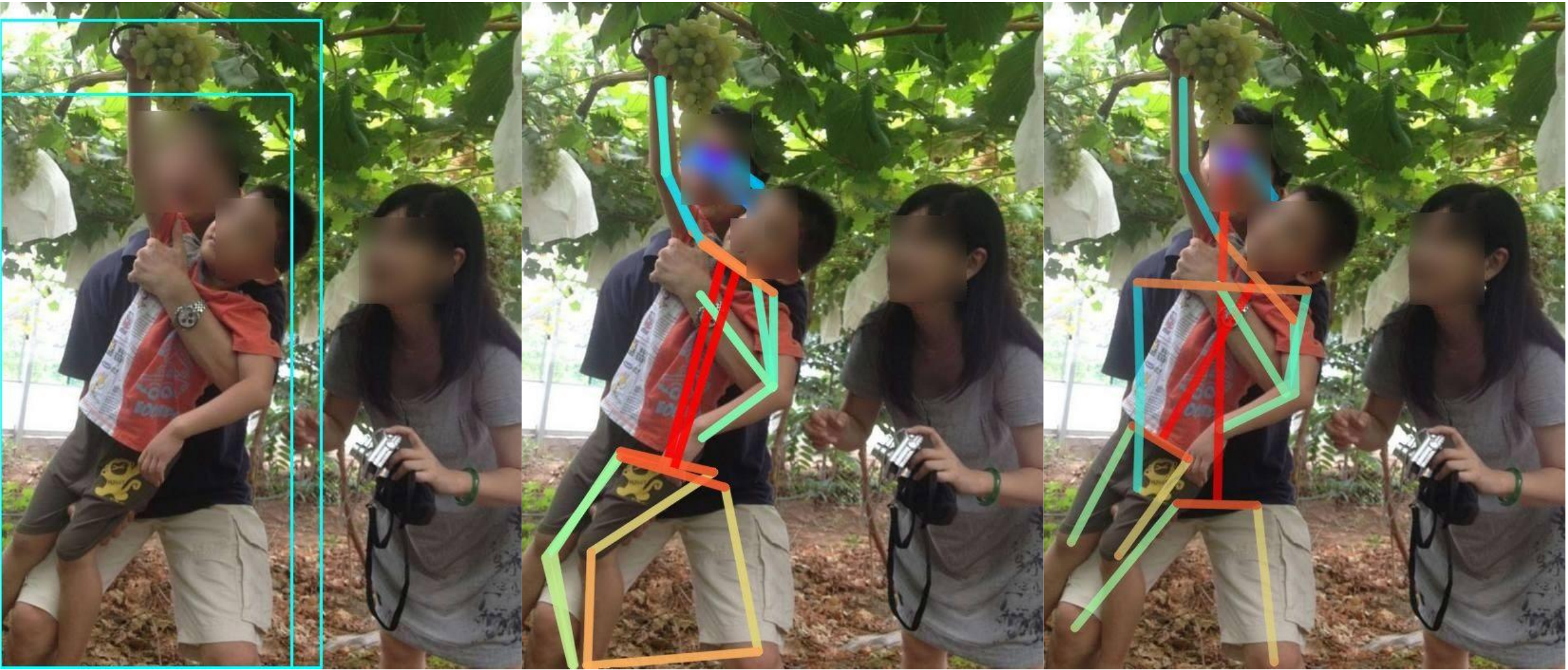}
\includegraphics[height=0.13\textheight,width=0.3\linewidth]{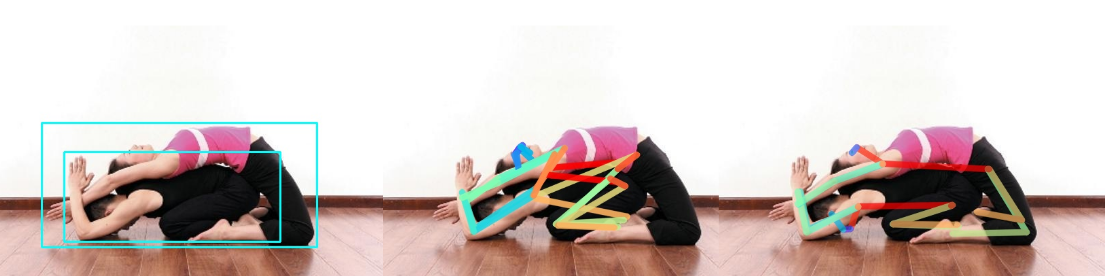}
\includegraphics[height=0.13\textheight,width=0.3\linewidth]{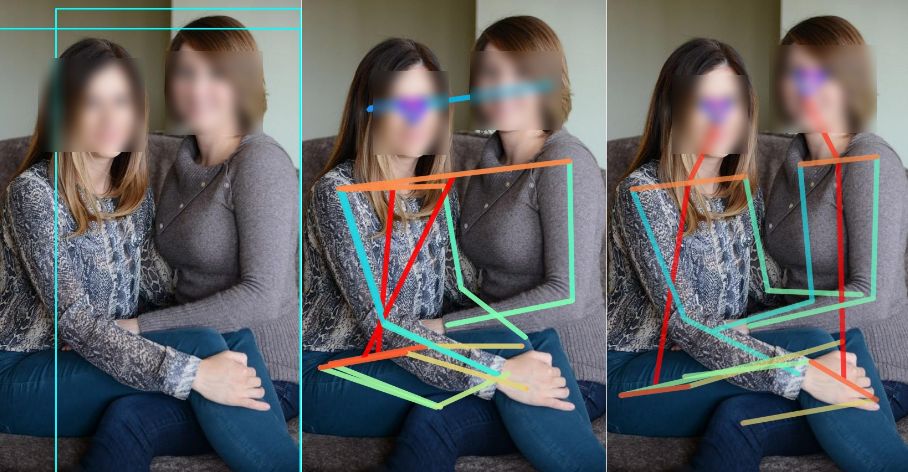}

\includegraphics[height=0.13\textheight,width=0.3\linewidth]{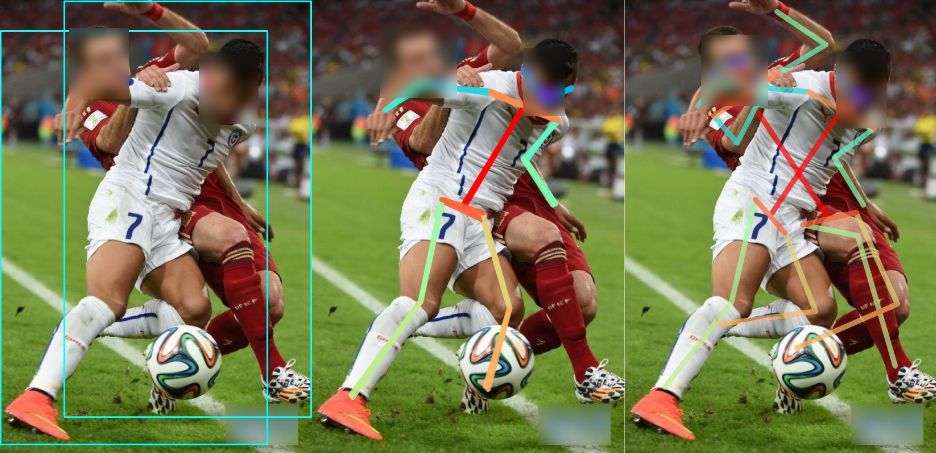}
\includegraphics[height=0.13\textheight,width=0.3\linewidth]{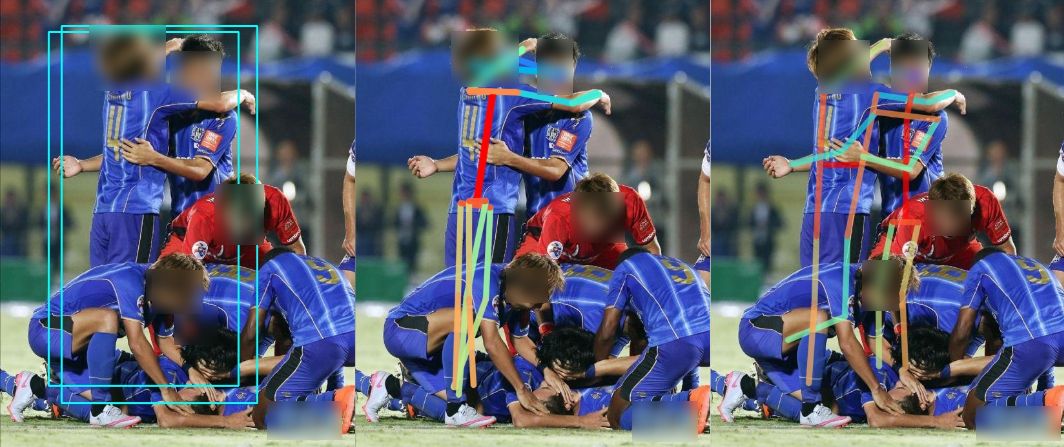}
\includegraphics[height=0.13\textheight,width=0.3\linewidth]{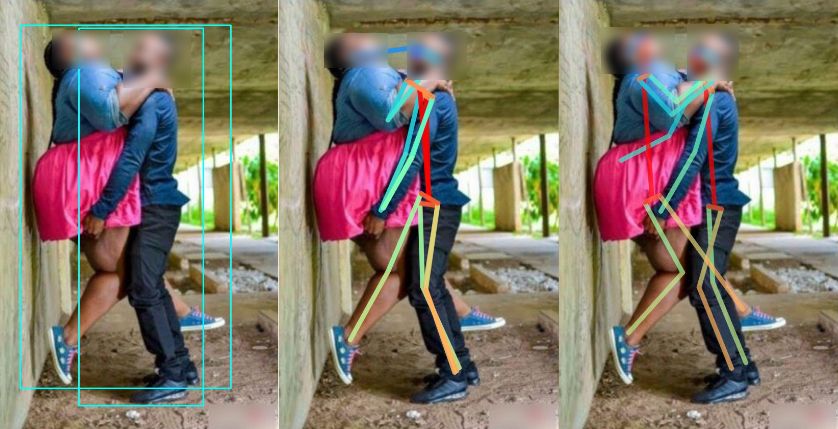}

\includegraphics[height=0.13\textheight,width=0.3\linewidth]{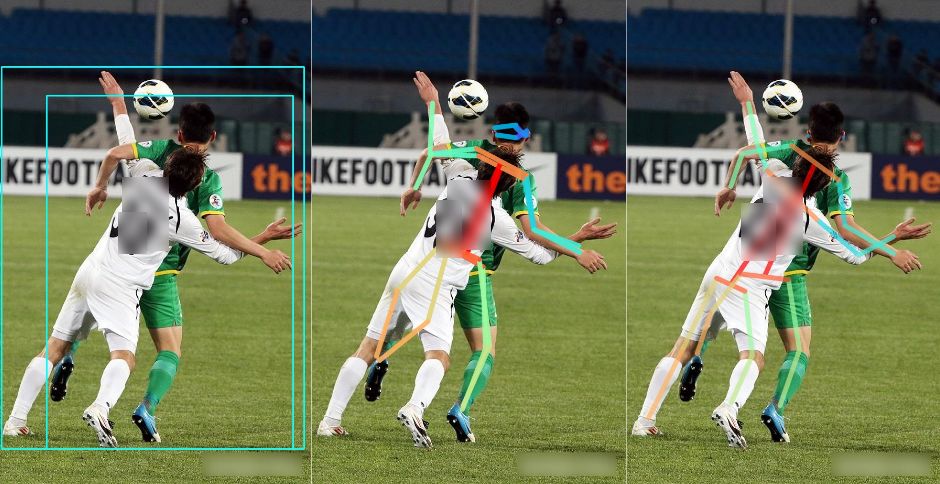}
\includegraphics[height=0.13\textheight,width=0.3\linewidth]{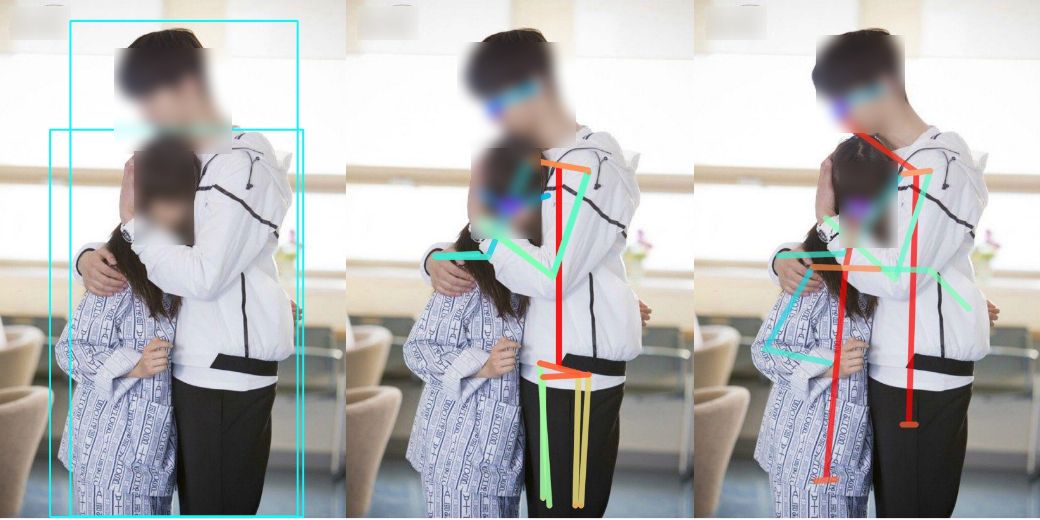}
\includegraphics[height=0.13\textheight,width=0.3\linewidth]{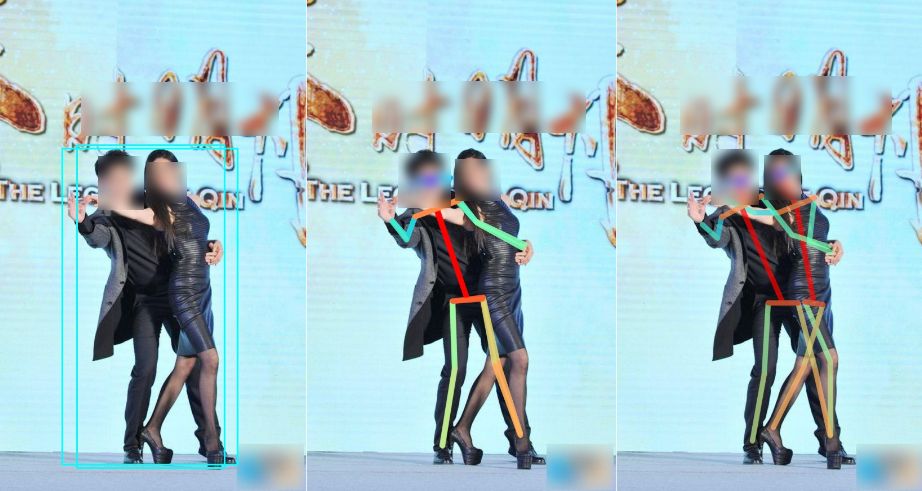}

\includegraphics[height=0.13\textheight,width=0.3\linewidth]{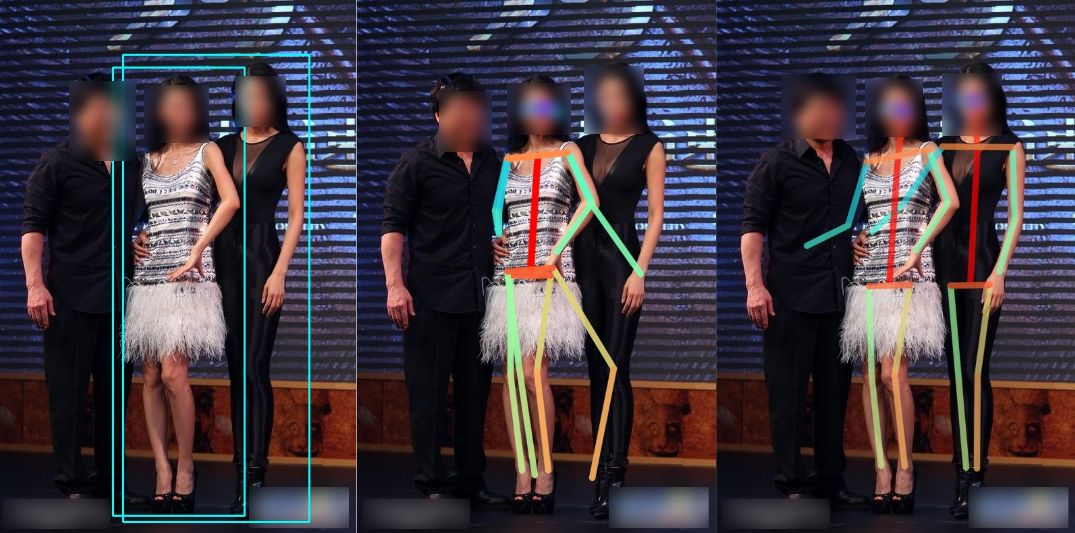}
\includegraphics[height=0.13\textheight,width=0.3\linewidth]{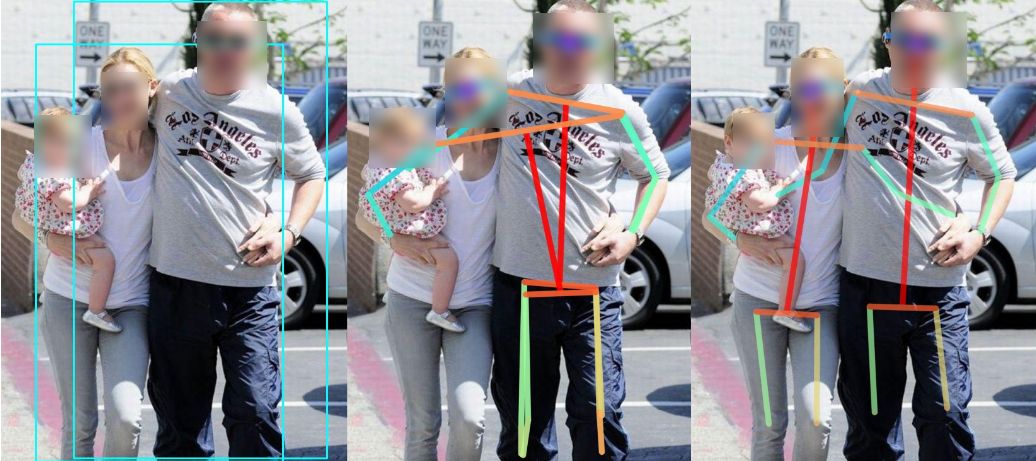}
\includegraphics[height=0.13\textheight,width=0.3\linewidth]{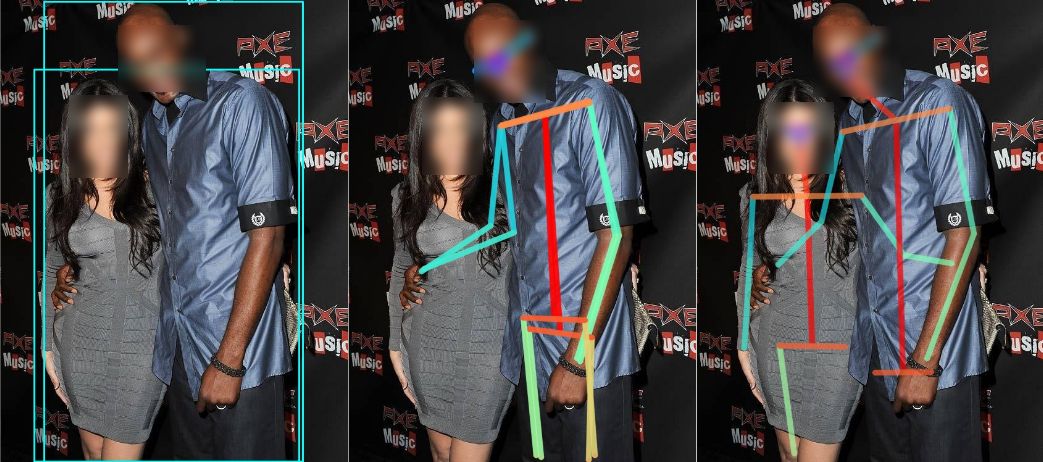}

\includegraphics[height=0.13\textheight,width=0.3\linewidth]{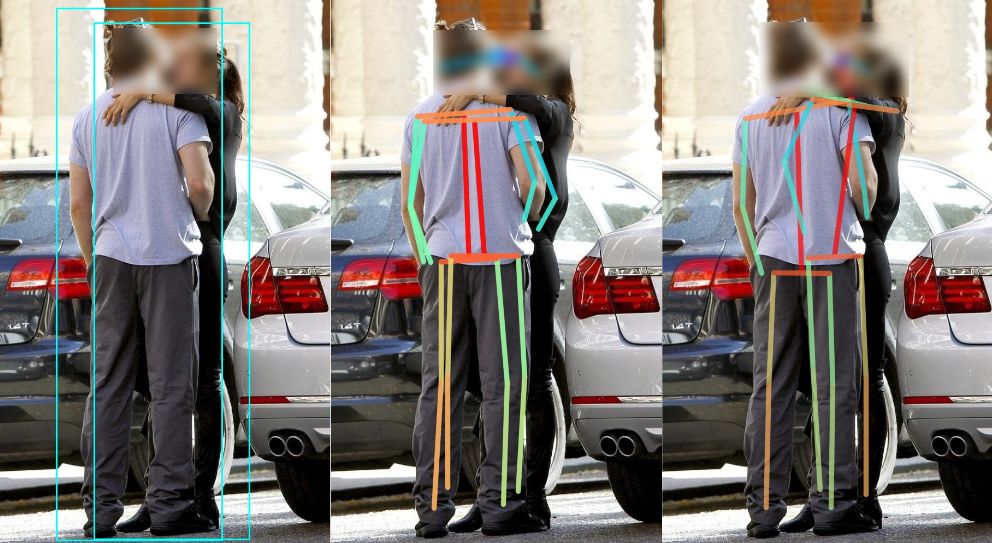}
\includegraphics[height=0.13\textheight,width=0.3\linewidth]{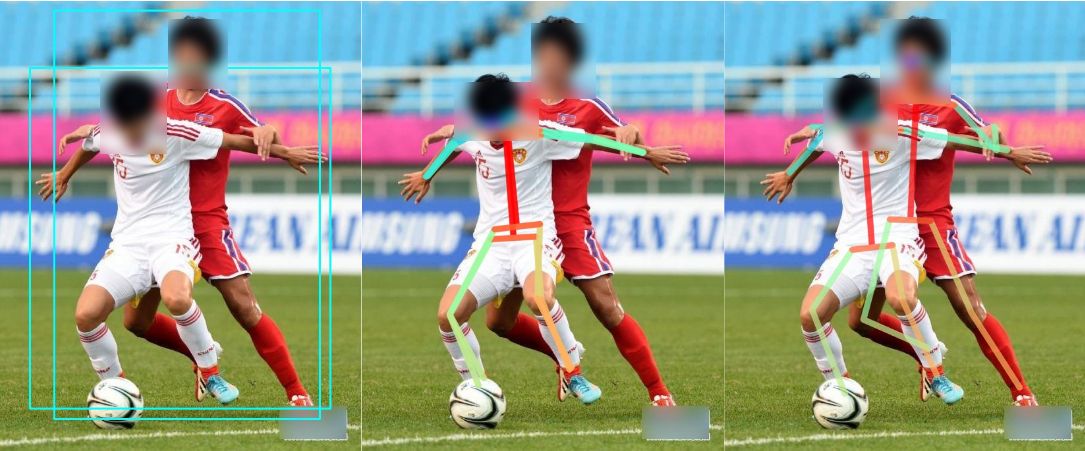}
\includegraphics[height=0.13\textheight,width=0.3\linewidth]{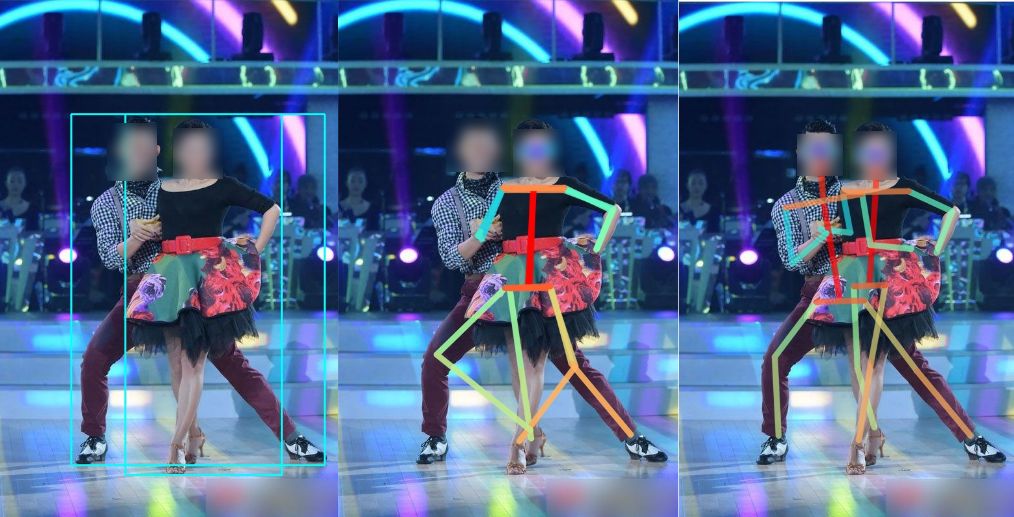}

\includegraphics[height=0.13\textheight,width=0.3\linewidth]{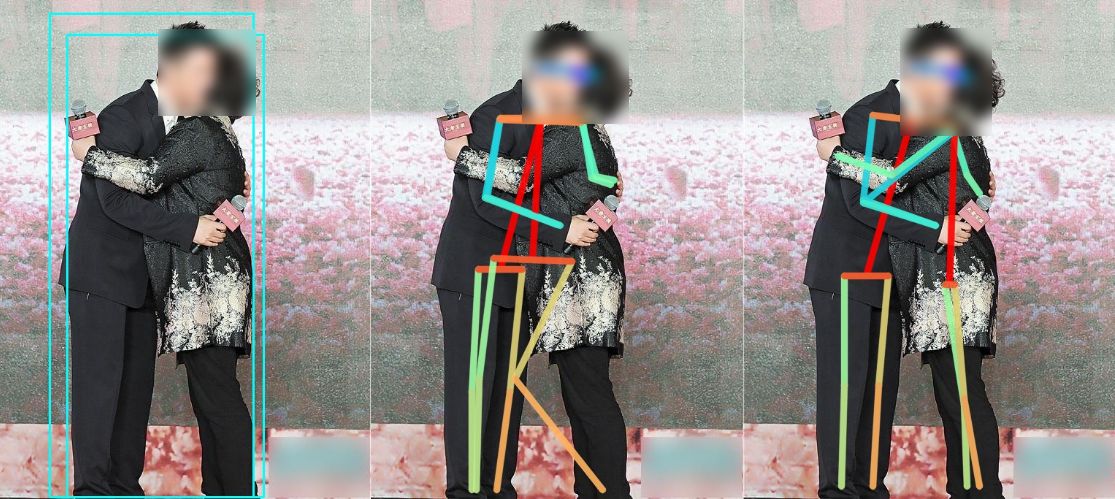}
\includegraphics[height=0.13\textheight,width=0.3\linewidth]{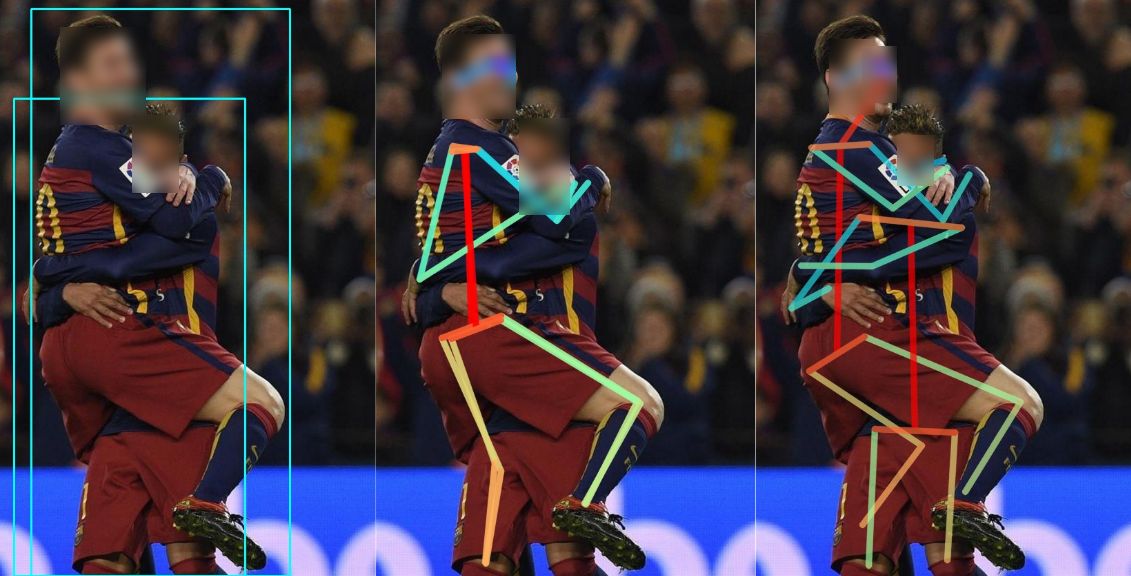}
\includegraphics[height=0.13\textheight,width=0.3\linewidth]{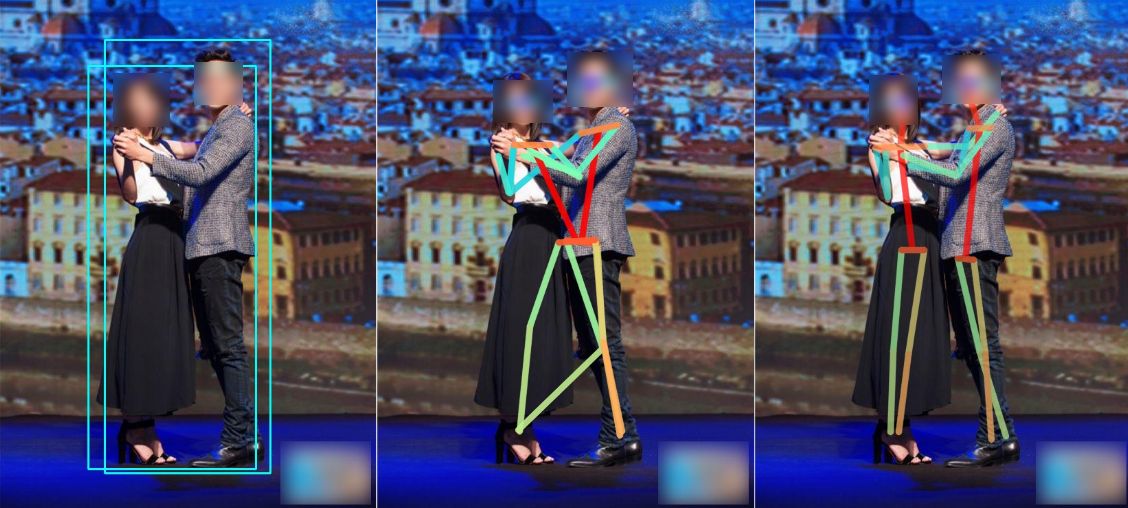}

\includegraphics[height=0.13\textheight,width=0.3\linewidth]{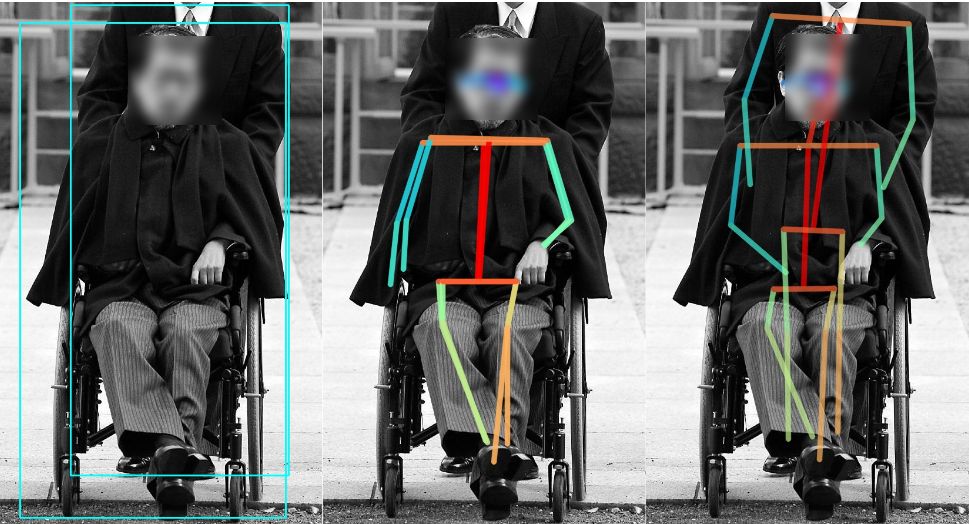}
\includegraphics[height=0.13\textheight,width=0.3\linewidth]{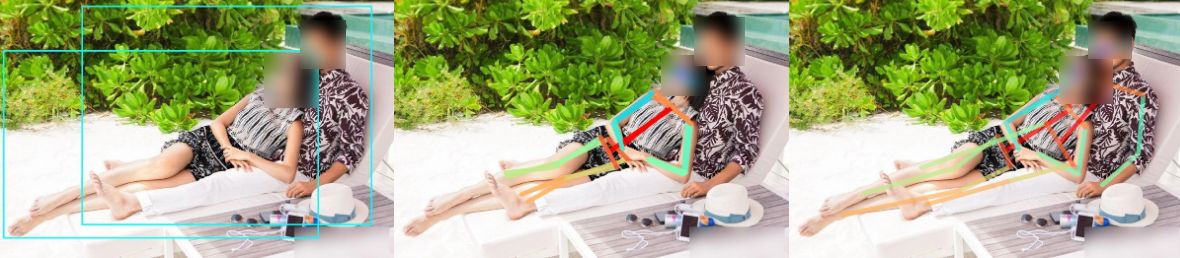}
\includegraphics[height=0.13\textheight,width=0.3\linewidth]{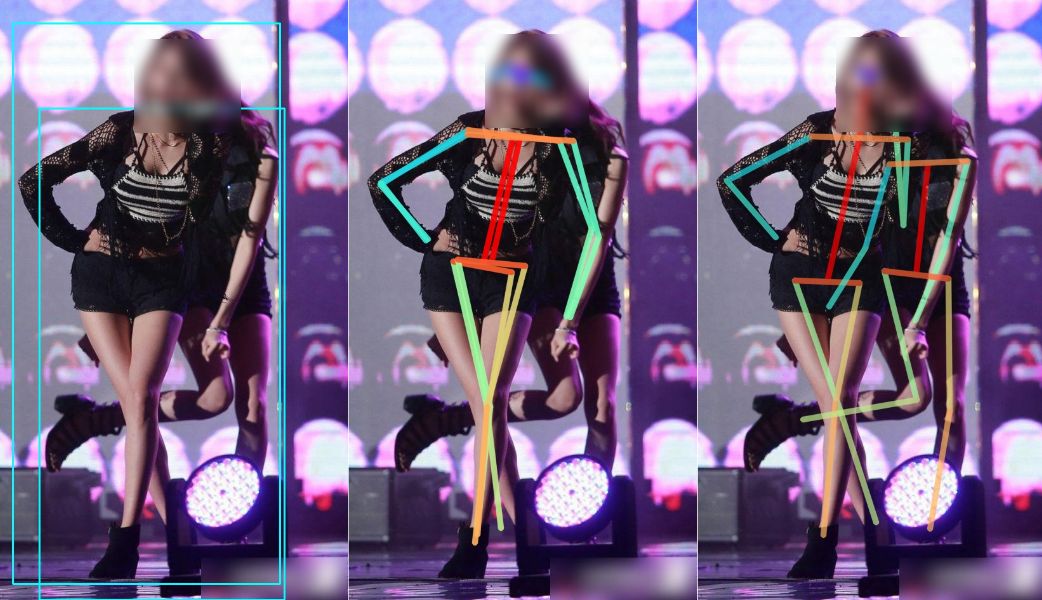}

\end{center}

\caption{Qualitative results of MIPNet. Each image (left to right) shows input bounding boxes, HRNet predictions and MIPNet predictions. }
\label{fig:supplementary:qualitative}
 \end{figure*}

\begin{figure*}
 \begin{center}
\includegraphics[height=0.13\textheight,width=0.3\linewidth]{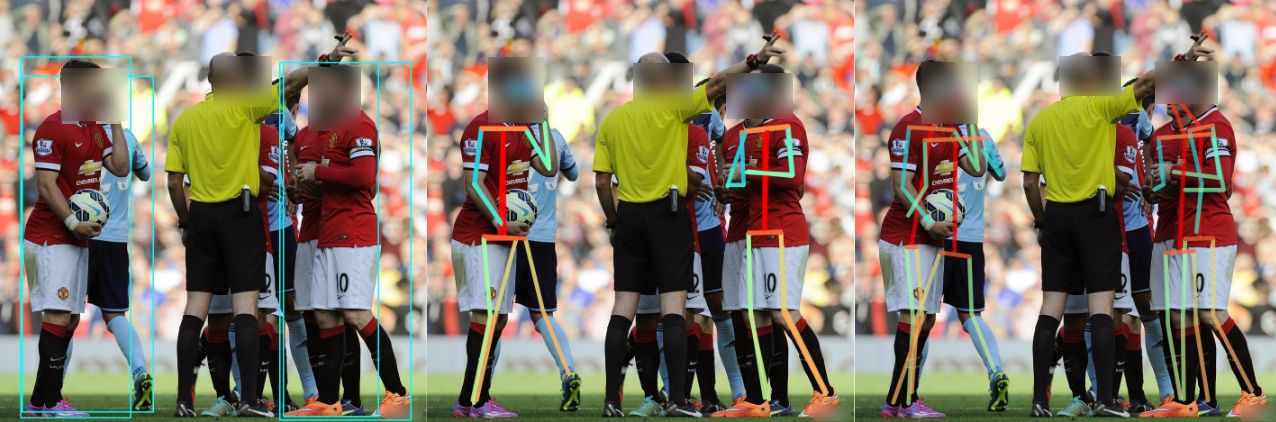}
\includegraphics[height=0.13\textheight,width=0.3\linewidth]{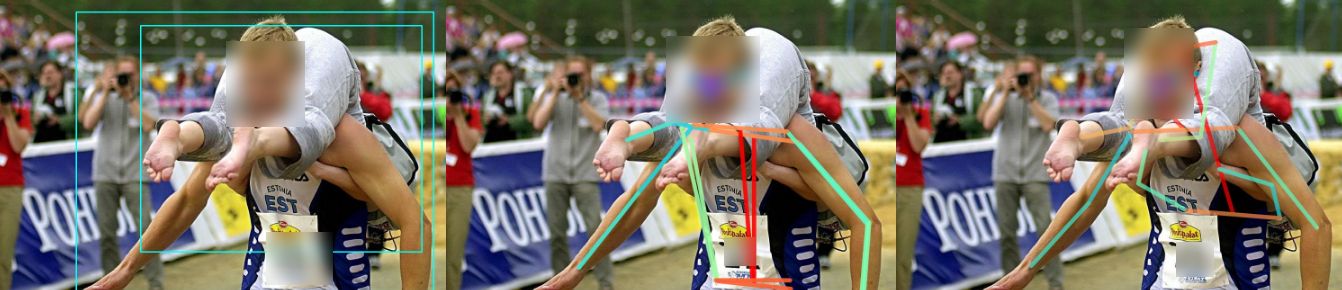}
\includegraphics[height=0.13\textheight,width=0.3\linewidth]{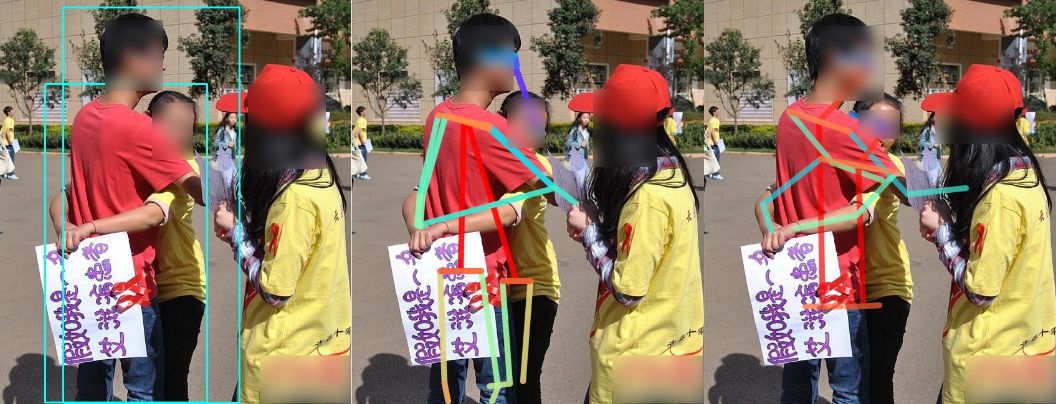}

\includegraphics[height=0.13\textheight,width=0.3\linewidth]{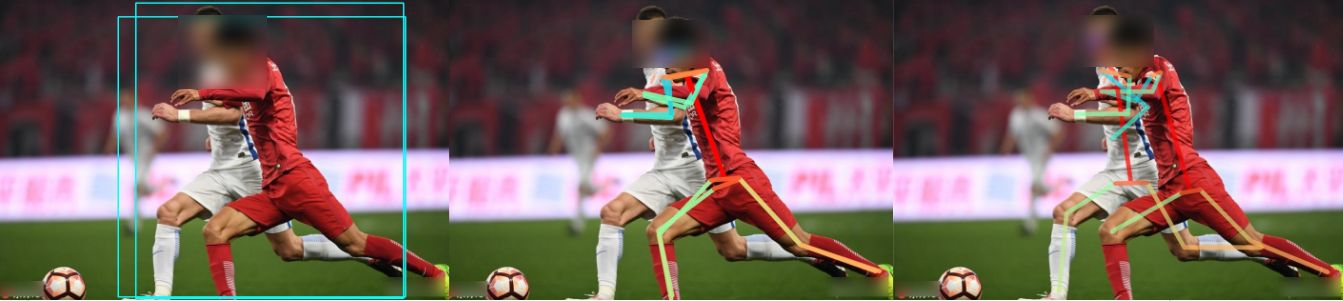}
\includegraphics[height=0.13\textheight,width=0.3\linewidth]{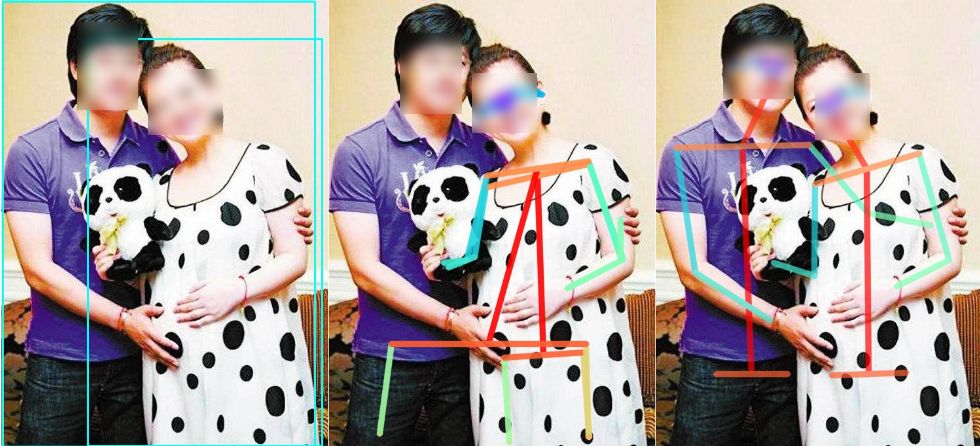}
\includegraphics[height=0.13\textheight,width=0.3\linewidth]{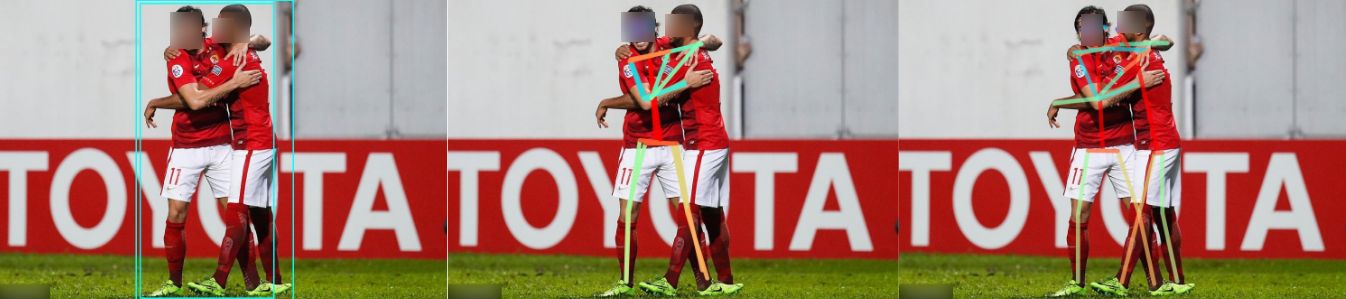}

\includegraphics[height=0.13\textheight,width=0.3\linewidth]{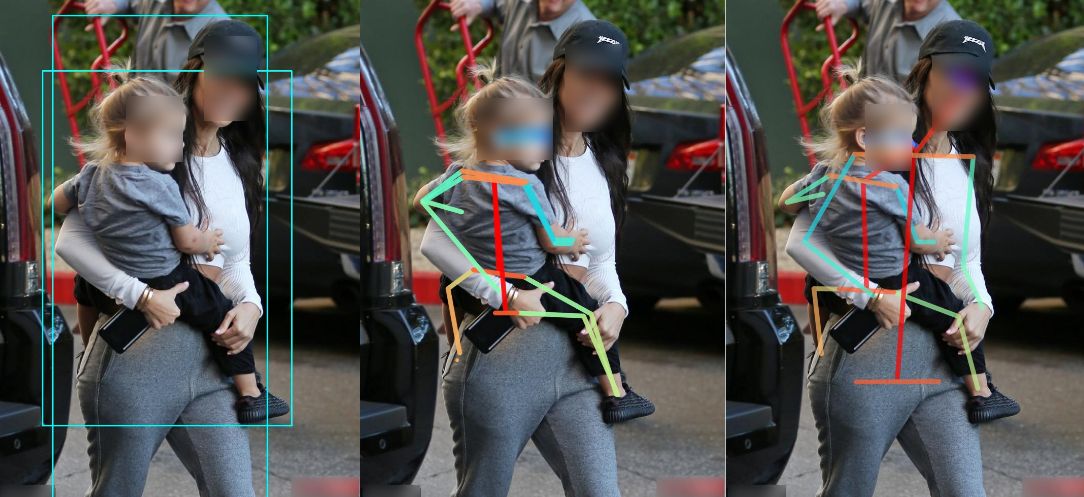}
\includegraphics[height=0.13\textheight,width=0.3\linewidth]{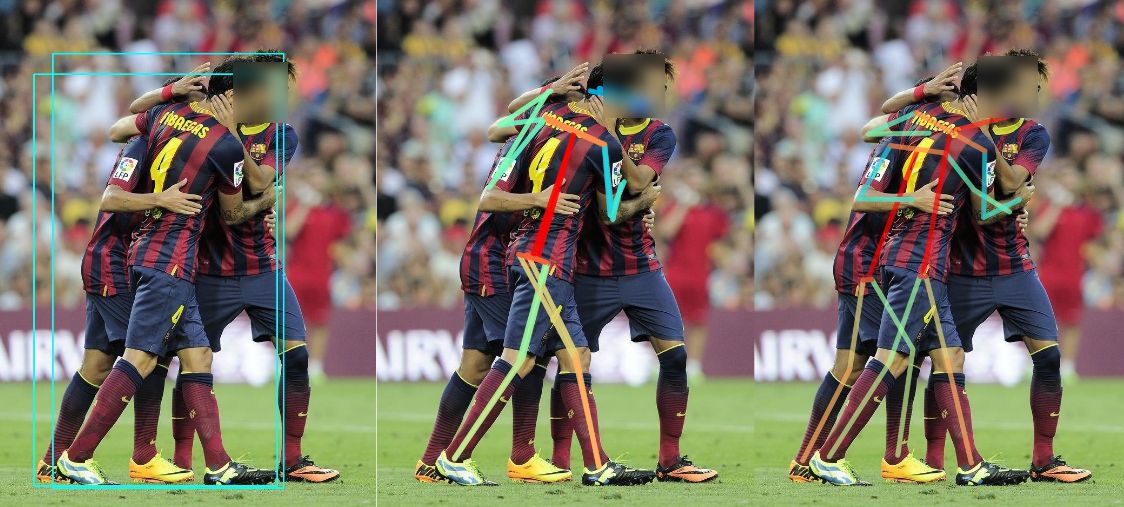}
\includegraphics[height=0.13\textheight,width=0.3\linewidth]{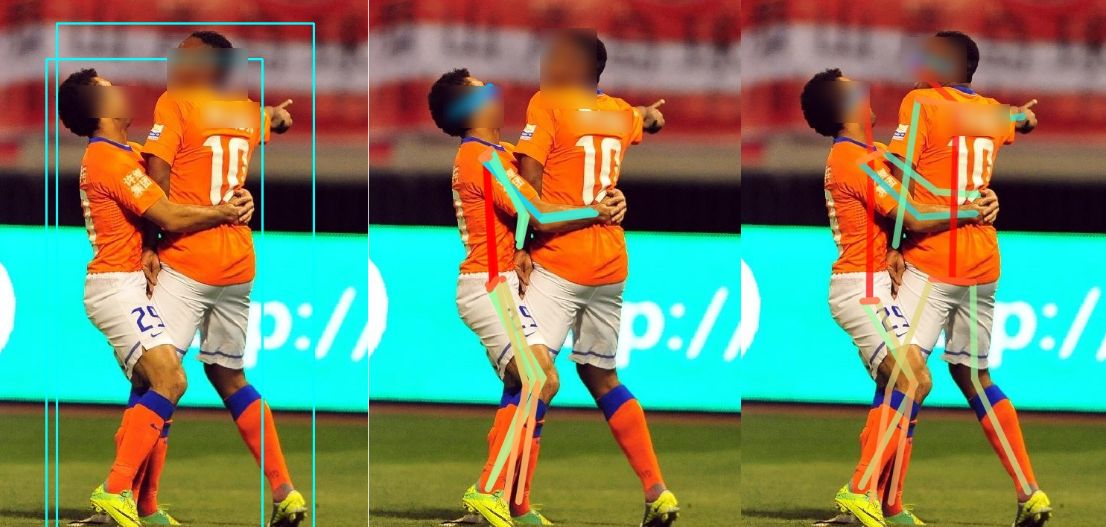}

\includegraphics[height=0.13\textheight,width=0.3\linewidth]{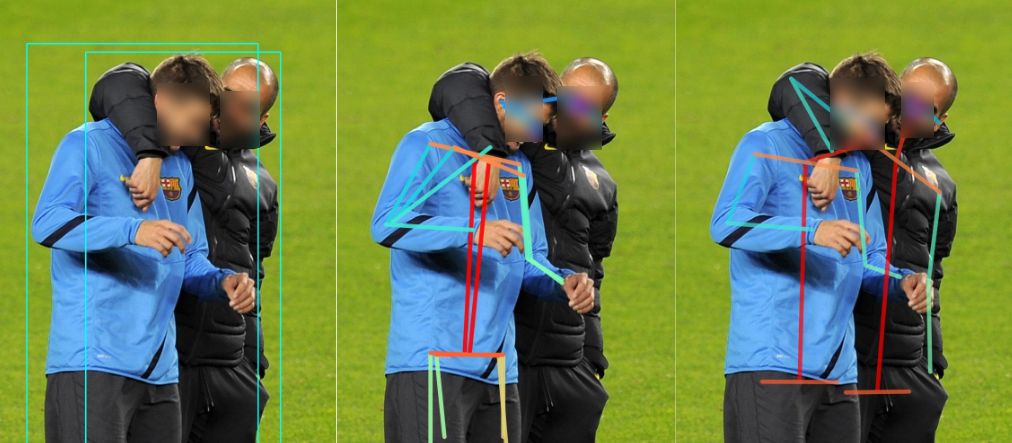}
\includegraphics[height=0.13\textheight,width=0.3\linewidth]{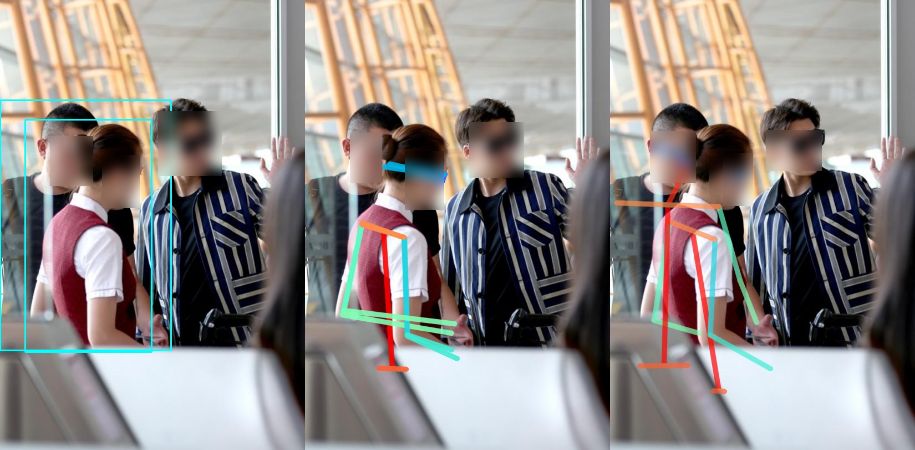}
\includegraphics[height=0.13\textheight,width=0.3\linewidth]{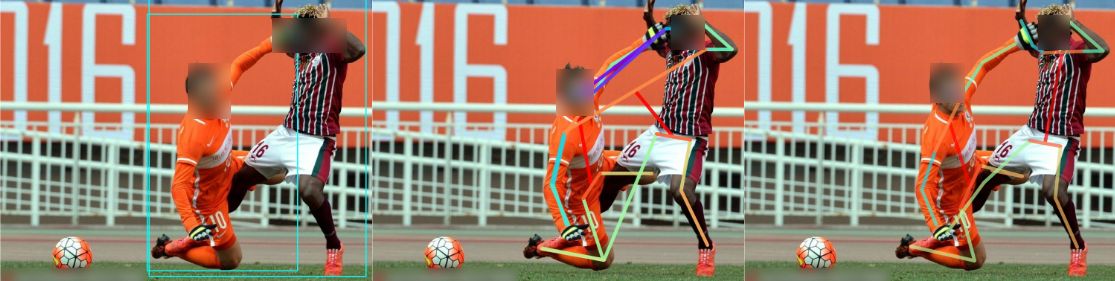}

\includegraphics[height=0.13\textheight,width=0.3\linewidth]{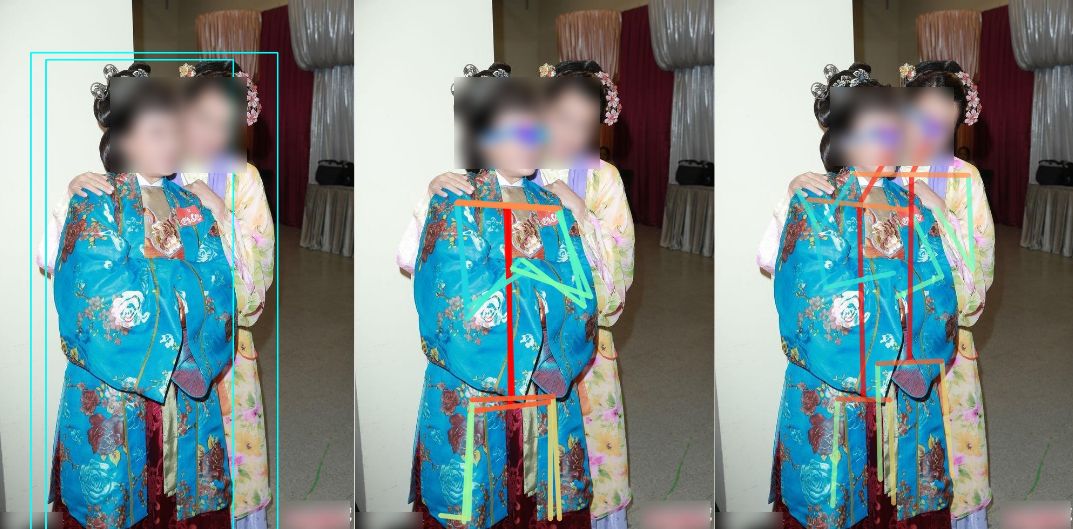}
\includegraphics[height=0.13\textheight,width=0.3\linewidth]{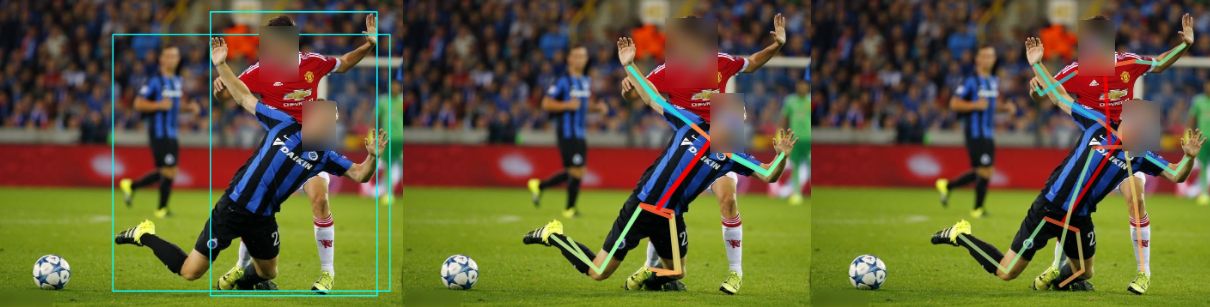}
\includegraphics[height=0.13\textheight,width=0.3\linewidth]{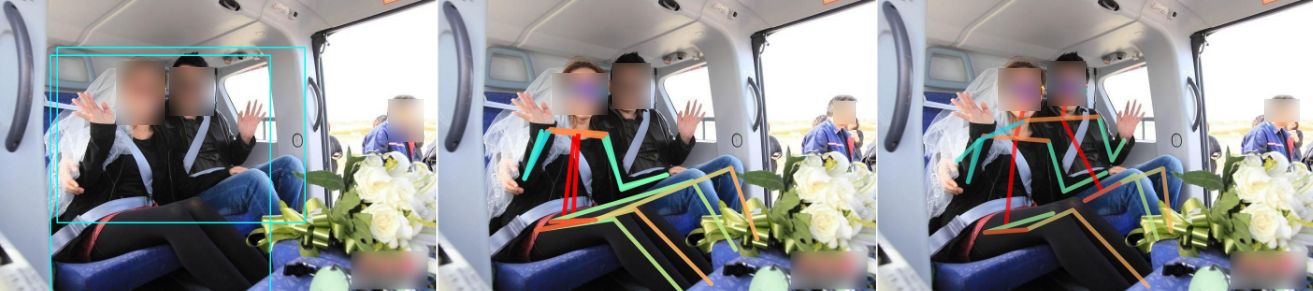}

\includegraphics[height=0.13\textheight,width=0.3\linewidth]{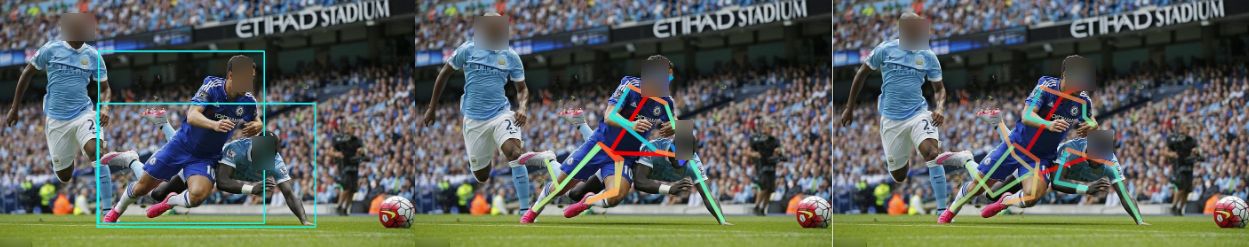}
\includegraphics[height=0.13\textheight,width=0.3\linewidth]{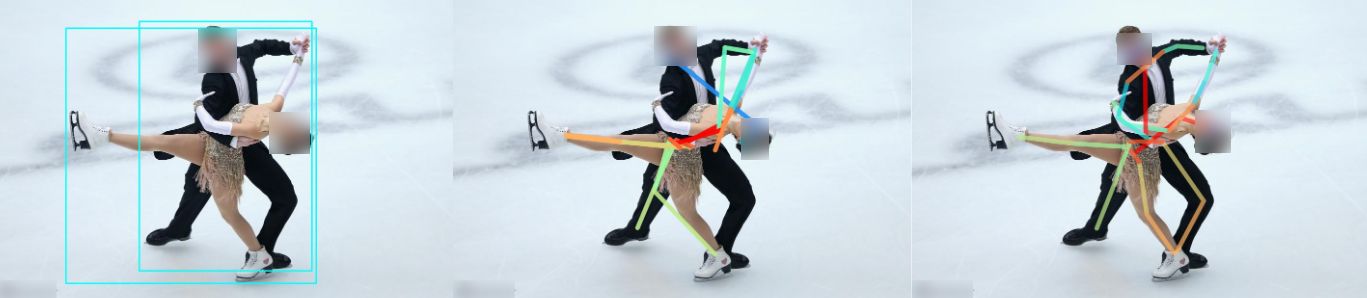}
\includegraphics[height=0.13\textheight,width=0.3\linewidth]{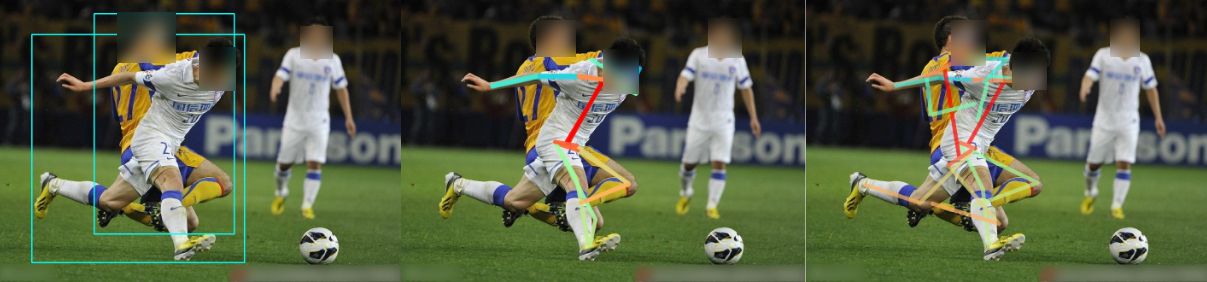}

\includegraphics[height=0.13\textheight,width=0.3\linewidth]{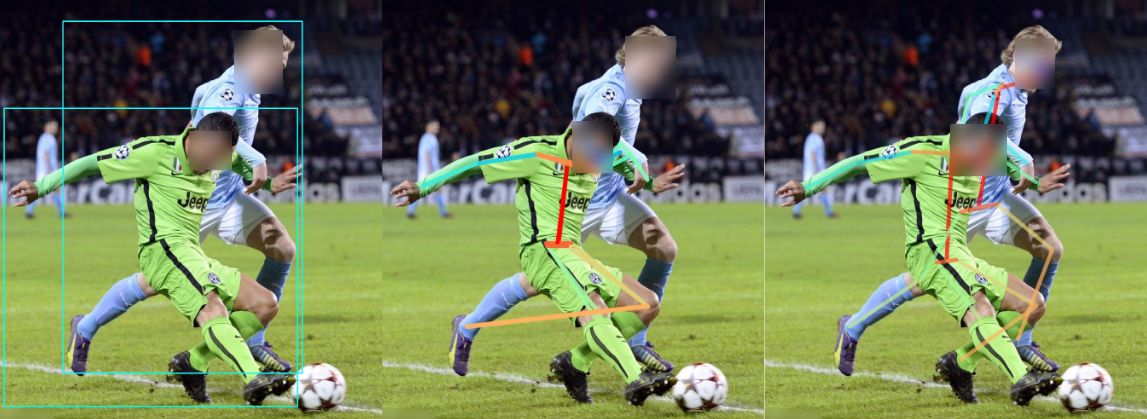}
\includegraphics[height=0.13\textheight,width=0.3\linewidth]{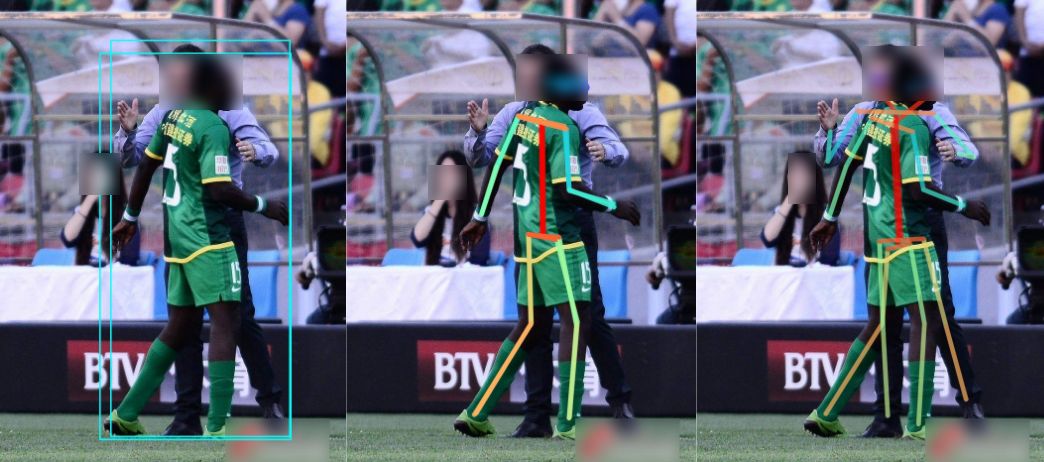}
\includegraphics[height=0.13\textheight,width=0.3\linewidth]{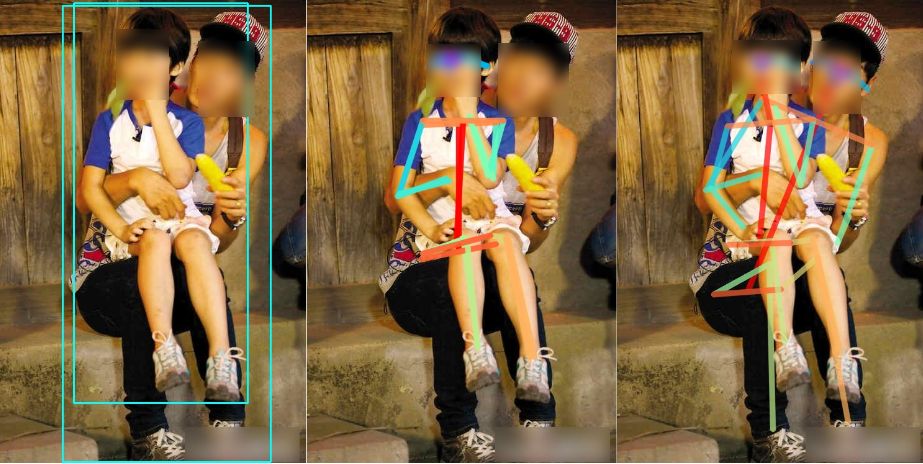}

\end{center}

\caption{Qualitative results of MIPNet. Each image (left to right) shows input bounding boxes, HRNet predictions and MIPNet predictions. }
\label{fig:supplementary:qualitative1}
 \end{figure*}

Figure~\ref{fig:supplementary:qualitative} and Figure~\ref{fig:supplementary:qualitative1} shows additional results on the OCHuman dataset, comparing MIPNet to HRNet. Note that in all of these cases, HRNet faces the problem of having highly overlapping bounding boxes because of the spatial proximity of humans in these images. Consequently, HRNet picks one dominant person and detects key-points on the same person within both bounding box instances. In contrast, MIPNet can clearly identify the correct set of key-points and associate them to the correct human(s) in each example. We especially want to point attention to the cases where people are  dancing in tandem, or tackling each other while playing sports. Such situations produce extremely complicated occlusions. However, MIPNet is able to successfully attribute the correct key-points to each human in the input bounding boxes in such situations, highlighting its usefulness in occlusion scenarios.
\newpage

 \newpage

\clearpage
\newpage
{\small
\bibliographystyle{ieee_fullname}
\bibliography{references}
}